\documentclass{article}
\usepackage{packages/arxiv}
\usepackage[utf8]{inputenc} 
\usepackage[T1]{fontenc}    
\usepackage{hyperref}       
\usepackage{url}            
\usepackage{booktabs}       
\usepackage{amsfonts}       
\usepackage{nicefrac}       
\usepackage{microtype}      
\usepackage{lipsum}		
\usepackage{graphicx}

\usepackage{doi}
\RequirePackage{amsmath, amssymb, algorithm, algorithmicx, algpseudocode}
\usepackage{physics}
\usepackage{float} 
\usepackage[numbers]{natbib}
\usepackage{subcaption}
\newsavebox{\measurebox}
\usepackage{longtable}
\usepackage{setspace}

\title{Real-Time Decorrelation-Based Anomaly Detection for Multivariate Time Series}

\date{} 					

\author{ 
        {Amirhossein Sadough} \\
	Dept. of Machine Learning and Neural Computing\\
	Radboud University\\
	Nijmegen, The Netherlands \\
	\texttt{amirhossein.sadough@donders.ru.nl} \\
	\And
        {Mahyar Shahsavari} \\
	Dept. of Machine Learning and Neural Computing\\
	Radboud University\\
	Nijmegen, The Netherlands \\
	\texttt{mahyar.shahsavari@donders.ru.nl} \\   
        \And
        {Mark Wijtvliet} \\
	Center of Competency\\
	ASMPT\\
	Beuningen, The Netherlands \\
	\texttt{mark.wijtvliet@asmpt.com} \\   
        \And
        {Marcel van Gerven} \\
	Dept. of Machine Learning and Neural Computing\\
	Radboud University\\
	Nijmegen, The Netherlands \\
	\texttt{marcel.vangerven@donders.ru.nl}  \\       
}

\renewcommand{\headeright}{}
\renewcommand{\undertitle}{}

\begin{document}
\maketitle


\begin{abstract}
Anomaly detection (AD) plays a vital role across a wide range of real-world domains by identifying data instances that deviate from expected patterns, potentially signaling critical events such as system failures, fraudulent activities, or rare medical conditions. The demand for real-time AD has surged with the rise of the (Industrial) Internet of Things, where massive volumes of multivariate sensor data must be processed instantaneously. Real-time AD requires methods that not only handle high-dimensional streaming data but also operate in a single-pass manner, without the burden of storing historical instances, thereby ensuring minimal memory usage and fast decision-making. We propose DAD, a novel real-time decorrelation-based anomaly detection method for multivariate time series, based on an online decorrelation learning approach. Unlike traditional proximity-based or reconstruction-based detectors that process entire data or windowed instances, DAD dynamically learns and monitors the correlation structure of data sample by sample in a single pass, enabling efficient and effective detection. To support more realistic benchmarking practices, we also introduce a practical hyperparameter tuning strategy tailored for real-time anomaly detection scenarios. Extensive experiments on widely used benchmark datasets demonstrate that DAD achieves the most consistent and superior performance across diverse anomaly types compared to state-of-the-art methods. Crucially, its robustness to increasing dimensionality makes it particularly well-suited for real-time, high-dimensional data streams. Ultimately, DAD not only strikes an optimal balance between detection efficacy and computational efficiency but also sets a new standard for real-time, memory-constrained anomaly detection.\end{abstract}

\section{Introduction}
Anomaly detection (AD) focuses on identifying data points that deviate from the expected patterns or structure within a dataset. These anomalies may arise due to unexpected processes affecting the data generation. For example, in chemistry, anomalies could result from an incorrectly conducted experiment, while in medicine, a disease may produce rare symptoms. In predictive maintenance, an anomaly might signal the onset of a potential system failure. The detection of anomalies in time series data has garnered significant attention in both statistics and machine learning. This is not surprising given the wide range of applications in which time series anomaly detection plays a critical role, from fraud detection to fault detection~\citep{bouman2024unsupervised, fisch2022real}.

The proliferation of sensors in the Internet of Things (IoT) has underscored the challenge of detecting anomalies in real-time, high-throughput streaming, multivariate data, making it a critical focus for both research and practical applications~\citep{fisch2022real}. Data from sensors and other connected devices provide real-time insights and enable targeted actions for industrial equipment and infrastructure. This network of interconnected industrial devices, sensors, machines, and manufacturing tools, facilitated by communication technologies, creates systems capable of monitoring, gathering, analyzing, and providing unprecedented insights~\citep{9235582, Jeschke2017, 9915308}.

The deployment of large-scale edge devices within the industrial IoT (IIoT) framework has enabled the development of various new edge-based applications, such as intelligent transportation, smart grids, digital health, smart logistics, and smart factories~\citep{lampropoulos2019internet, chen2017smart}. Edge devices facilitate efficient, scalable, and swift real-time decision-making, driving enhancements in quality and production. Edge computing allows for the deployment of advanced machine learning (ML) models in real-time applications like IoT anomaly detection, all while reducing processing delays, network traffic, and cloud expenses~\citep{borsatti2021enabling}. At the network edge, edge devices collect sensor data from numerous IIoT devices, process the incoming data, and use ML to identify patterns and monitor different operating conditions of the IoT devices~\citep{chen2017smart}. Consequently, these time series data are leveraged to recognize abnormal patterns and highlight anomalous behavior in industrial devices~\citep{liu2021towards, 9915308}.

Anomaly detection techniques are commonly divided into three main categories: supervised, semi-supervised, and unsupervised. Supervised approaches require a labeled dataset where both normal and abnormal instances are explicitly identified for model training and tuning. These methods typically offer high accuracy when labeled data is abundant. However, a known risk of supervised methods is that ground truth labels may not capture all types of anomalies during annotation. In contrast, semi-supervised methods use a small portion of labeled data combined with a larger amount of unlabeled data~\citep{ADbench}. Unsupervised methods work with entirely unlabeled data, making them highly adaptable to situations where labeling is impractical or unavailable~\citep{ahmad2017unsupervised}. Despite this flexibility, these methods tend to suffer from higher rates of false positives and can be sensitive to the inherent structure of the data. Each approach has its advantages and challenges, and the selection of the most appropriate method depends on the data characteristics and the specific needs of the anomaly detection task.

Unsupervised, data-driven anomaly detection for multivariate time series has become a common technique in ML, as these methods can detect unknown abnormal behavior automatically without needing a large collection of known anomalies~\citep{boniol2024dive}. Over the years, various unsupervised methods have been developed to detect anomalies, with some focusing on solving particular challenges like high-dimensionality, while others aim for more general anomaly detection with an emphasis on high performance or low computational complexity~\citep{bouman2024unsupervised}.

A recent benchmarking study on unsupervised anomaly detection methods for multivariate time series, presented by~\citep{bouman2024unsupervised}, aimed to answer the question: "How many unsupervised anomaly detection algorithms are needed for real-world data?" According to their findings, if a user lacks prior knowledge of whether their dataset contains local or global anomalies, the extended isolation forest (EIF) is the most suitable choice. In cases where a dataset is known or suspected to have local anomalies, $k$-nearest neighbors ($k$-NN) is the best-performing method. For datasets with only global anomalies, EIF stands out as the most effective method. Ultimately, they concluded that a combination of $k$-NN and EIF would be sufficient to achieve strong performance in anomaly detection.

The computational complexity of anomaly detection is crucial in determining the real-time throughput achievable on resource-constrained devices, particularly edge devices, which are essential for many applications. The performance of real-time anomaly detection reflects its ability to accurately identify outliers, directly influencing the overall functionality of the system. Therefore, when developing real-time anomaly detection methods, the primary focus should be on optimizing performance, especially in systems like IIoT, where operational efficacy is key to decision-making. While the throughput of a real-time anomaly detection method must meet the system's data rate requirements, it should not come at the cost of performance. Both performance efficacy and computational efficiency are essential in such environments.

The evaluation by~\citep{bouman2024unsupervised} lacks runtime measurements for the various methods, and their conclusions about which anomaly detection methods are most suitable for different scenarios do not provide sufficient information to generalize their benchmarking results to the selection of real-time anomaly detection methods. They claim that methods like EIF have the lowest computational complexity, making them typically more desirable in practice. However, based on our own intuition regarding runtime measurements, we found that EIF actually takes the longest time during their benchmarking runs.

Given the identified research gap in runtime evaluation of unsupervised anomaly detection methods for multivariate time series and the absence of effective approaches for real-time deployment, this work not only investigates the runtime performance of such methods but also introduces a novel real-time anomaly detection method for multivariate time series that effectively balances the trade-off between efficacy and throughput. 

The main contributions of this work are as follows. First, we propose decorrelation-based anomaly detection (DAD) as a new anomaly detection method  for multivariate time series, offering an optimal trade-off between efficacy and efficiency for real-time deployment. Second, we validate the proposed method through empirical evaluation using realistically generated synthetic data. Third, we introduce a practical hyperparameter tuning approach for anomaly detection benchmarking. While the proposed method achieves maximal performance, it maintains a relatively low computational complexity, making it suitable for real-time applications. Ultimately, the proposed real-time anomaly detection method emerges as the optimal solution for real-time exploitation on edge devices. 

\section{Background and Motivation}
Anomaly detection is essential across various domains, utilizing techniques from traditional statistical models to modern machine learning (ML), deep learning (DL), and hybrid ML/DL approaches~\citep{agyemang2006comprehensive}. The surge in time series data from IoT devices has introduced new complexities, particularly due to the temporal dependencies in IoT data -- current measurements often correlate with past observations~\citep{marjani2017big,han2019review}. These temporal correlations, known as short-term and long-term dependencies, are crucial for understanding and predicting sensor behavior~\citep{de200625}. Leveraging these dependencies can enhance anomaly detection in data streams, as unusual patterns may only emerge when contextualized within their historical timeline~\citep{9915308}. Therefore, an effective anomaly detection approach for streaming data must capture both the high-dimensional nature and temporal structure of these dynamic environments

\subsection{Classical Unsupervised Anomaly Detection Methods}
This work focuses on unsupervised anomaly detection with an emphasis on real-time applicability. It is important to first introduce classical unsupervised methods that have laid the foundation for many recent developments. These include distance-based, density-based, classification-based, tree-based, and spectral-based approaches.

\paragraph{Distance-based AD} Distance-based anomaly detection assumes that anomalous instances lie farther from their nearest neighbors than normal data points. These methods compute an anomaly score based on the proximity between a point and its k-nearest neighbors, making them effective in identifying outliers in spatially clustered datasets. Notable algorithms include $k$-NN~\citep{knox1998algorithms, knorr2000distance} and its variant $k$th-NN~\citep{ramaswamy2000efficient, samariya2023comprehensive}. In the standard $k$-NN approach, a point is considered anomalous if the average distance to its $k$ nearest neighbors is significantly larger than that of typical points. $k$th-NN refines this by focusing solely on the distance to the $k$-th nearest neighbor, improving sensitivity in detecting isolated outliers. Despite being distribution-agnostic, these techniques often suffer from high computational costs -- typically requiring $O(n^2)$ pairwise distance computations -- which limits their scalability to large datasets.

\paragraph{Density-based AD} Density-based anomaly detection operates on the principle that anomalies occur in regions where the local density significantly deviates from that of surrounding points. Prominent examples include local outlier factor (LOF)~\citep{LOFref} and connectivity-based outlier factor (COF)~\citep{tang2002enhancing}. LOF evaluates the anomaly score of a data point by comparing its local density to that of its neighbors, flagging instances with notably lower density as outliers. COF modifies the way density is computed, focusing more on the connectivity structure among neighbors. These approaches are popular due to their conceptual simplicity. However, they suffer from computational inefficiency, especially due to the need for pairwise distance calculations. They scale poorly with high-dimensional or large-scale data and are highly sensitive to the choice of parameters, particularly the neighborhood size $k$. Furthermore, they may underperform in datasets with heterogeneous density regions~\citep{samariya2023comprehensive}.

\paragraph{Classification-based AD} Classification-based approaches treat anomaly detection as a binary classification task, where a decision function is trained to separate normal instances from potential anomalies. In scenarios where only normal data is available during training, one-class classification techniques are employed. A prominent example is one-class support vector machine (OCSVM)~\citep{chandola2009anomaly}, which constructs a boundary that encloses the normal data. Any instance falling outside this region is considered an anomaly. While effective in certain low-dimensional settings, standard OCSVMs rely on fixed kernel assumptions (e.g., Gaussian or polynomial), which may not adapt well to complex data distributions. Additionally, these methods often struggle with scalability and performance in high-dimensional or large-scale datasets due to the curse of dimensionality and computational overhead~\citep{ruff2018deep}.

\paragraph{Tree-based AD} Tree-based methods partition data hierarchically using decision trees and are particularly effective in handling high-dimensional or non-Gaussian data distributions~\citep{barbariol2022review,belay2023unsupervised}. A prominent example is isolation forest (IF)~\citep{d12}, which isolates anomalies through recursive random splits, leveraging the idea that outliers can be separated more quickly due to their rarity and distinctiveness. IF constructs multiple isolation trees (iTrees) using randomly selected subsamples and random axis-aligned splits. During inference, it computes anomaly scores based on the average path length required to isolate each point. The extended isolation forest (EIF)~\citep{8888179} introduces randomly sloped cuts via hyperplanes, enhancing isolation in complex distributions. While EIF often yields performance improvements, it introduces higher computational and memory demands, limiting its use in resource-constrained environments. Moreover, tree-based methods typically assume global separability of anomalies, making them less effective in detecting local anomalies embedded in dense neighborhoods~\citep{samariya2023comprehensive}.

\paragraph{Spectral-based AD} Spectral-based methods project high-dimensional data into a lower-dimensional subspace that captures the most significant variability in the dataset. These techniques assume that anomalies deviate from the dominant patterns in this reduced space~\citep{chandola2009anomaly}. A widely used spectral method is principal component analysis (PCA)~\citep{aggarwal2015data}, which transforms the original correlated variables into a set of uncorrelated principal components ranked by variance. PCA achieves dimensionality reduction through the singular value decomposition of the empirical covariance matrix, retaining the principal components that represent the majority of the variance. This not only improves computational efficiency but also enables the detection of multivariate anomalies by quantifying how far each data point deviates from the learned subspace. A common approach is to compute the distance of each data point from the subspace defined by the top principal components and use this as an anomaly score~\citep{chandola2009anomaly}. However, PCA’s primary limitation lies in its static decorrelation matrix, which is computed once over the entire dataset. This makes it ill-suited for time-dependent data, as it is not capable of adapting to evolving temporal patterns. Consequently, PCA struggles to capture both short- and long-term dependencies in multivariate time series, reducing its effectiveness for dynamic anomaly detection scenarios.

\subsection{Motivation} 
The increasing dimensionality of modern features -- often containing tens to thousands of features -- poses significant challenges for real-time anomaly detection in data streams. High-dimensional data is prevalent in areas such as financial fraud detection~\citep{hemalatha2015minimal,cai2020mifi}, cybersecurity~\citep{xie2018line, 6200273, dong2018threaded}, and industrial fault monitoring~\citep{zhang2016sliding}, where identifying unusual behavior from continuous sensor measurements is crucial. Most unsupervised anomaly detection methods rely on proximity-based metrics like distance and density. However, these become ineffective in high-dimensional spaces due to data sparsity and the tendency of points to become nearly equidistant. This weakens the contrast between normal and abnormal points and obscures true anomalies with irrelevant features, a problem rooted in the curse of dimensionality~\citep{schubert2013generalized}.

Data streams further complicate the problem. Unlike static datasets, which are finite and stationary, streaming data is infinite, arrives rapidly, and often shifts in distribution over time. This demands algorithms that can process each data point as it arrives, using minimal memory and without revisiting past instances. Storing entire sequences is not viable; instead, the model must maintain a compact internal representation of relevant patterns. To address these challenges, existing solutions often use summarization techniques such as sliding windows, reservoir sampling, and forgetting mechanisms. For example, window-based approaches maintain only a portion of recent data, updating the model incrementally while discarding older points. While effective in some cases, these methods struggle with feature drift and concept evolution, as the fixed-size window may either fail to adapt to rapid changes or mix multiple evolving patterns if too large~\citep{souiden2022survey}.

The deployment of anomaly detection methods on edge devices -- particularly in IIoT applications -- intensifies the need for memory and compute efficiency. These devices must process high-throughput input rates and make decisions in real-time, yet they often lack the resources to store or process large volumes of high-dimensional data. To address these limitations, this work proposes a real-time anomaly detection method that avoids storing past data altogether. Instead, it learns and updates an internal representation of historical behavior in a single pass, allowing for memory-efficient, low-latency decision-making while maintaining awareness of short- and long-term temporal dependencies.

\section{Methodology}
\subsection{Problem Formulation}

Let $X = \{x_1, x_2, \dots, x_T\}$ represent a multivariate time series, where each observation $x_t \in \mathbb{R}^d$ represents a $d$-dimensional vector at time step $t$. A real-time anomaly detection method assigns an anomaly score $s_t \in \mathbb{R}$ to each observation $x_t$, quantifying its deviation from the expected behavior given the historical context $X_{1:t-1}$. The anomaly detection task can be formulated as learning a function $f(\cdot)$ that maps the current observation and its historical context to an anomaly score:
\[
s_t = f(x_t \mid X_{1:t-1}; \theta)
\]
where $\theta$ denotes the hyperparameters of the model. The objective of the anomaly detection model is to learn $f(\cdot)$ such that it effectively differentiates between normal and anomalous observations based on patterns in historical data.

\subsection{Decorrelation-based Anomaly Detection}
We are motivated by the fact that anomalies often disrupt the expected correlation structure or interrelations of features in time series data. The goal of multivariate analysis is to understand these interrelations and detect deviations that indicate potential abnormalities~\citep{rousseeuw1999fast}. The relationships among features in multivariate time series inherently give rise to dynamic correlation structures, shaped by the temporal dependencies between them. Therefore, we hypothesize that a significant shift in these interrelations at a given time step, compared to the historical pattern, can signal an anomaly. These correlation structures capture the inherent relationships within the data, and disruptions in these patterns are indicative of abnormal behavior.

Hence, we aim to develop an approach to effectively learn interrelations. To achieve this, we draw inspiration from the work of~\citep{ahmad2022constrained}, where they introduce a method for decorrelating layer inputs in neural network training. This technique enhances training efficiency by mitigating feature correlations that can act as a barrier to learning~\citep{ahmad2024correlations}. In their approach, layer inputs are decorrelated using a trainable decorrelation matrix $R \in \mathbb{R}^{d\times d}$ , where an input vector $x$ is decorrelated through multiplication by $R^T$, yielding the decorrelated input $\hat{x}=xR^T$. The decorrelation matrix $R$ is then updated according to 
\begin{equation}
\label{eq1}
R \leftarrow R - \eta \left(\hat{x}^T\hat{x} - \text{diag}\left(\hat{x}^T\hat{x}\right) \right)R
\end{equation}
where $\eta$ is a small constant learning rate and $R$ is initialised as the identity matrix $I_d$. Motivated by this approach, we adapt their learning rule for time series applications, specifically for real-time anomaly detection, where understanding the interrelations of features is crucial. 


In this context, we use the concept of decorrelation as the foundation for learning a matrix that can effectively decorrelate a given input based on its historical values. We assume that, after sufficient exposure to samples, the learned decorrelation matrix will capture the underlying statistical structure of the input sequence. Specifically, it will enable the transformation of the current input $x_t$ into a decorrelated representation through the operation $\hat{x}_t=x_{t}R_{t}^{T}$ where $R_t$ is the learned decorrelation matrix at time $t$. So, outliers are detected when the converged decorrelation matrix fails to perform effectively which indicates that the interrelations of the data at time step $t$ have significantly changed. This deviation manifests as a shift in the decorrelation matrix’s learning, which will be illustrated in Section~\ref{sec:Decorrelation_Learning}. Returning to Eq.~\eqref{eq1}, we now extend it to the decorrelation problem for multivariate time series data:
\begin{equation}
\label{eq2}
R_t \gets R_{t-1} - \frac{\eta}{(d-1)} \left(\hat{x_t}^T\hat{x_t} - \text{diag}\left(\hat{x_t}^T\hat{x_t}\right) \right)R_{t-1}
\end{equation}
where $\flatfrac{1}{(d-1)}$  acts as a normalization factor, ensuring that the update is appropriately scaled based on the number of dimensions $d$. By tracking the behavior of the decorrelation matrix $R$, the magnitude of change in the decorrelation matrix reflects the learning dynamics, which can be interpreted as an anomaly score for $x_t$. To measure this, we compute the absolute magnitude of the change in the matrix $R$ between consecutive time steps, given by $\left| \| R_t \| - \| R_{t-1} \| \right|$, where $ \| \cdot \| $ denotes the Frobenius norm. To enhance the stability of the anomaly scores and reduce the sensitivity to minor fluctuations, we incorporate a momentum term. This momentum smooths the score over time by combining the current value with the previous score, as described by the following update rule:
\begin{equation}
\label{eq4}
s_t \gets (1 - \gamma) \cdot s_{t-1} + \gamma \cdot \left| \| R_t \| - \| R_{t-1} \| \right|
\end{equation}
where $\gamma$ is the momentum factor. To further stabilize and mitigate the effects of normal fluctuations among features, the update rule for the decorrelation matrix in (\ref{eq2}) is extended to incorporate temporal information as follows:
\begin{equation}
\label{eq5}
R_t \gets R_{t-1} - \frac{\eta}{(p+1)(d-1)} \left((\hat{x}_{t-p:t}^{T}\hat{x}_{t-p:t}) - \text{diag}\left(\hat{x}_{t-p:t}^{T}\hat{x}_{t-p:t}\right) \right)R_{t-1}
\end{equation}
where $p$ represents the temporal window size (\textit{i.e.}, specifies the number of past samples contributing to the decorrelation matrix update). This sliding window-based decorrelation mechanism helps dampen the impact of minor fluctuations that do not indicate anomalies, as the update integrates both current and past observations. Algorithm~\ref{alg:DAD_alg} presents the pseudo-code for the proposed real-time anomaly detection method.

\begin{algorithm}[t]
\caption{Real-time Anomaly Detection using DAD}
\label{alg:DAD_alg}
\begin{scriptsize}
\begin{algorithmic} [1]
\State \textbf{Input:} Stream $X = \{x_1, x_2, \dots, x_T\}$ ($x_t \in \mathbb{R}^d$), sliding window size $p$, learning rate $\eta$, momentum factor $\gamma$
\State \textbf{Output:} $s_t$ anomaly score assigned at time step $t$
\State \textbf{Initialize:} $R_0 \gets {I}_d$, $s_0 \gets 0$
\For{$t = 1$ to $T$}
\If{$t > p$}
    \State $R_t \gets R_{t-1} - \frac{\eta}{(p+1)(d-1)}  \left[ \left( x_{t-p:t} R_{t-1}^T \right)^T \left( x_{t-p:t} R_{t-1}^T \right) - \operatorname{diag} \left( \left( x_{t-p:t} R_{t-1}^T \right)^T \left( x_{t-p:t} R_{t-1}^T \right) \right) \right] R_{t-1} $
    \State $s_t \gets (1 - \gamma) \cdot s_{t-1} + \gamma \cdot \left| \| R_t \| - \| R_{t-1} \| \right|$
\Else
    \State $R_t \gets {I}_d$
    \State ${s}_t \gets 0$
\EndIf  
\EndFor
\end{algorithmic}
\end{scriptsize}
\end{algorithm}

Several key observations can be drawn from the introduced method. First, no valid anomaly score can be expected for the initial observations. This is logical as unsupervised real-time anomaly detection methods cannot detect anomalies without first being exposed to enough data to learn the underlying patterns and build a statistical model of normal behavior. Second, the assigned anomaly scores cannot be guaranteed to be accurate until the decorrelation matrix converges. This aligns with the initialization delay in unsupervised real-time anomaly detection applications, which is an inherent characteristic of such methods, as discussed in~\citep{vajda2024machine}. The length of this initialization delay may vary across different methods, but it is essential for the system to learn the baseline patterns before anomaly detection becomes effective.
Third, the decorrelation matrix, influenced by the momentum factor, learns the interrelations of the majority of the data samples and converges on this majority. This suggests that the problem should predominantly consist of normal samples, as the model will mainly capture the interrelations of typical patterns over time. This aligns with the underlying assumption that anomalous data is rare, as highlighted in~\citep{chandola2009anomaly,d12}, where the focus is on learning the majority class (normal data). Finally, the method’s ability to differentiate between normal and abnormal samples is highly sensitive to the hyperparameter $\eta$. This sensitivity plays a crucial role in the method’s performance, and further analysis of this hyperparameter is provided in Section~\ref{sec:Hyper_sen}.

\begin{algorithm}[t]
\begin{scriptsize}
\caption{Automated DAD}
\label{alg:DAD_alg_aut}
\begin{algorithmic} [1]
\State \textbf{Input:} Stream $X = \{x_1, x_2, \dots, x_T\}$ ($x_t \in \mathbb{R}^d$), learning rate set ${\eta}_{G} = \{\eta_0,\eta_1,\eta_2, \dots,\eta_{g-1}\}$, exploration length $n$, momentum factor $\gamma$
\State \textbf{Output:} $s_t$ anomaly score assigned at time step $t$
\State \textbf{Initialize:} ${R}_{i,0} \gets {I}_d$
\For{$t = 1$ to $T$}
\If{$t \geq n$} \Comment{Operating Phase: Anomaly Detection}
    \State $R_t \gets R_{t-1} - \frac{\eta}{d-1} \left[ \left( x_t R_{t-1}^T \right)^T \left( x_t R_{t-1}^T \right) - \operatorname{diag} \left( \left( x_t R_{t-1}^T \right)^T \left( x_t R_{t-1}^T \right) \right)\right]  R_{t-1}$
    \State $s_t \gets (1 - \gamma) \cdot s_{t-1} + \gamma \cdot \left| \| R_t \| - \| R_{t-1} \| \right|$
\Else \Comment{Burn-in Phase: Determining the Optimal Learning Rate}
    \For{$i = 0$ to $(g-1)$}
        \State ${R}_{i,t} \gets {R}_{i,t-1} - \frac{{\eta}_{G}[i]}{d-1} \left[ \left( x_{t} {R}_{i,t-1}^T \right)^T \left( x_{t} {R}_{i,t-1}^T \right) - \operatorname{diag} \left( \left( x_{t} {R}_{i,t-1}^T \right)^T \left( x_{t} {R}_{i,t-1}^T \right) \right) \right] {R}_{i,t-1}$
        \State $c_{i,t} = \frac{1}{d^2} \sum_{row}^{d}\sum_{col}^{d} \left[ \left( x_t R_{i,t}^T \right)^T \left( x_t R_{i,t}^T \right) \right]_{row,col}$ 
    \EndFor
    \If{$t=(n-1)$}
        \State $i^\star \gets \arg\max_i \left\{ \bar{c}_i \mid \bar{c}_i = \frac{1}{n} \sum_{t=0}^{n{-}1} c_{i,t},~ \bar{c}_i \leq 1.025 \min_j \bar{c}_j \right\}$
        \State ${R}_{n-1} \gets {R}_{i^{\star},n-1}$ , ${\eta} \gets {\eta_G}[i^{\star}]$
    \EndIf
    \State ${s}_t \gets 0$    
\EndIf  
\EndFor
\end{algorithmic}
\end{scriptsize}
\end{algorithm}

To simplify the real-time deployment of our method, where labeled data is often unavailable in real-world scenarios, we propose a solution to automate the hyperparameter selection internally, eliminating the need for dedicated effort in manually tuning the optimal hyperparameter for each specific case. Our assumption is that correlation exists in multivariate time series, and the DAD method operates based on a decorrelator core that identifies deviations from historical patterns. By measuring how well this decorrelator core reduces the correlation over time, we can obtain the minimum necessary information to assess whether the method is functioning correctly.

To achieve this, we propose an exploration phase that occurs when the method is first executed and receives input data. This phase involves measuring the correlation level sample by sample for a set of hyperparameters. Notably, sliding window-based decorrelation is excluded from this automation process in order to reduce the complexity of the hyperparameter grid. The objective is to identify a hyperparameter that minimizes the correlation level effectively. We specifically seek a hyperparameter that ensures the decorrelation process functions as intended, while also balancing the learning speed. A moderate learning rate is necessary to prevent learning too quickly, as a high learning rate may hinder the ability to distinguish deviations from the normal pattern. This assumption is crucial for accurately detecting deviations during the learning process. By maintaining a reasonable decorrelation rate, the model can adapt to changes without being overly sensitive to noise or outliers. 

The pseudo-code for the automated version of DAD is provided in Algorithm~\ref{alg:DAD_alg_aut}, where an empirical approach is employed to ensure the algorithm effectively decorrelates the input. As shown, the exploration length $n$ determines the number of time steps dedicated to hyperparameter finding from a learning rate set $\eta_G =\{\eta_0,\eta_1,\eta_2,\dots,\eta_{g-1}\}$ based on the described concept. A recommended learning rate set is provided in Section~\ref{sec:synth_bench_setup}. To ensure sufficient data for robust selection, we suggest a minimum of $n = 50$ time steps during this exploration phase. Once these samples are processed, the optimal hyperparameter satisfying our assumptions is chosen, and the real-time operation begins. Also, a momentum factor $\gamma = 0.25$ is suggested for stability. From that point on, the algorithm assigns an anomaly score at each time step based on the selected hyperparameter. The performance of the automated variant of DAD will be discussed in Section~\ref{sec:Hyper_sen}, where the hyperparameter sensitivity is analyzed.

\subsection{Hyperparameter Tuning: A Practical Approach}
\label{sec:hyper_opt}
Hyperparameter selection plays a crucial role in the performance of anomaly detection methods. Despite its importance, the literature lacks consensus on how to effectively tune these parameters when comparing algorithms in practice. 

According to~\citep{HPO_ref}, researchers typically address this issue using one of two approaches. The first, more conservative strategy, involves bypassing the tuning process altogether and instead applying a fixed hyperparameter configuration across all datasets~\citep{d1,d8,d9,d20,d22,d23}. This approach aligns with the idea of relying on "reasonable defaults". Conversely, the alternative strategy focuses on selecting hyperparameters that maximize the detector's performance for each individual dataset (peak performance)~\citep{campos2016evaluation,p3,p4,p7}. Both of these approaches come with notable limitations. Neither method effectively addresses the fundamental question posed during the empirical evaluation of a new anomaly detection method: can the proposed algorithm demonstrate practical utility by outperforming existing methods on certain datasets? Relying on default settings may underestimate an algorithm's potential, especially for methods that are highly sensitive to hyperparameter adjustments. Conversely, optimizing hyperparameters to maximize performance directly on the test set introduces a major methodological flaw, as this constitutes "tuning on the test set".  Moreover, both strategies are unlikely to reflect real-world scenarios accurately. In practical applications, users are unlikely to rely solely on default settings and would typically attempt some form of parameter tuning.

Another approach involves averaging performance across a range of hyperparameter values~\citep{bouman2024unsupervised}. However, this method raises concerns about its practical relevance, as real-time applications ultimately require a single, well-tuned configuration. Moreover, this averaging strategy may obscure the potential of methods that are more sensitive to hyperparameter tuning yet capable of outperforming others when properly optimized. 

Therefore, we adopt a more practical evaluation approach for anomaly detection methods, grounded in realistic hyperparameter tuning. This strategy aims to bridge the gap between theoretical evaluation and real-world, real-time applications. To address the evaluation challenge of anomaly detection methods, a practical strategy was introduced in~\citep{HPO_ref}, which advocates using a small labeled validation set for hyperparameter tuning. This method strikes a balance between minimizing the effort required to obtain labeled data and achieving sound, reproducible evaluations. The proposed strategy ensures that the validation set contains at least one anomaly and that its contamination ratio closely reflects that of the full dataset. By doing so, this method offers a realistic and practical solution for hyperparameter optimization in anomaly detection.

Inspired by this approach, we propose an enhancement to further improve the representativity of the validation set. Specifically, our method involves downsampling the dataset at ratios between 10\% to 60\% in steps of 10\%. Among the downsampled subsets, we select the one that demonstrates the highest histogram similarity to the original dataset using the Jensen-Shannon divergence (JSD) with 20 bins. Additionally, we ensure that the chosen validation set maintains a non-zero contamination ratio to preserve the presence of anomalies. This strategy aims to maximize the representativeness of the validation set, ultimately improving the reliability of hyperparameter tuning in real-time anomaly detection applications.

Algorithm~\ref{alg:find_L} presents the pseudo-code for finding the optimal subset $ D^{*}_j \subset D_j $ of each dataset $D_j$ based on our proposed enhancement. After obtaining the optimal subsets, each method $M$ in the evaluation explores its optimal hyperparameters $\theta_j^*$ across its own hyperparameter grid $\Theta$ on individual subsets by maximizing the quality criterion $q$, formulated as follows: 
\begin{equation}
\label{optimal_hyperparameter}
\theta_j^* = \arg\max_{\theta \in \Theta} q(M(\theta, D^*_j)) \,.
\end{equation}
Thus, each method in the evaluation identifies its optimal hyperparameters for each dataset individually. This aligns with the common practice of tuning hyperparameters based on a small labeled subset, aiming to optimize the algorithm's performance on this subset~\citep{ruff2019deep}. This approach also allows us to assess the algorithm's ability to generalize to the entire dataset after tuning the hyperparameters using a subset. This mirrors practical scenarios where, after tuning, the method is expected to maintain consistent performance with the subset. Consequently, our proposed method not only enhances hyperparameter tuning by incorporating a representatively downsampled validation set but also evaluates the robustness and performance of the tuned method across various datasets.

\begin{algorithm}[t]
\begin{scriptsize}
\caption{Finding the optimal subset $D_j^{*}$}
\label{alg:find_L}
\begin{algorithmic}[1]
\State \textbf{Input:} dataset $D_j \in \mathbb{R}^{k\times d}$ with sample length $k$ and number of features $d$ , $L_R = \{l_0, l_1, \dots, l_{r-1}\}$ downsampling ratios, number of histogram bins $B$
\State \textbf{Output:} \( D_j^{*} \)
\State $H_j \gets \text{Hist}(D_j;B)$, $H_j \in \mathbb{R}^{d\times B}$   
\For{each \( l \in \{l_0, l_1, \dots, l_{r-1}\} \)}
    \State \( D_j^{l} \gets \text{Downsample}(D_j; l) \)
    \State $H_j^{l} \gets \text{Hist}(D_j^{l};B)$, $H_j^{l} \in \mathbb{R}^{d\times B}$ 
    \State $JSD^{l} \gets \frac{1}{d}\sum_{i=0}^{d-1} JSD(H_j^{l}[i]\|H_j[i])$
\EndFor
\State \( {L} \gets \arg\min_l JSD^{l} \)
\State \( D_j^{*} \gets \text{Downsample}(D_j; L) \)
\end{algorithmic}
\end{scriptsize}
\end{algorithm}

As highlighted by~\citep{HPO_ref}, one may wonder why this standard practice has not yet become common in anomaly detection. A possible explanation lies in the requirement for a labeled data subset, which seemingly contradicts the principles of unsupervised learning. However, based on observations, assuming that practitioners are willing to provide a small amount of labeled data in exchange for improved anomaly detection accuracy is far more reasonable than assuming they would rely solely on default settings or attempt to guess the optimal hyperparameters for achieving peak performance. This enhanced strategy will be used in our anomaly detection benchmarking on real-world data.

\subsection{Empirical Validation using Synthetic Data}
\label{sec:EmpiricalValidation}

This section presents experimental setup to evaluate DAD's performance on a specific task based on a synthetic data. This synthetic data designed to study some types of abnormalities and DAD's corresponding ability to capture them. Using the synthetic data we provide an overview of a potential problem in the literature that our proposed method can deal with it. Furthermore, we test our method on real-world data in the following sections.

\subsubsection{Synthetic Data Generation}
\label{sec:SynthDataGen}

To evaluate the proposed anomaly detection method, we generate synthetic multivariate time series by switching between two different Gaussian distributions at predefined time points. These distributions differ in their mean, covariance structure, or correlation strength, thereby introducing controlled anomalies. This setup enables a systematic assessment of the method's ability to detect changes in the underlying data-generating process.

\paragraph{Approach 1: Covariance Structure Modulation}
The first approach introduces anomalies by modifying the covariance structure of the data while keeping the mean unchanged. This is achieved using Algorithm~\ref{alg:mgaussian_generation}, which generates a multivariate Gaussian distribution with $d$ features and $m$ samples. In this method, varying the seed value -- sampled from a uniform random distribution -- allows for the easy generation of diverse covariance structures. In Algorithm~\ref{alg:mgaussian_generation}, the matrix ${A} \in \mathbb{R}^{d \times d}$ is randomly generated, ensuring that the resulting covariance matrix ${\Sigma}$ remains positive semi-definite. By switching the random seed at specific time steps, a new covariance matrix is introduced, causing abrupt changes in feature dependencies. These transitions act as anomalies, resembling real-world phenomena such as sensor failures or concept drift, where the data distribution shifts unexpectedly.

\paragraph{Approach 2: Correlation Strength Modulation}
The second approach introduces anomalies by modulating the correlation strength between features while maintaining a consistent covariance structure. This is achieved using Algorithm~\ref{alg:correlation_strength}, which generates a multivariate Gaussian dataset with $d$ features and $m$ samples. The correlation strength parameter $s$ controls the dependency among features. By modifying $s$ at specific time points while keeping the seed value fixed, abrupt shifts in feature correlation strengths are induced. These shifts serve as anomalies, mimicking real-world scenarios such as environmental changes or system drifts, where the interdependencies between features change over time.

\begin{algorithm}[t]
\begin{scriptsize}
\caption{Generation of multivariate Gaussian data with a random covariance structure}
\label{alg:mgaussian_generation}
\begin{algorithmic}[1]
\State \textbf{Input:} Number of features \(d\), number of samples \(m\)
\State \textbf{Output:} A dataset \({X} \in \mathbb{R}^{m \times d}\) sampled from a multivariate Gaussian distribution
\State Draw \(S\) from a uniform distribution: \(S \sim U(0, M)\), where \(M\) is the maximum possible seed value
\State Set the mean vector: \({\mu} = {0} \in \mathbb{R}^{d}\)
\State Generate a random matrix: \({A} \in \mathbb{R}^{d \times d} \sim \mathcal{N}(0,1)\)
\State Compute the covariance matrix: \({\Sigma} = {A} {A}^T\) 
\State Sample \(m\) points  
${x}^{(i)} \sim \mathcal{N}({\mu}, {\Sigma})$
\end{algorithmic}
\end{scriptsize}
\end{algorithm}

\begin{algorithm}[t]
\begin{scriptsize}
\caption{Generation of correlated multivariate data}
\label{alg:correlation_strength}
\begin{algorithmic}[1]
\State \textbf{Input:} Number of features \(d\), number of samples \(m\), correlation strength \(s\)
\State \textbf{Ouput:} A dataset \({X} \in \mathbb{R}^{m \times d}\) with controllable correlation strength
\State Set the random seed: \(S = \text{seed}\), where \(S\) controls the pseudo-random number generator
\State Set the mean vector: \({\mu} = {0} \in \mathbb{R}^{d}\)
\State Construct the covariance matrix: \({\Sigma} = (1 - s) {I}_d + s {1}_d {1}_d^T\)
\State Sample \(m\) points  
${x}^{(i)} \sim \mathcal{N}({\mu}, {\Sigma})$
\State Normalize the dataset: \({x^{(i)}} = \frac{{x^{(i)}} - \hat{\mu}}{\hat{\sigma}}\) with $\hat{\mu}$ and $\hat{\sigma}$ the empirical mean and standard deviation
\end{algorithmic}
\end{scriptsize}
\end{algorithm}

Both approaches generate controlled anomalies by switching between distinct statistical properties of the data. The first approach focuses on \textit{abrupt structural changes} in covariance, while the second approach targets \textit{gradual or sudden shifts in correlation strength}. Together, these approaches provide a robust synthetic benchmark for evaluating real-time anomaly detection methods.

We generate two distinct clusters. The first cluster, with two features ($d=2$), allows for a clear visualization of scattered data points. The second cluster, with eight features ($d=8$), represents a higher-dimensional multivariate time series. We create 11 randomly designed scenarios for the first cluster and 8 for the second, applying both Approach 1 (Covariance Structure Modulation) or Approach 2 (Correlation Strength Modulation) in a random manner. Additionally, a separate scenario involving only a mean shift is included in the first cluster to isolate and evaluate the method’s sensitivity to changes in the mean. Each dataset consists of $m=50,000$ samples drawn from a single distribution, with a randomly selected window containing either 100 or 1000 samples replaced by data generated from the alternate distribution, following the methodology described previously. This transition from a background pattern to a temporal pattern alters the interrelation of the data, which is interpreted as abnormal behavior. We demonstrate the first cluster by scatter plot, and the second cluster by a plot consist of covariance matrix values over time as well as a value-time step plot for both clusters. To ensure diversity and eliminate manual selection bias, the data for each scenario is generated by iteratively running a synthetic data generator, where each iteration is seeded with random integers drawn from a uniform distribution. This guarantees that the underlying data distributions vary across scenarios without human intervention. 

\begin{figure*}[t]
    \centering
    \begin{tabular}{cccc}
        \begin{minipage}[c]{0.20\textwidth}
            \centering
            \subfloat[$L_4$]{\includegraphics[width=\linewidth]{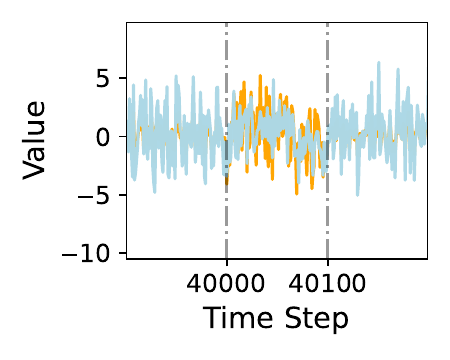}}\\
            \includegraphics[width=\linewidth]{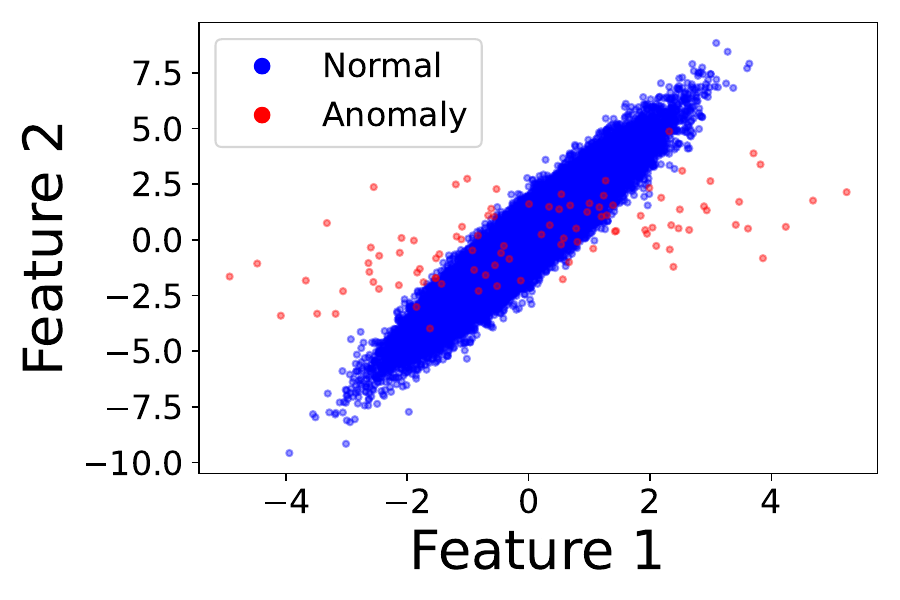}
        \end{minipage}
        &
        \begin{minipage}[c]{0.20\textwidth}
            \centering
            \subfloat[$L_7$]{\includegraphics[width=\linewidth]{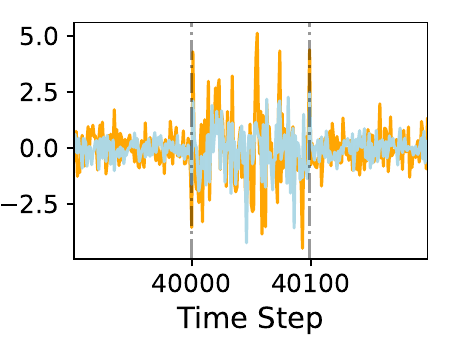}}\\
            \includegraphics[width=\linewidth]{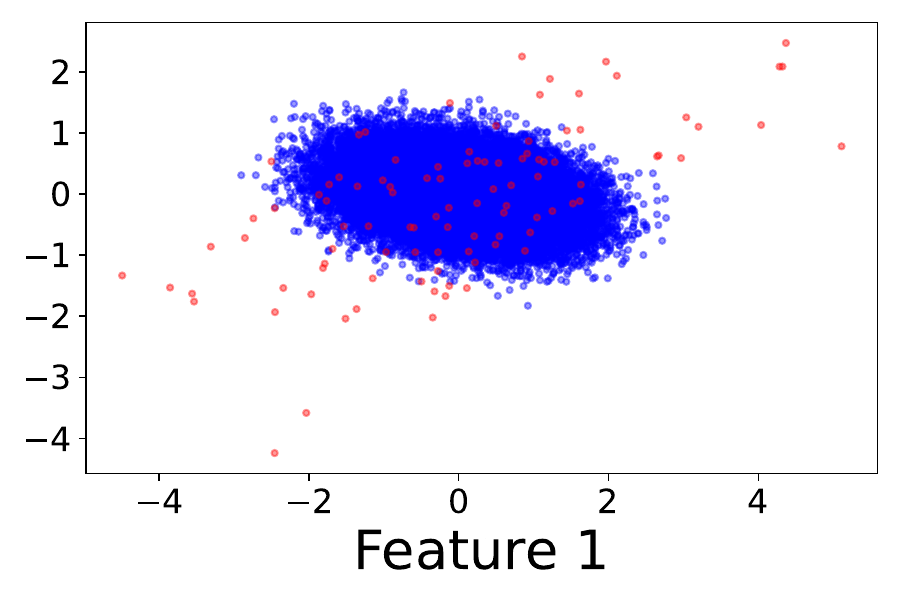}
        \end{minipage}
        &
        \begin{minipage}[c]{0.20\textwidth}
            \centering
            \subfloat[$L_9$]{\includegraphics[width=\linewidth]{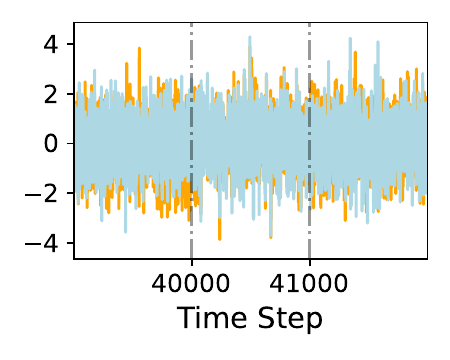}}\\
            \includegraphics[width=\linewidth]{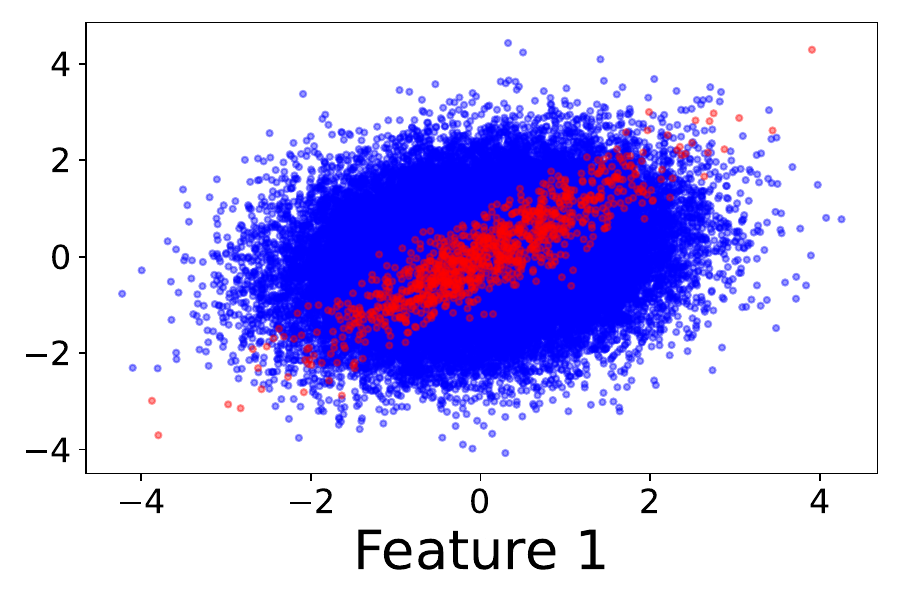}
        \end{minipage}
        &
        \begin{minipage}[c]{0.20\textwidth}
            \centering
            \subfloat[$L_{11}$]{\includegraphics[width=\linewidth]{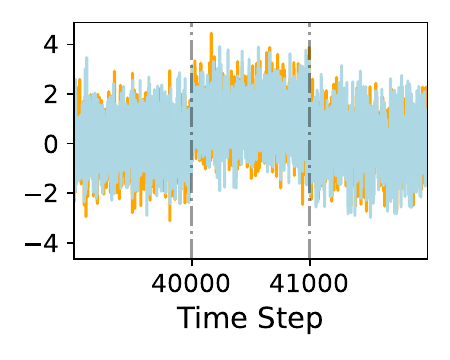}}\\
            \includegraphics[width=\linewidth]{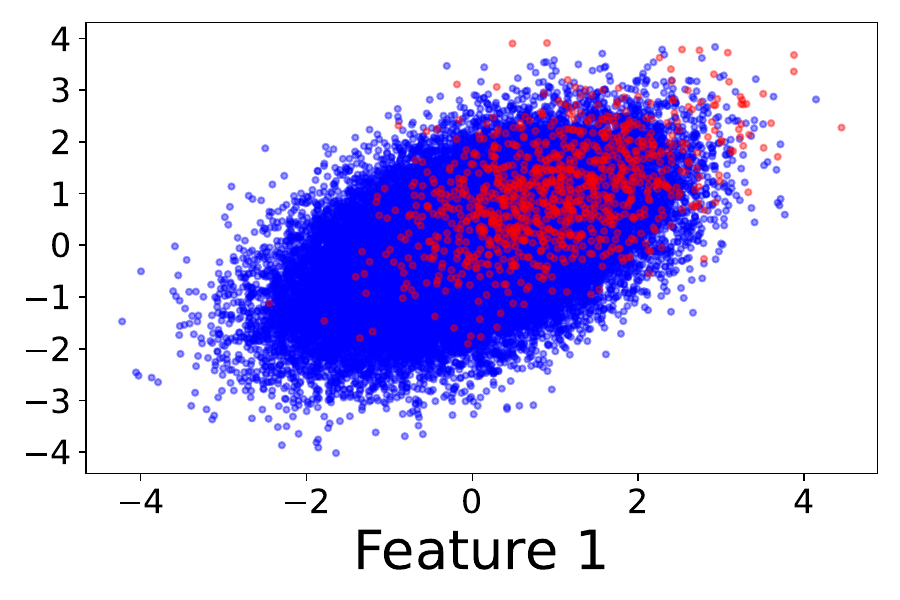}
        \end{minipage}
    \end{tabular}
    \caption{Some instances of randomly generated synthetic data with feature size $d=2$. From top to bottom: value versus time step plot and scatter plot visualizing the distribution of data points.}
    \label{fig:fig1}
\end{figure*}

Figure~\ref{fig:fig1} depicts four instances of the first cluster to provide an insightful over view of interpreted normal and abnormal samples at a glance. The notation $L_i$ refers to a low-dimensional synthetic dataset instance with $d=2$ features, where $i$ denotes the index of the specific generated scenario. Similarly, $H_i$ is used to indicate a high-dimensional synthetic dataset instance with $d=8$ features. As shown in the figure, $L_4$ exhibits similar mean and correlation levels in both the normal and abnormal distributions; however, a noticeable shift in the orientation of the scattered points is observed. In contrast, $L_7$ shows changes in both the magnitude of values and the orientation, while $L_{11}$ presents only a drift in the mean of the data as an abnormal behavior. A particularly distinct case is $L_9$, which represents a scenario where feature correlations remain weak initially but abruptly increase -- making the correlation level itself the only changing factor.

\begin{figure*}[t]
    \centering
    \begin{tabular}{cccc}
        \begin{minipage}[c]{0.20\textwidth}
            \centering
            \subfloat[$H_2$]{\includegraphics[width=\linewidth]{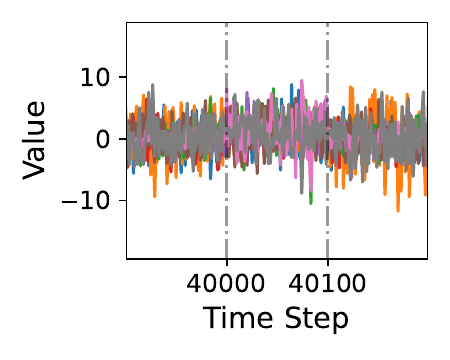}}\\
            \includegraphics[width=\linewidth]{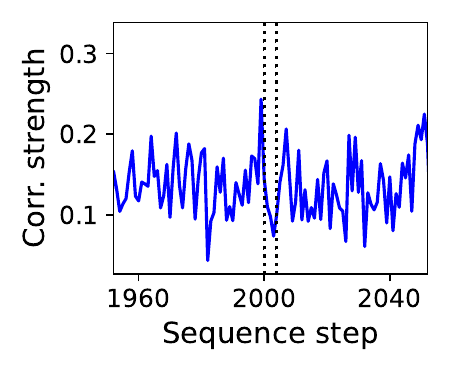}\\
            \includegraphics[width=0.9\linewidth]{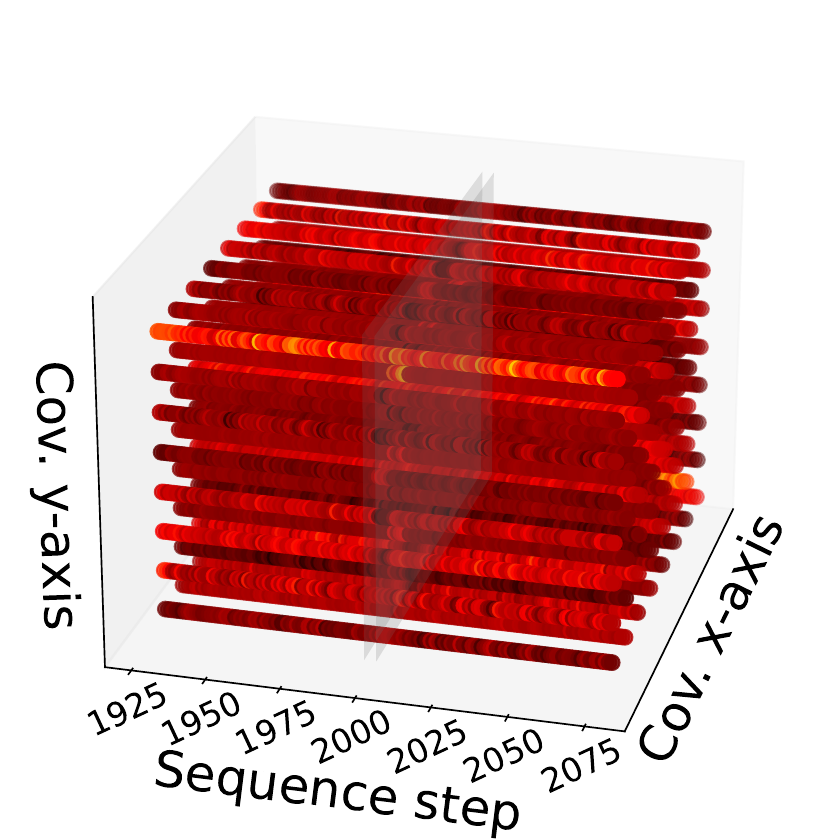}
        \end{minipage}
        &
        \begin{minipage}[c]{0.20\textwidth}
            \centering
            \subfloat[$H_3$]{\includegraphics[width=\linewidth]{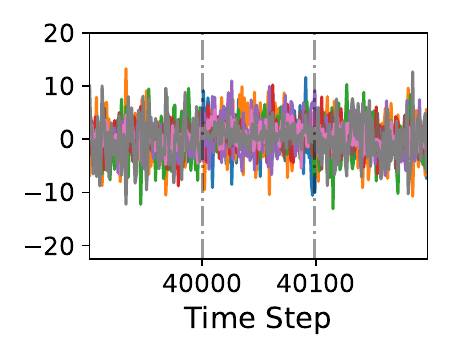}}\\
            \includegraphics[width=\linewidth]{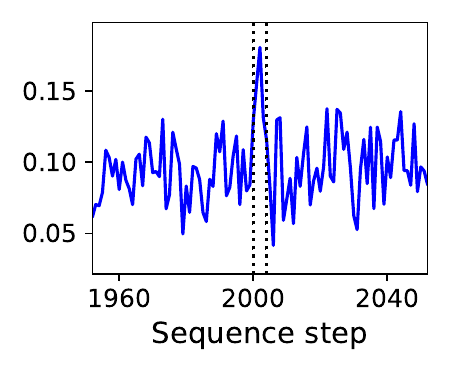}\\
            \includegraphics[width=0.9\linewidth]{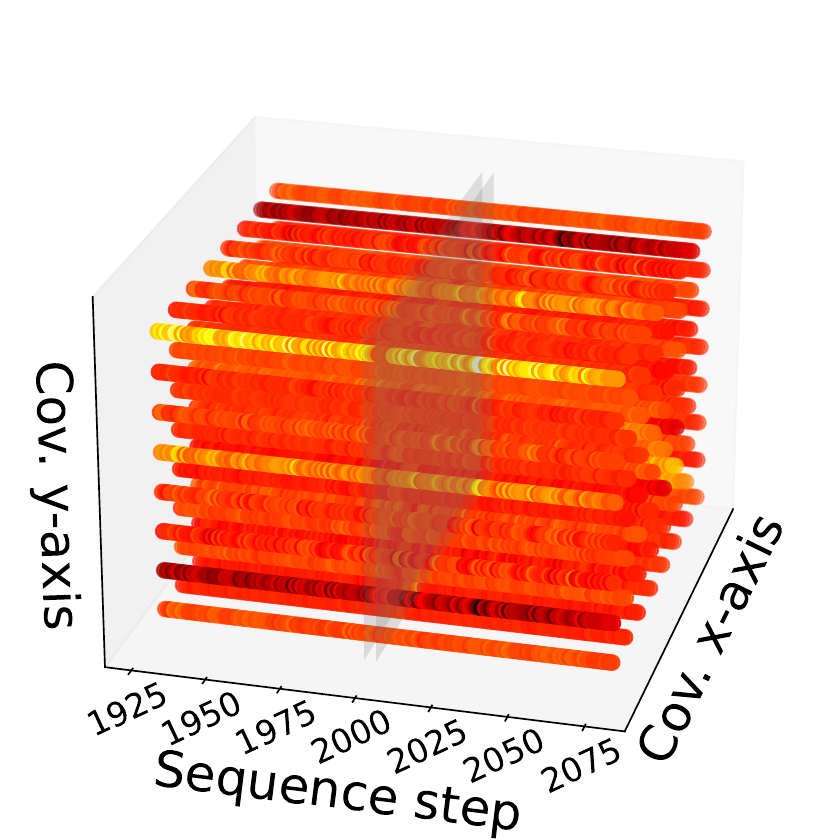}
        \end{minipage}
        &
        \begin{minipage}[c]{0.20\textwidth}
            \centering
            \subfloat[$H_5$]{\includegraphics[width=\linewidth]{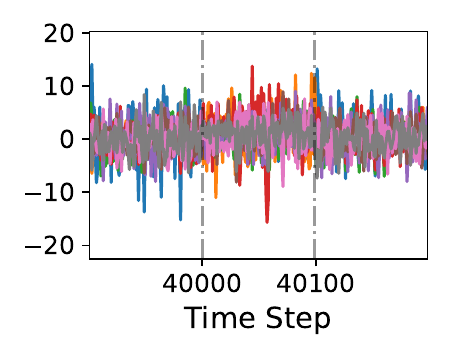}}\\
            \includegraphics[width=\linewidth]{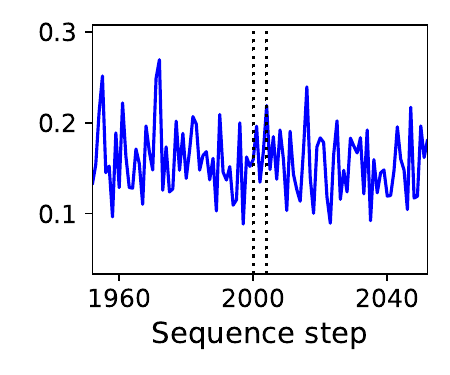}\\
            \includegraphics[width=0.9\linewidth]{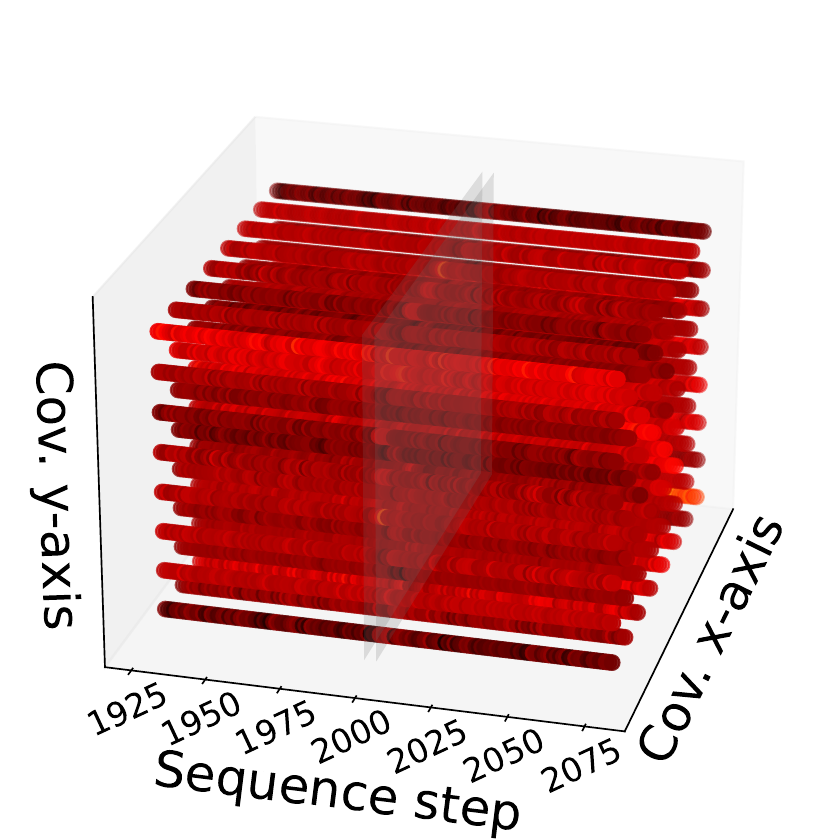}
        \end{minipage}
        &
        \begin{minipage}[c]{0.20\textwidth}
            \centering
            \subfloat[$H_8$]{\includegraphics[width=\linewidth]{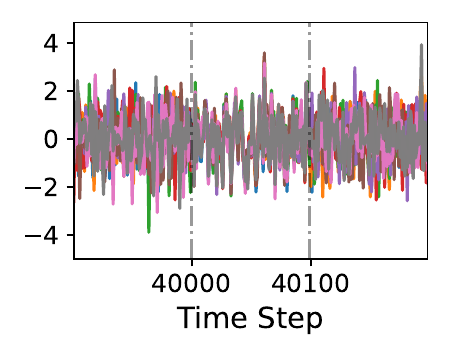}}\\
            \includegraphics[width=\linewidth]{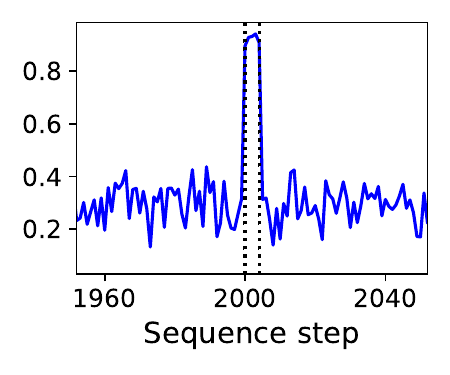}\\
            \includegraphics[width=0.98\linewidth]{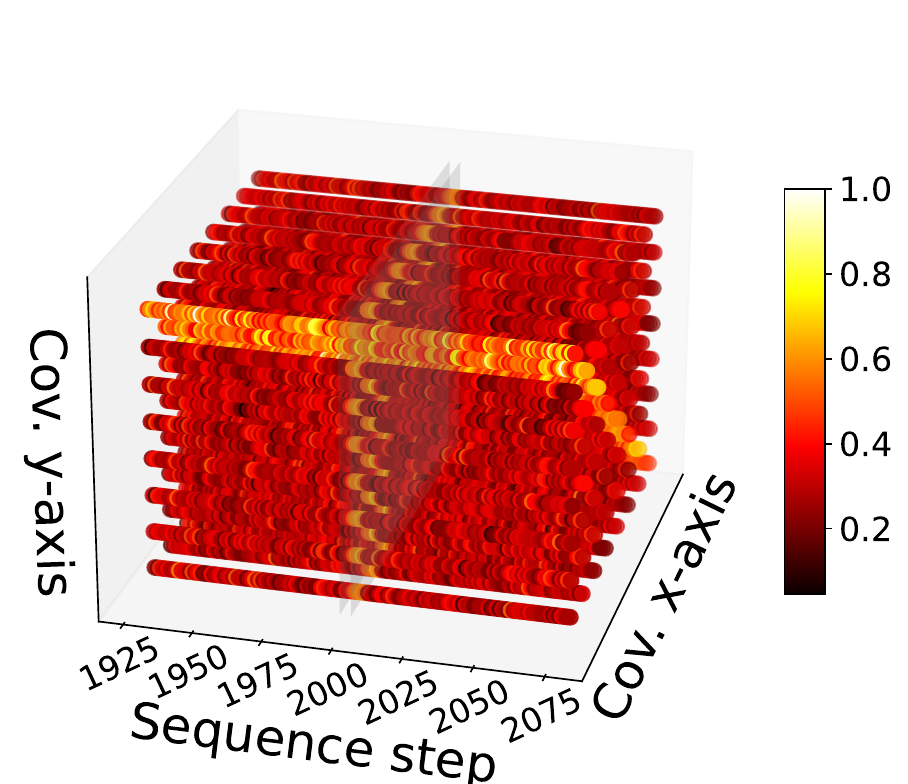}
        \end{minipage}
    \end{tabular}
    \caption{Some instances of randomly generated synthetic data with feature size $d=8$. From top to bottom: value versus time step plot, correlation strength versus sequence steps, and covariance versus sequence steps.}
    \label{fig:fig2}
\end{figure*}

To better visualize abnormal behavior in the higher-dimensional scenario of cluster 2 and visualize any changes in the covariance structure at specified time steps, a non-overlapping sliding covariance computation with a window size of $w_c = 20$ samples is applied. Specifically, the covariance for the $i$th sequence is computed as $\operatorname{Cov}_{\text{sequence } i} = [{X_{i:i+w_c}^T \, X_{i:i+w_c}}]/[{w_c - 1}]$ with respect to the fact that the mean is zero for all samples. Using the computed covariance matrices for each sequence, we visualize the covariance structure of the synthetic datasets through a 3D plot. In this representation, the covariance matrix values are mapped onto the $x$- and $z$-axes as points, while the heatmap colors indicate their magnitudes. The $y$-axis represents the sequence step, allowing us to observe the temporal evolution of covariance patterns. Additionally, we calculate and visualize the mean of the correlation matrix for each sequence step, following the same approach used for the covariance matrices. This helps in understanding the evolution of correlation strength over time, offering a more comprehensive view of how the interrelations between features change throughout the sequences. Figure~\ref{fig:fig2} illustrates the value-time, correlation strength, and the 3D covariance plots for the second cluster. Among the eight generated synthetic datasets in this cluster, we selected four instances that exhibit the most distinguishable normal and abnormal patterns for clearer visualization. The 3D plot includes two gray layers highlighting the specified sequence window where abnormal behavior occurs. Each sequence step represents the covariance pattern computed over 20 samples.

\subsubsection{Alignment of Synthetic Data with Real-World Scenarios}
\label{sec:Alignment}
Generally, anomalies in multivariate time series are defined as samples that deviate from a consistent, normal temporal pattern. In this context, the normal pattern is characterized by a set of samples that exhibit stable and predictable statistical behavior over time. Accordingly, our synthetic data generation approach first establishes a baseline normal pattern for the majority of samples, and then introduces anomalies at specific time steps by altering key statistical properties -- namely, the correlation between features, shifts or changes in the covariance structure, and variations in the mean. These induced deviations are designed to closely mimic the types of anomalies observed in real-world applications.

Covariance shift, also known as covariate shift -- as exemplified by $L_4$ in Figure~\ref{fig:fig1} -- is a prevalent phenomenon where the data distribution evolves over time. A broad range of real-world systems face this challenge of input data shift, requiring continuous monitoring of process behavior and prompt adaptive corrections~\citep{raza2015ewma}. This issue is evident in numerous applications, including spam filtering, brain–computer interfaces (BCIs)~\citep{app132312800}, and EEG prediction\citep{NEURIPS2022_8511d06d}.

In addition, real-time anomaly detection in IIoT applications plays a crucial role. For example, in smart factories, industrial equipment functions as interconnected intelligent nodes. When machines equipped with sensors exhibit anomalous behavior, they can trigger disruptions in production, leading to significant economic losses~\citep{9915308}. In IIoT settings, sensors embedded in machinery and production lines generally maintain stable covariance relationships during normal operations. However, an abrupt change in the covariance structure -- for instance, a sudden alteration in the joint behavior of temperature and vibration sensors -- may indicate issues such as mechanical misalignment or process disturbances. Similarly, IIoT environments rely on consistent correlations among sensor readings. Under typical conditions, sensors monitoring parameters like pressure, flow, and energy consumption display predictable correlation strengths. A sudden drop or spike in these correlations -- for example, if a normally strong correlation between pressure and flow is lost, or if two independent sensor readings become unexpectedly synchronized -- can signal operational anomalies such as blockages, leaks, or coupling faults.

From an observational standpoint, we leverage the widely used IIoT benchmark dataset, DAMADICS (Development and Application of Methods of the Actuator Diagnosis in Industrial Control Systems)~\citep{refDamadics}, to underscore the similarity between our synthetic datasets and real-world data. We employ visualization techniques for high-dimensional multivariate time series -- similar to those shown in Figure~\ref{fig:fig2} -- along with an analysis of the mean correlation matrix, to demonstrate how the average correlation strength between features evolves over time.

\begin{figure*}[t]
    \centering
    \begin{tabular}{cccc}
        \begin{minipage}[c]{0.20\textwidth}
            \centering
            \subfloat[$DAM_{0}, w_{c} = 100$]{\includegraphics[width=\linewidth]{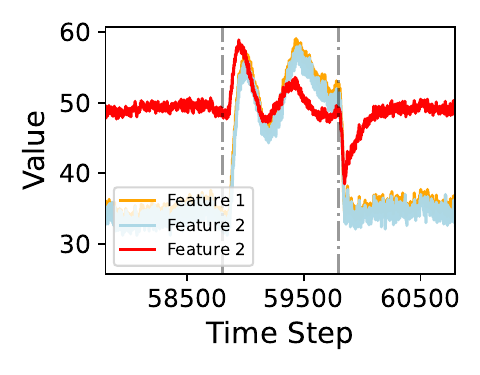}}\\
            \includegraphics[width=\linewidth]{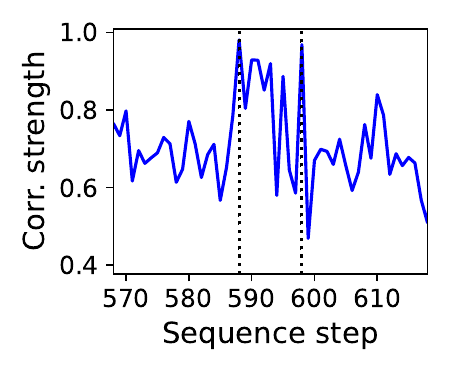}\\
            \includegraphics[width=0.9\linewidth]{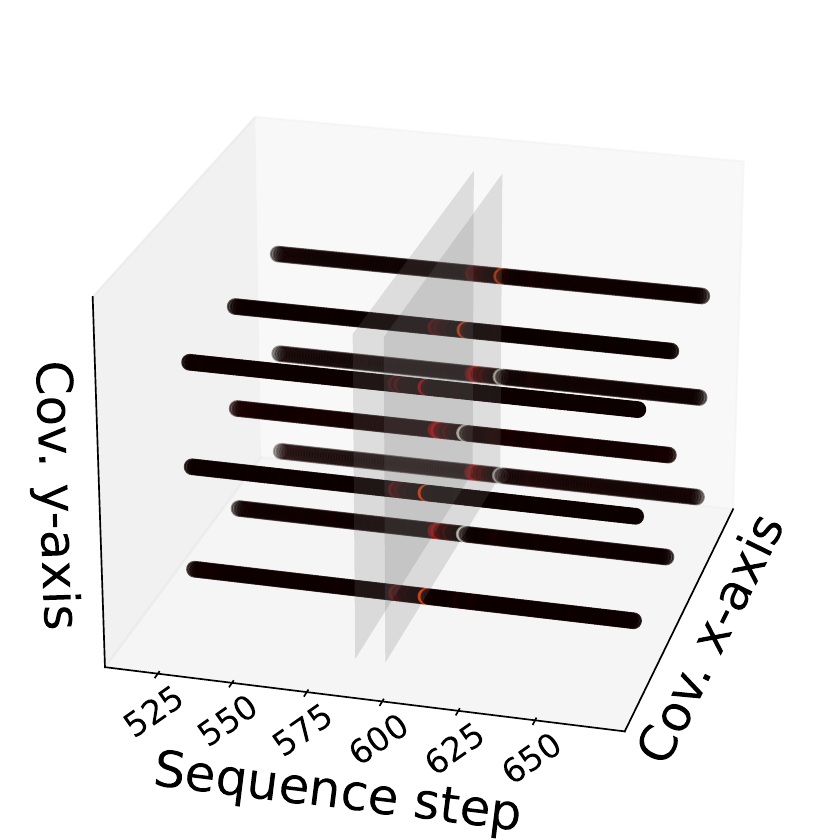}
        \end{minipage}
        &
        \begin{minipage}[c]{0.20\textwidth}
            \centering
            \subfloat[$DAM_{2}, w_{c} = 200$]{\includegraphics[width=\linewidth]{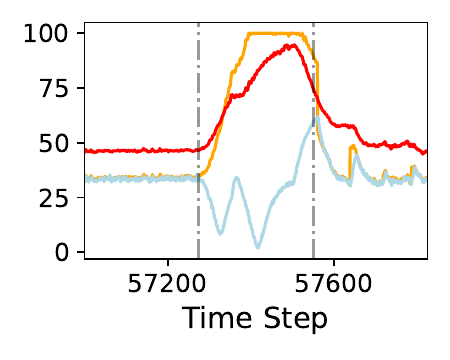}}\\
            \includegraphics[width=\linewidth]{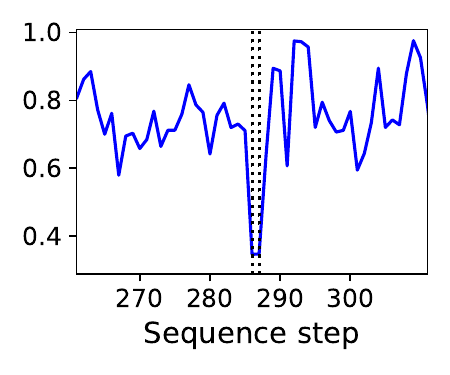}\\
            \includegraphics[width=0.9\linewidth]{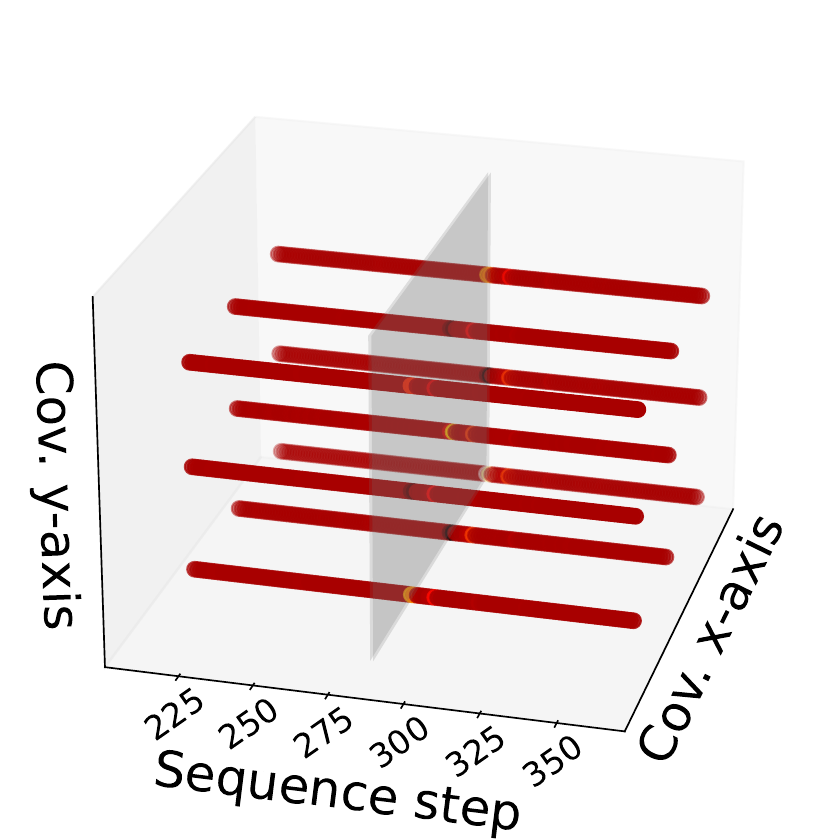}
        \end{minipage}
        &
        \begin{minipage}[c]{0.20\textwidth}
            \centering
            \subfloat[$L_9,w_{c}=20$]{\includegraphics[width=\linewidth]{figures/Toy9.pdf}}\\
            \includegraphics[width=\linewidth]{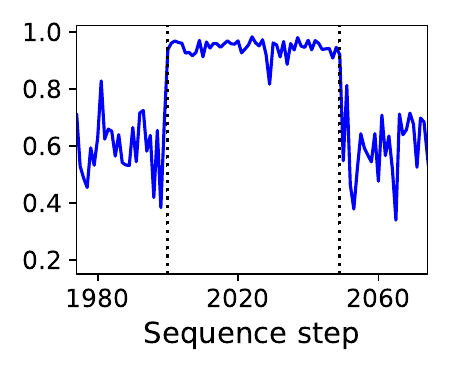}\\
            \includegraphics[width=0.9\linewidth]{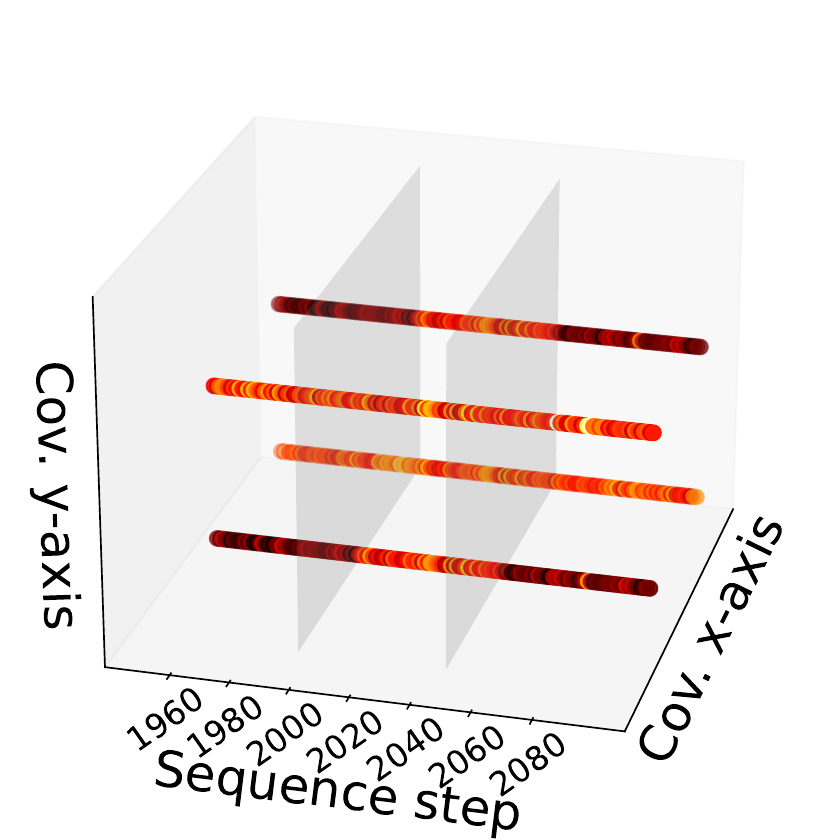}
        \end{minipage}
        &
        \begin{minipage}[c]{0.20\textwidth}
            \centering
            \subfloat[$L_{10}, w_{c}=20$]{\includegraphics[width=\linewidth]{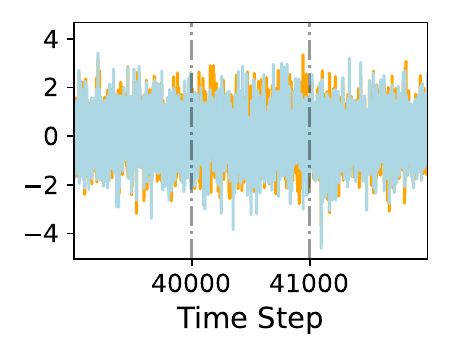}}\\
            \includegraphics[width=\linewidth]{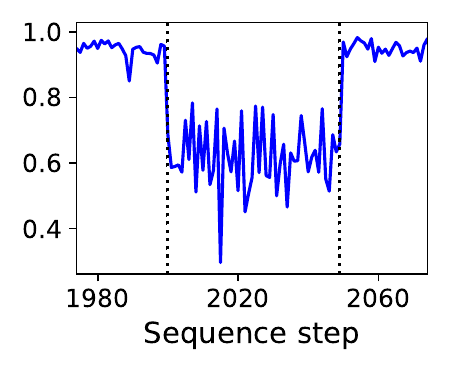}\\
            \includegraphics[width=0.98\linewidth]{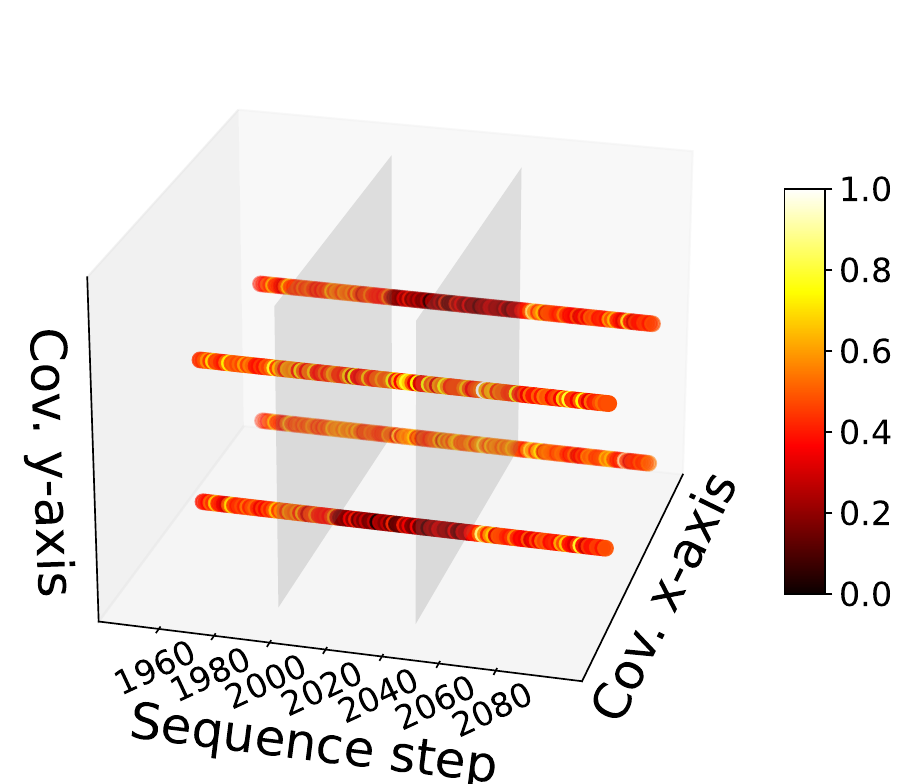}
        \end{minipage}
    \end{tabular}
    \caption{Visual comparison between two real-world data scenarios from DAMADICS dataset and the two generated synthetic datasets. From top to bottom: value versus time step plot, correlation strength across sequence steps, and covariance across sequence steps.}
    \label{fig:fig6}
\end{figure*}

Figure~\ref{fig:fig6} illustrates two fault instances from the DAMADICS dataset (where $DAM_{i}$ denotes the $i$-th scenario from the dataset) alongside two examples of our synthetic data. As observed, $DAM_{0}$ shows a transition from a weak correlation between features in the normal pattern to a higher correlation strength in the abnormal sequences, similar to the behavior seen in $L_9$ (and also in $H_8$ depicted in Figure~\ref{fig:fig2}). In the time plot, all three features increase in value and develop a positive correlation. Conversely, $DAM_{2}$ exemplifies a transition from a high to a lower correlation strength, reflecting the behavior observed in $L_{10}$. Although anomalies in the DAMADICS dataset manifest as changes in the data range -- creating a relatively easier detection problem -- our synthetic data generation approach deliberately avoids introducing deviations in the overall data range. This design choice aligns with the concept of contextual anomalies, where observations may remain within the normal global range yet deviate from the expected pattern when examined in context.

Altogether, the incorporation of shifts in mean, covariance, and correlation strength in our synthetic datasets mirrors a wide range of real-world anomaly scenarios. This alignment not only reinforces the realism of our synthetic data generation approach but also demonstrates its applicability to practical, real-time systems. Consequently, it supports our argument for its alignment with real-world scenarios. 

\subsubsection{Benchmark Setup}
\label{sec:synth_bench_setup}
In unsupervised anomaly detection, evaluating anomaly scores is generally more practical and insightful than using binary labels, as noted by~\citep{bouman2024unsupervised}. An anomaly score is a continuous measure where higher values indicate a stronger probability that a given sample is anomalous, based on the method’s assessment. This scoring approach enables the ranking of samples according to their degree of  anomaly intensity, offering deeper understanding of the anomalies’ characteristics. For a given dataset, anomaly scores are calculated across all data points at once, without the need for cross-validation or train-test splits typically employed in supervised methods. These unsupervised scores are subsequently evaluated against ground truth labels -- where a sample is marked as an anomaly (1) or not (0) -- to assess their effectiveness. To assess the performance of our DAD method, we compare our results against state-of-the-art (SOTA) unsupervised anomaly detection methods for time series data. The evaluation is based on the area under the receiver operating characteristic (ROC) curve (AUC). 
This is a widely recognized metric in anomaly detection assessments~\citep{goldstein2016comparative, campos2016evaluation, xu2018comparison}.

We leverage a comprehensive framework for evaluating unsupervised anomaly detection methods, introduced in~\citep{bouman2024unsupervised}, which includes 33 methods -- 27 of which are sourced from the widely used Python library for anomaly detection, PyOD~\citep{zhao2019pyod}, while the remaining methods were implemented independently. We integrate our twenty randomly generated synthetic datasets into the framework and implement our method following their standardized template for incorporating custom methods. As each method has its own hyperparameters, the authors provided "a set of sensible hyperparameters per method" for their benchmarking. We will similarly define our hyperparameter list (see Table~\ref{tab:DAD_hyperparam}) and integrate our method into their framework to complement the evaluation materials. 

We evaluate DAD under two configurations: (1) a compact configuration without windowing, defined by $p = 0$, referred to as DAD -- offering a smaller hyperparameter search space; and (2) a more flexible configuration that allows variation in spatio-temporal size with $p \in [0, 1]$, referred to as DAD\_s.

\begin{table*}[t]
\centering
\caption{Recommended hyperparameters for the proposed DAD method.}
\label{tab:DAD_hyperparam}
\begin{small}\begin{tabular}{l|c}
\textbf{Hyperparameter} & \textbf{Values} \\
\hline
Learning rate ($\eta$) & ${0.8, 0.2, 0.08, 0.02, \dots, 2 \times 10^{-6}}$ \\
Momentum factor ($\gamma$) & $0.25$ \\
Sliding window size ($p$) & 0, 1 \\
\end{tabular}\end{small}
\end{table*}

Among the methods available in the evaluation framework, several are based on neural networks, which we exclude from this study to maintain a fair comparison focused on statistics-based approaches. This decision is supported  by~\citep{bouman2024unsupervised}, who observed that "many of the neural networks do not perform well" in this context. Since evaluating all methods available in the framework requires a significant amount of time, we selected a subset of widely used SOTA methods for benchmarking based on synthetic datasets. The selected methods include MCD, GMM, $k$-NN, $k$th-NN, INNE, EIF, IF, OCSVM, gen2out, PCA, DHBOS (DynamicHBOS, referred to as DHBOS throughout this work), ECOD, and COPOD.
To highlight the potential of these methods in handling synthetic data scenarios, we report the maximum performance achieved for each method-hyperparameter-dataset combination. 

\subsection{Empirical Validation using Real-World Data}
\label{sec:benchmark_setup}

\begin{figure}[!ht]
    \centering
    \includegraphics[width=0.5\textwidth]{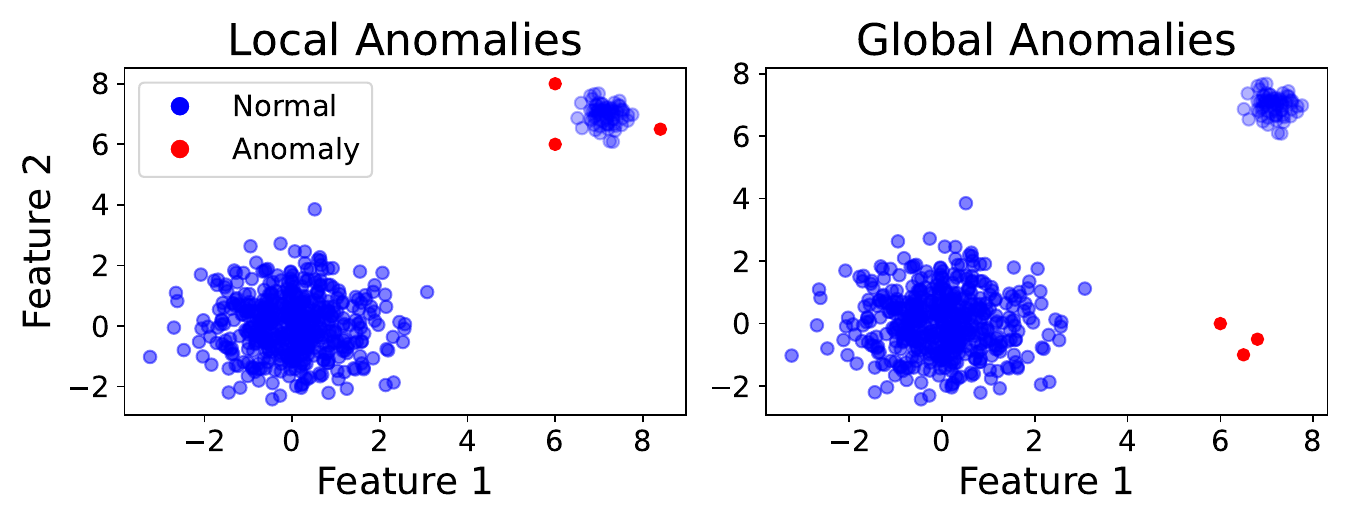}      
    \caption{Illustration of local anomalies (left) and global anomalies (right).}
    \label{fig:local_global_anomalies}
\end{figure}
To evaluate the performance of our proposed anomaly detection method, DAD, we also benchmark it on real-world data collected in the benchmarking framework by~\cite{bouman2024unsupervised}. This framework comprises 50 tabular datasets, offering a broad range of scenarios for evaluation. Accordingly, two distinct clusters were identified: \textit{local anomaly} datasets and \textit{global anomaly} datasets. Local anomalies are characterized by occupying regions of lower density relative to their neighboring samples, whereas global anomalies exist in generally low-density regions within the feature space (Figure~\ref{fig:local_global_anomalies}). Likewise, we present performance results separately for both global and local anomaly types. Local anomaly datasets include \textit{aloi, fault, glass, internetads, ionosphere, landsat, letter, magic.gamma, nasa, parkinson, pen-local, pima, skin, speech, vowels, waveform, and wilt}, while global anomaly datasets contain the remaining datasets, excluding \textit{vertebral}. For a detailed list of real-world datasets in the benchmark, see Table~\ref{tab:HPT_based_L_Exploration}.

\subsubsection{Benchmark Setup}
Since our evaluation follows the practical hyperparameter tuning strategy introduced in Section~\ref{sec:hyper_opt}, we selected only those SOTA methods from the framework that include both hyperparameters and their corresponding value ranges for real-world data benchmarking (see Table~\ref{tab:methods_hyper}). We applied Algorithm~\ref{alg:find_L} across all real-world datasets within the benchmark to determine the most suitable subset per dataset for hyperparameter tuning. Details regarding the identified downsampling ratios and subset sizes for each dataset are provided in Appendix Table~\ref{tab:HPT_based_L_Exploration}. Next, we used Eq.~\eqref{optimal_hyperparameter} to identify the optimal hyperparameter for each method-dataset combination based on the selected subset, where the quality criterion $q$ is AUC. Finally, AUC is computed for each method using the full dataset. As in the synthetic benchmark setup, both DAD and DAD\_s configurations are evaluated on the real-world benchmark datasets. 

To benchmark performance on real-world datasets (detailed in Section~\ref{sec:benchmark_setup}), we follow the methodology of the benchmark framework for reporting results. Specifically, we report the performance distribution of each method across datasets in terms of the maximum AUC, defined as:
\begin{equation}
\label{eq:auc}
\overline{\text{AUC}}(M, D) = \frac{\text{AUC}(M, D)}{\max_{M' \in M} \text{AUC}(M', D)} \times 100
\end{equation}
where $M$ is one of the methods and $D$ is one of the datasets. 

To ensure reproducibility, we build upon the previously introduced benchmarking framework, which we have extended and modified to incorporate our proposed hyperparameter tuning-based evaluation strategy. In addition, we introduce new benchmarking settings that assess performance under peak, mean, and default hyperparameter configurations (detailed in Appendix~\ref{sec:app_traditional}). Our method, DAD, along with its automated variant, has been integrated into the benchmarking suite. Furthermore, we have added support for the proposed synthetic dataset and the industrial IIoT dataset, DAMADICS, expanding the applicability of the benchmark. The complete modified benchmarking package is publicly available at: \url{https://github.com/AmirhosseinSadough/DAD}.

\section{Results}
\label{sec:Results}
We validate our method across two benchmark settings: (1) synthetic datasets introduced in Section~\ref{sec:EmpiricalValidation}, to evaluate the underlying concept, and (2) real-world datasets, as detailed in Section~\ref{sec:benchmark_setup}, to assess performance across diverse, widely-used benchmarks.

\subsection{Synthetic Data}
\label{sec:Synthetic_result}
\begin{figure*}[t]
    \centering
    \begin{tabular}{cc}
        \subfloat[]{\includegraphics[width=0.49\textwidth]{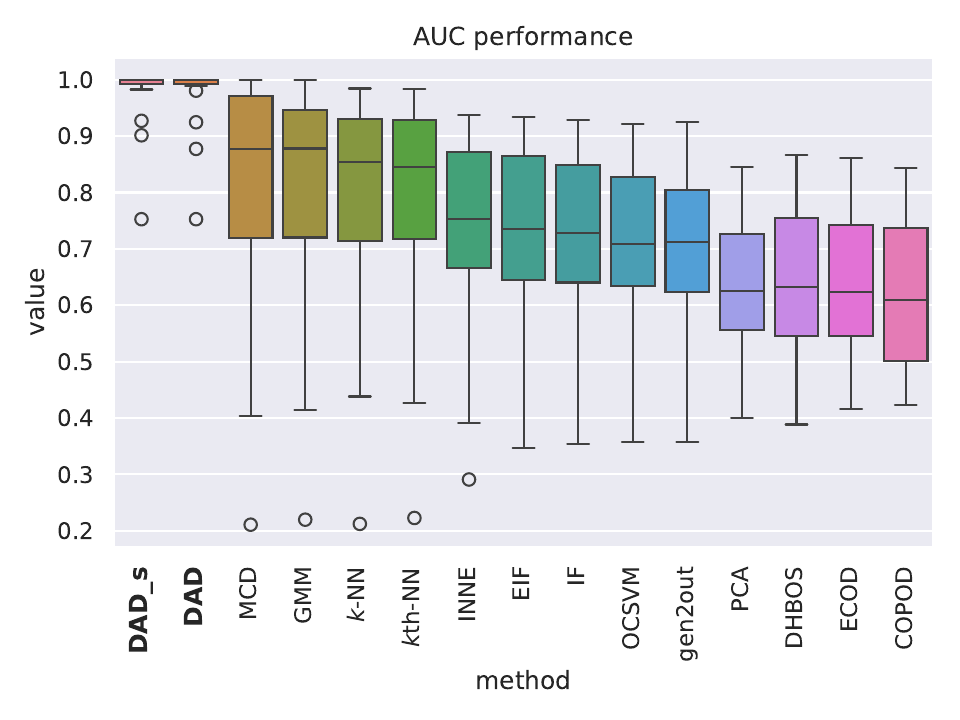}}  &  
        \subfloat[]{\includegraphics[width=0.38\textwidth]{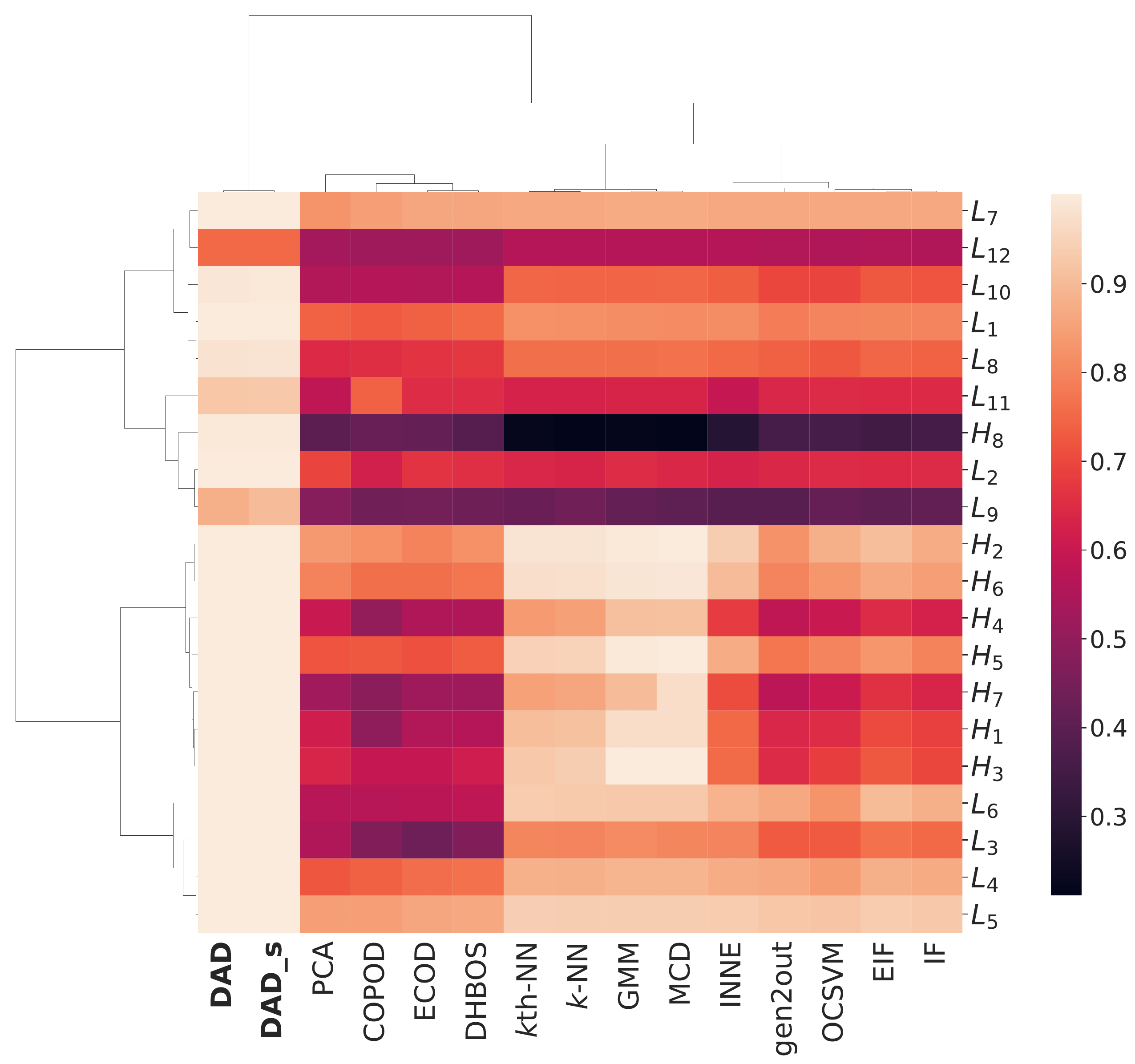}} 
    \end{tabular}
    \caption{(a) Boxplots of method performance on synthetic datasets in terms of AUC. Whiskers extend 1.5 times the interquartile range beyond the quartiles; outliers are marked as circles. (b) Clustered heatmap of AUC performance using hierarchical clustering with average linkage and Pearson correlation.}
    \label{fig:fig3}
\end{figure*}

Figure~\ref{fig:fig3}~(a) shows a box plot illustrating the performance distribution of each method across the synthetic datasets. The results indicate that MCD, GMM, $k$-NN, and $k$th-NN achieve comparable performance and outperform INNE, EIF, IF, OCSVM, and gen2out, which exhibit similar performance levels concerning the median levels. Likewise, PCA, DHBOS, ECOD, and COPOD demonstrate relatively close performance, ranking as the least effective methods. In contrast, our DAD method (both configurations) demonstrates superior anomaly detection capabilities across the synthetic data scenarios. 

Following the methodology outlined in~\citep{bouman2024unsupervised}, we also utilize the Iman-Davenport test to determine if the performance variations between methods are statistically significant. The analysis yielded a test statistic of 9.76, which substantially exceeds the critical value of 0.46, prompting us to reject the null hypothesis. To pinpoint which methods demonstrate superior performance, we subsequently employ the Nemenyi post-hoc test for further analysis.

Table~\ref{tab:outperforms_max} presents the results of the Nemenyi post-hoc analysis conducted on the synthetic dataset, highlighting significant differences between the evaluated methods. 
The results clearly illustrate that several methods significantly outperform others. Notably, both versions of our DAD method excel, surpassing 11/14 methods evaluated. Following DAD, MCD and GMM are the next most effective methods, outperforming 5/14 method. $k$-NN and $k$th-NN rank third, outperforming 4/14 methods. The mean AUC of our DAD method is approximately 0.97, which is about 0.17 higher than the second-best performing method, MCD. This substantial difference underscores the superior performance of our method in effectively addressing these types of problems.

\begin{table*}[t]
\caption{Nemenyi post-hoc analysis results showing significant performance differences between methods on the synthetic dataset, along with mean AUC values.
Symbols indicate the statistical significance of performance differences: $+{}+$ and $+$ denote that the method in the respective row significantly outperforms the method in the respective column at $p=0.05$ and $p=0.10$, respectively. Conversely, $-{}-$ and $-$ indicate that the method in the row is significantly outperformed by the method in the column at $p=0.05$ and $p=0.10$, respectively. Rows and columns are sorted by descending and ascending mean performance, respectively. Notably, columns are omitted when the corresponding method is not significantly outperformed by any other method at $p=0.05$ or $p=0.10$. The final column reports the mean AUC for each method, offering an overview of overall performance. 
}
\label{tab:outperforms_max}
\resizebox{\textwidth}{!}{%
\begin{scriptsize}
\begin{tabular}{lccccccccccccccc|c}
\toprule
 & {COPOD} & {ECOD} & {DHBOS} & {PCA} & {gen2out} & {OCSVM} & {IF} & {EIF} & {INNE} & {$k$th-NN} & {$k$-NN} & {GMM} & {MCD} & {DAD} & {DAD\_s} & {\textbf{Mean AUC}} \\
\midrule
DAD\_s & ++ & ++ & ++ & ++ & ++ & ++ & ++ & ++ & ++ & ++ & ++ &  &  &  &  & \textbf{0.9773} \\
DAD & ++ & ++ & ++ & ++ & ++ & ++ & ++ & ++ & ++ & + & ++ &  &  &  &  & \textbf{0.9758} \\
MCD & ++ & ++ & ++ & ++ & ++ &  &  &  &  &  &  &  &  &  &  & 0.8014 \\
GMM & ++ & ++ & ++ & ++ & ++ &  &  &  &  &  &  &  &  &  &  & 0.7991 \\
$k$-NN & ++ & ++ & + & ++ &  &  &  &  &  &  &  &  &  & -{}- & -{}- & 0.7830 \\
$k$th-NN & ++ & ++ & ++ & ++ &  &  &  &  &  &  &  &  &  & - & -{}- & 0.7825 \\
INNE &  &  &  &  &  &  &  &  &  &  &  &  &  & -{}- & -{}- & 0.7358 \\
EIF &  &  &  &  &  &  &  &  &  &  &  &  &  & -{}- & -{}- & 0.7272 \\
IF &  &  &  &  &  &  &  &  &  &  &  &  &  & -{}- & -{}- & 0.7153 \\
OCSVM &  &  &  &  &  &  &  &  &  &  &  &  &  & -{}- & -{}- & 0.7030 \\
gen2out &  &  &  &  &  &  &  &  &  &  &  & -{}- & -{}- & -{}- & -{}- & 0.6942 \\
PCA &  &  &  &  &  &  &  &  &  & -{}- & -{}- & -{}- & -{}- & -{}- & -{}- & 0.6431 \\
DHBOS &  &  &  &  &  &  &  &  &  & -{}- & - & -{}- & -{}- & -{}- & -{}- & 0.6374 \\
ECOD &  &  &  &  &  &  &  &  &  & -{}- & -{}- & -{}- & -{}- & -{}- & -{}- & 0.6315 \\
COPOD &  &  &  &  &  &  &  &  &  & -{}- & -{}- & -{}- & -{}- & -{}- & -{}- & 0.6269 \\
\bottomrule
\end{tabular}
\end{scriptsize}
}
\end{table*}

To visualize the similarities between methods and datasets, a heatmap is generated to show the performance of each dataset/method combination, accompanied by dendrograms from two hierarchical clusterings: one for the datasets and one for the methods. Figure~\ref{fig:fig3}~(b) demonstrates the clustered heatmap, suggesting that the proposed DAD method consistently handles all scenarios except for $L_{12}$, while the other methods struggle with nearly half of the synthetic datasets.

The two exceptional cases, $L_9$ and $H_8$, highlighted in the heatmap, indicate a specific challenge that none of the evaluated methods -- except our DAD method -- can effectively capture. To better understand these problem scenarios, we further examine the characteristics of these synthetic datasets. As shown in Figure~\ref{fig:fig1}, case $L_9$ presents a situation where the mean remains unchanged between normal and abnormal distributions. However, the key abnormality lies in the correlation strength -- while normal samples exhibit weak dependencies, anomalies display significantly stronger correlations. This problem requires detecting a shift from weak to strong dependency over time. Similarly, Figure~\ref{fig:fig2} illustrates the case of $H_8$, where the covariance matrix for normal sequence steps appears nearly diagonal, indicating low correlation among features. However, in anomalous sequence steps, the covariance matrix deviates from this diagonal structure, revealing a substantial increase in correlation. These findings highlight the ability of our DAD method to capture complex correlation-based anomalies that remain undetected by the other approaches.

Clustering analysis reveals two distinct groups of methods: our DAD method forms its own standalone cluster, while all other methods are grouped into a subcluster. This distinction further highlights the superior performance of DAD in effectively tackling the challenges presented by this problem environment. Similarly, the datasets are clustered into two main classes. The upper cluster consists of datasets where only our DAD method demonstrates strong performance, while none of the other methods exhibit consistent or meaningful results. The lower cluster is further divided into two subgroups:~\textbf{Group A}, comprising $k$-NN, $k$th-NN, MCD, and GMM, performs relatively well, achieving effective results on 7/11 datasets. However, our method remains the only one to maintain strong performance across all 11 datasets in this cluster.~\textbf{Group B}, which includes PCA, COPOD, ECOD, and DHBOS, is the least effective, failing to meet the requirements of this problem environment. While EIF, IF, OCSVM, gen2out, and INNE perform better than the methods in Group B, they still fall short of being viable candidates for effectively handling this task. 

These observations reinforce the distinctive ability of DAD in identifying complex anomaly patterns, such as changes in covariance structure or correlation shifts, where no existing unsupervised multivariate method in the literature demonstrates a viable solution. We used this empirical approach to validate the interrelation learning mechanism proposed by our DAD method. 

\subsubsection{DAD's Decorrelation Matrix Learning Behavior}
\label{sec:Decorrelation_Learning}
\begin{figure*}[t]
    \centering
    \begin{tabular}{cc}
        \subfloat[$L_1, {\eta=2e^{-5}, \gamma=0.25, p=1}$]{\includegraphics[width=0.45\textwidth]{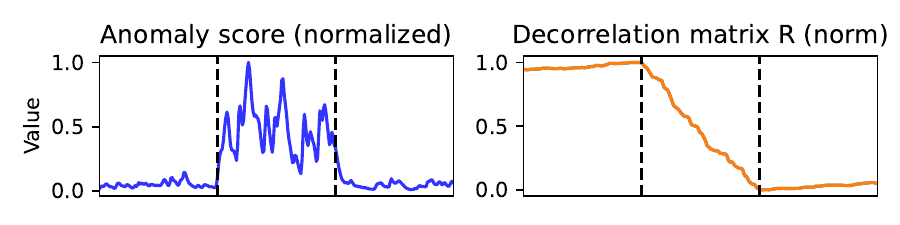}}    &    
        \subfloat[$L_2, {\eta=2e^{-4}, \gamma=0.25, p=1}$]{\includegraphics[width=0.45\textwidth]{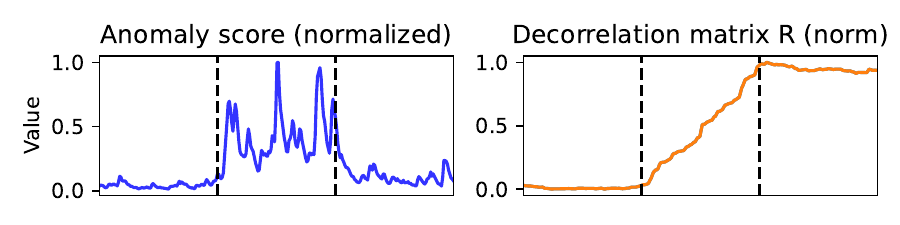}}    \\            
        \subfloat[$L_9, {\eta=2e^{-3}, \gamma=0.25, p=1}$]{\includegraphics[width=0.45\textwidth]{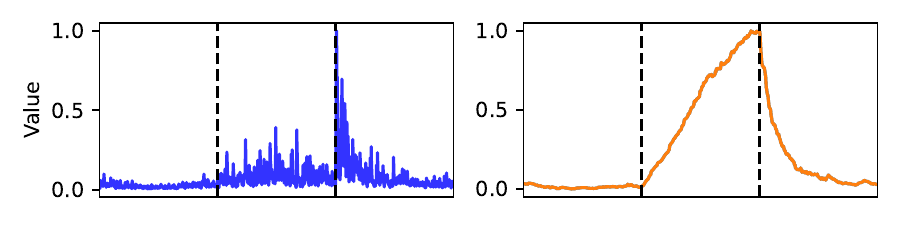}}    &   
        \subfloat[$L_{11}, {\eta=8e^{-6}, \gamma=0.25, p=0}$]{\includegraphics[width=0.45\textwidth]{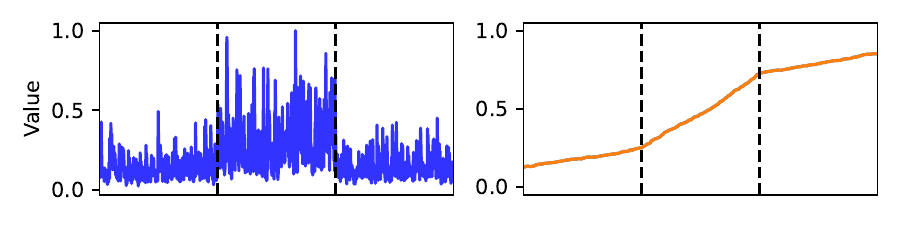}}  \\        
        \subfloat[$H_2, {\eta=2e^{-4}, \gamma=0.25, p=0}$]{\includegraphics[width=0.45\textwidth]{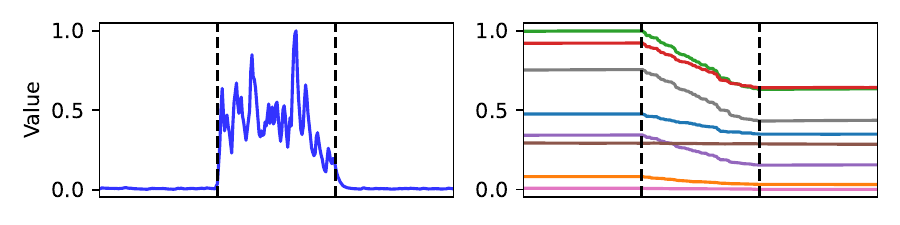}}   &           
        \subfloat[$H_3, {\eta=2e^{-3}, \gamma=0.25, p=0}$]{\includegraphics[width=0.45\textwidth]{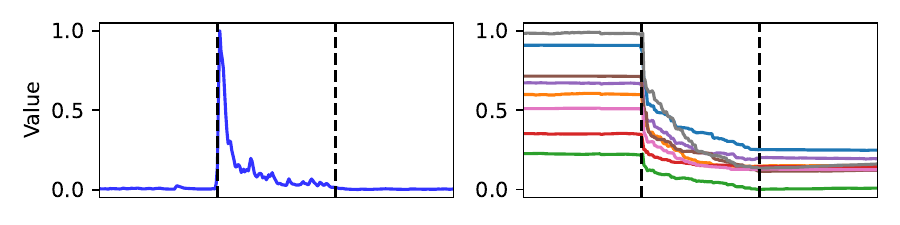}}   \\    
        \subfloat[$H_6, {\eta=2e^{-4}, \gamma=0.25, p=0}$]{\includegraphics[width=0.45\textwidth]{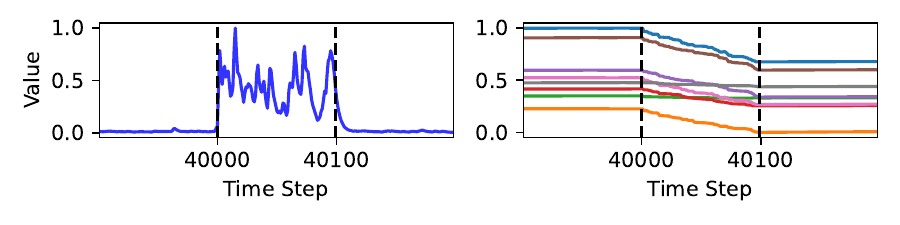}}   &    
        \subfloat[$H_8, {\eta=2e^{-5}, \gamma=0.25, p=0}$]{\includegraphics[width=0.45\textwidth]{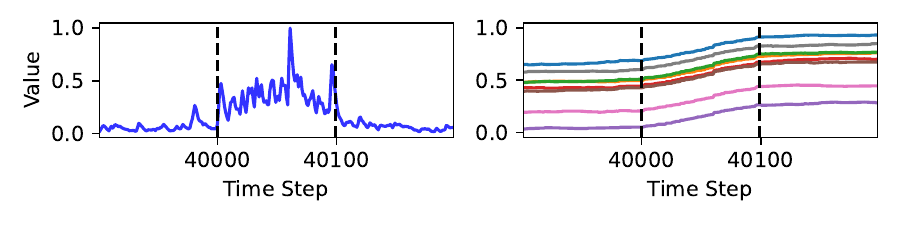}}             
    \end{tabular}
\caption{Visualization of the learning behavior of the decorrelation matrix ($R$), based on the Frobenius norm of $R$ with respect to each feature's update at each time step, along with the corresponding anomaly scores across multiple synthetic data scenarios.}    
\label{fig:fig5}
\end{figure*}

As described earlier, tracking the behavior of the decorrelation matrix $R$ allows us to monitor the learning dynamics of the data, with the magnitude of change in $R$ serving as an effective anomaly score for $x_t$. To provide a visual understanding of this process, Figure~\ref{fig:fig5} illustrates the learning curve of the decorrelation matrix -- measured by the Frobenius norm of $R$ with respect to each feature's update (\textit{i.e.}, the matrix column used to decorrelate the corresponding feature) -- alongside the corresponding anomaly scores across various synthetic data scenarios. As observed, abnormal behavior is consistently reflected by a noticeable change in the orientation of the decorrelation matrix learning curve. This occurs when the method identifies the need for a substantial update, recalculating $R$ to better decorrelate the given input.

\subsection{Real-World Data}
\label{sec:RealWorld_result}
\begin{figure}
    \centering
    \setlength{\tabcolsep}{0.001pt}
    \begin{tabular}{cc}
        \subfloat[DAD\_s vs SOTA on all benchmark datasets]{\includegraphics[width=0.5\textwidth]{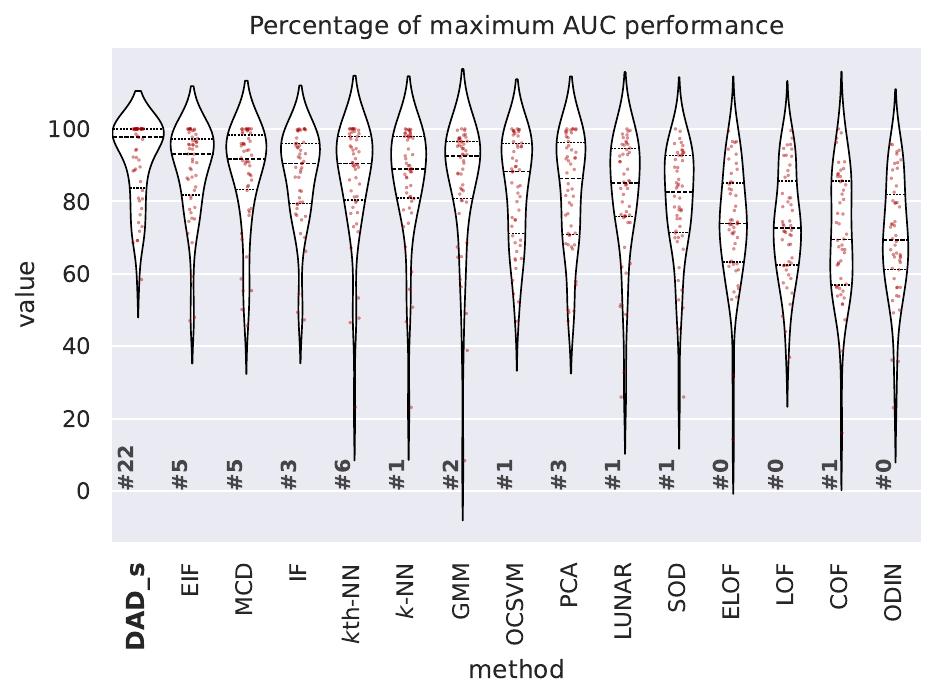}} &    
        \subfloat[DAD vs SOTA on all benchmark datasets]{\includegraphics[width=0.5\textwidth]{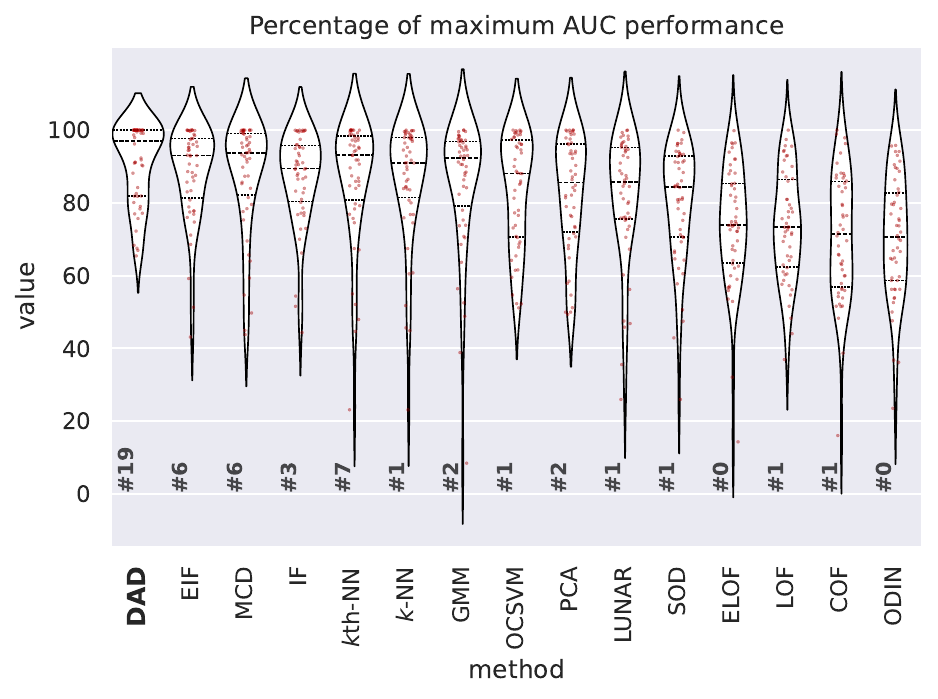}} \\
        \subfloat[DAD\_s vs SOTA on the global cluster]{\includegraphics[width=0.5\textwidth]{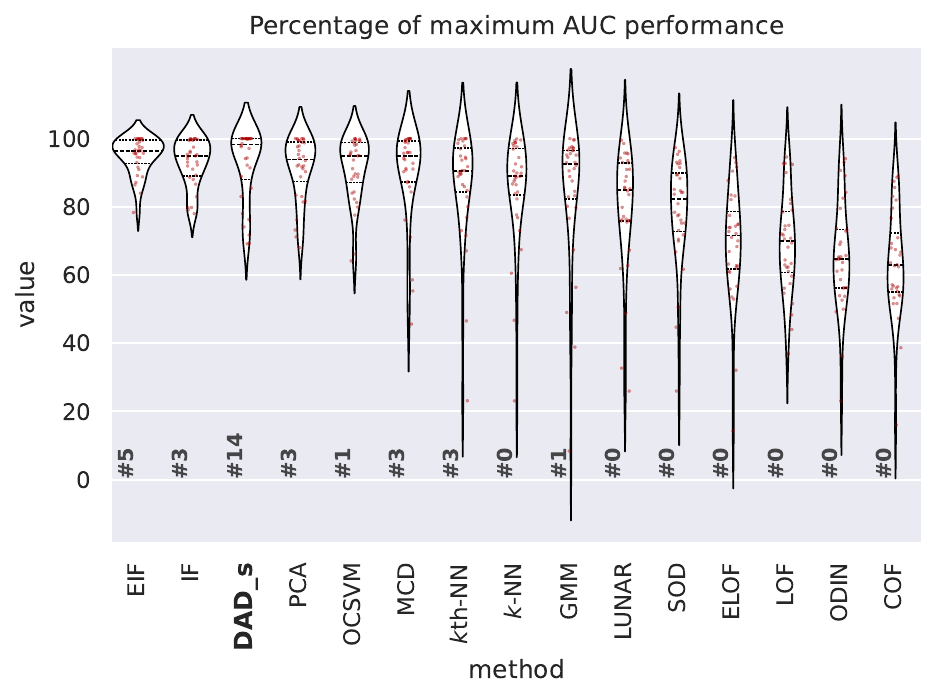}} &    
        \subfloat[DAD vs SOTA on the global cluster]{\includegraphics[width=0.5\textwidth]{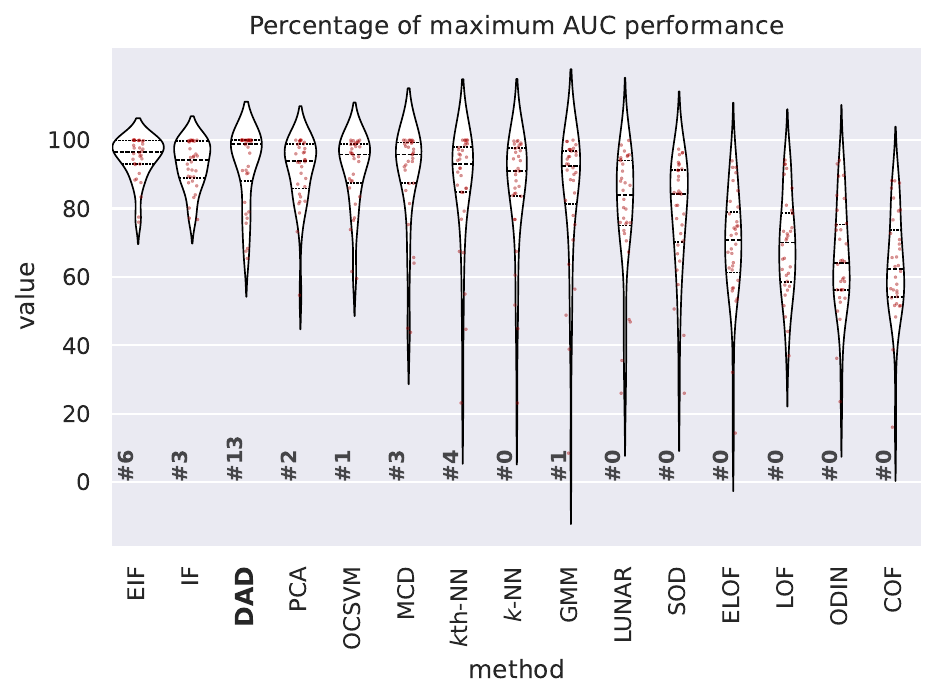}} \\
        \subfloat[DAD\_s vs SOTA on the local cluster]{\includegraphics[width=0.5\textwidth]{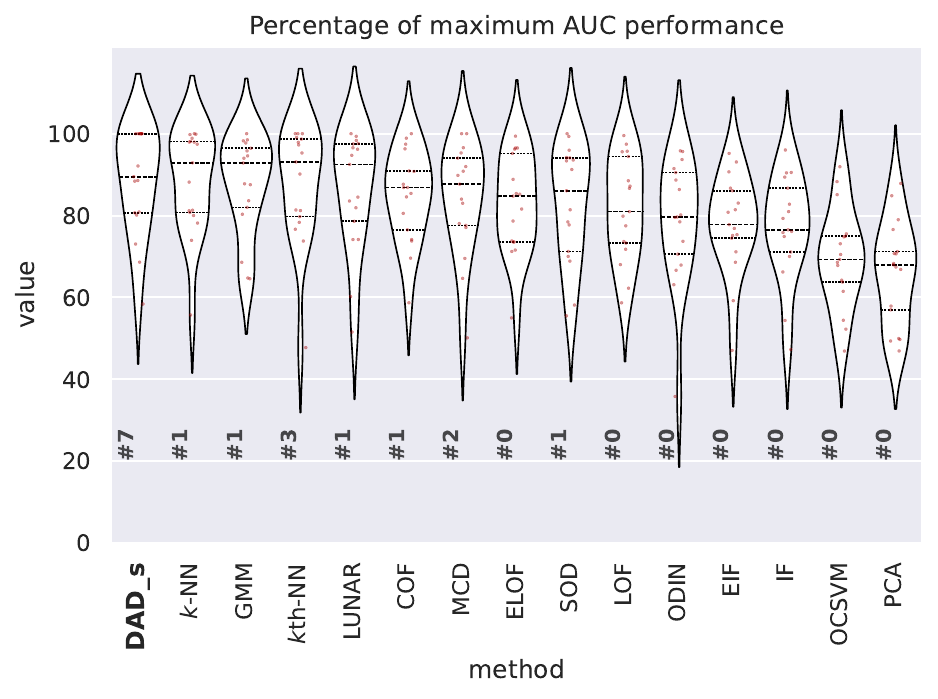}} &    
        \subfloat[DAD vs SOTA on the local cluster]{\includegraphics[width=0.5\textwidth]{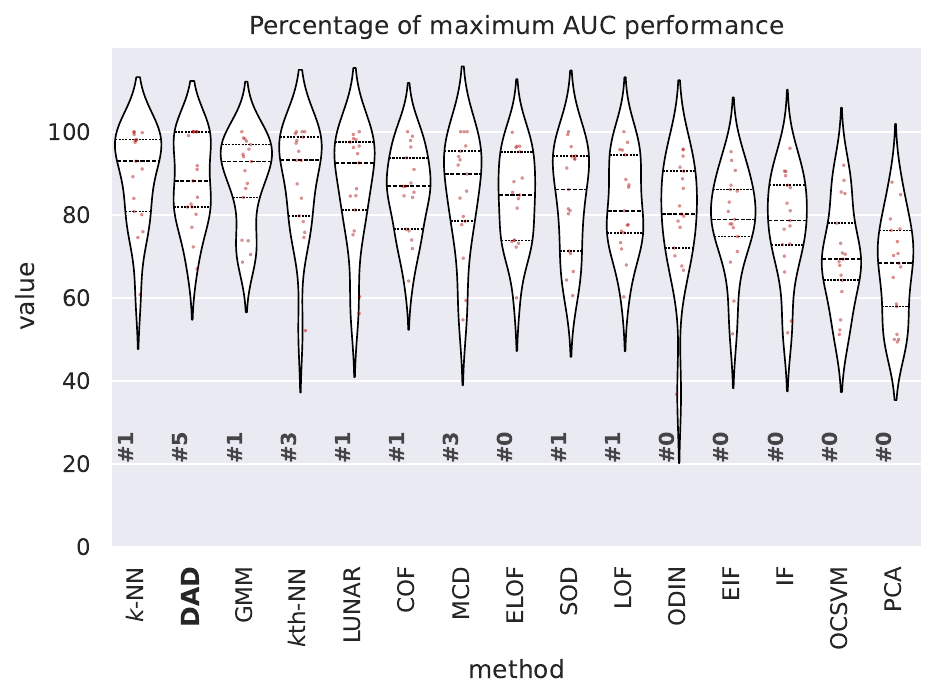}}     
    \end{tabular}
    \caption{Performance distribution of each method on: (a) and (b) all benchmark datasets, c) and d) the global cluster, e) and f) the local cluster.}        
    \label{fig:bresult_hpt_all}
\end{figure}

Figure~\ref{fig:bresult_hpt_all} presents the performance distribution of various anomaly detection methods across all the real-world benchmark datasets, the global cluster, and the local cluster. The results are shown as violin plots with scatter points, each representing a method’s performance on a specific dataset, expressed as a percentage of the maximum AUC achieved on that dataset, as defined in Eq.\eqref{eq:auc}. The numbers annotated beneath each violin plot (\textit{e.g.}, $\#value$) indicate how many datasets that method achieved the top performance on, according to Eq.~\eqref{eq:auc}. Worth mentioning is that the DAD and DAD\_s results are displayed in separate plots to avoid visual distortion, as they often alternate between first and second top performance across some datasets.

Figure~\ref{fig:bresult_hpt_all}~(a) presents the benchmark results for DAD\_s in comparison to SOTA anomaly detection methods. Notably, DAD\_s demonstrates superior performance, with its median value exceeding the upper quartile of EIF, the second-best performing method, and its lower quartile still surpassing the SOTA methods. Visually, the methods can be categorized into four classes based on their performance, ranked from highest to lowest. The first class is represented solely by DAD\_s. The second class includes EIF, MCD, IF, $k$th-NN, $k$-NN, and GMM, which show similar performance. The third class comprises OCSVM, PCA, LUNAR, and SOD. The fourth class consists of ELOF (ensemble-LOF, referred to as ELOF throughout this work), LOF, COF, and ODIN, which are the worst-performing methods. Similarly, Figure~\ref{fig:bresult_hpt_all}~(b) displays the results for DAD in comparison to SOTA anomaly detection methods. Although DAD shows a lower median than DAD\_s, it still achieves a higher median than SOTA methods. Its upper quartile is close to the maximum bound, similar to DAD\_s, and still higher than SOTA methods. Accordingly, DAD\_s obtained the top performance for 22/50 datasets, significantly more than other methods. This is highlighted by the greater concentration of performance scatter points at 100 percent.

Following the methodology in~\citep{bouman2024unsupervised}, and similar to Table~\ref{tab:outperforms_max}, Table~\ref{tab:outperforms_HPT}~(a) presents the results of the outperforming analysis on the benchmark dataset. Accordingly, we observe that DAD\_s outperforms a larger number of methods at $p=0.05$, including ODIN, COF, LOF, ELOF, SOD, and PCA ($6/15$). Moreover, its mean AUC is notably higher than that of EIF, the second-best performing method after the two different configurations of DAD (mean AUC: DAD\_s = $0.8027$ vs EIF = $0.7777$). Although $k$th-NN outperforms $6/15$ methods at $p=0.1$, its mean AUC is much lower than that of DAD\_s (mean AUC: DAD\_s = $0.8027$, $k$th-NN = $0.7571$). This indicates that DAD\_s consistently delivers superior performance across datasets within the benchmark compared to SOTA anomaly detection methods. Furthermore, EIF and MCD exhibit similar performance, while $k$-NN variants ($k$-NN and $k$th-NN) and GMM show comparable results with mean AUC values around $0.75$. However, GMM outperforms a smaller number of methods compared to $k$-NN. In contrast, ELOF, LOF, COF, and ODIN are the weakest-performing methods, with mean AUC values close to $0.6$, indicating that their performance is nearly equivalent to a random classifier, making them generally unsuitable choices for anomaly detection. For a heatmap view of the method-dataset combination performance on the benchmark dataset, see Figure~\ref{fig:bresult_cluster_hpt_all}.

As a result, our DAD\_s method consistently delivers strong performance across many datasets in the benchmark, highlighting its robustness across various scenarios and positioning it as the best option for anomaly detection across different data and anomaly types.

\begin{table*}[t]
\caption{Results of the Nemenyi post-hoc analysis across benchmark, global cluster, and local cluster. See~\ref{tab:outperforms_max} for an explanation of the symbols.}
\label{tab:outperforms_HPT}
\centering
\begin{subtable}[t]{\textwidth}
\caption{All benchmark datasets}
\resizebox{\textwidth}{!}{%
\begin{scriptsize}\begin{tabular}{lcccccccccccccccc|c}
\toprule
 & {ODIN} & {COF} & {LOF} & {ELOF} & {SOD} & {LUNAR} & {PCA} & {OCSVM} & {GMM} & {$k$-NN} & {$k$th-NN} & {IF} & {MCD} & {EIF} & {DAD} & {DAD\_s} & {\textbf{Mean AUC}}  \\
\midrule
DAD\_s & ++ & ++ & ++ & ++ & ++ &  & ++ &  &  &  &  &  &  &  &  &  & \textbf{0.8027} \\
DAD & ++ & ++ & ++ & ++ & ++ &  &  &  &  &  &  &  &  &  &  &  & 0.7985 \\
EIF & ++ & ++ & ++ & ++ & ++ &  &  &  &  &  &  &  &  &  &  &  & 0.7777 \\
MCD & ++ & ++ & ++ & ++ &  &  &  &  &  &  &  &  &  &  &  &  & 0.7704 \\
IF & ++ & ++ & ++ & ++ &  &  &  &  &  &  &  &  &  &  &  &  & 0.7639 \\
$k$th-NN & ++ & ++ & ++ & ++ & ++ &  & + &  &  &  &  &  &  &  &  &  & 0.7571 \\
$k$-NN & ++ & ++ & ++ & ++ & + &  &  &  &  &  &  &  &  &  &  &  & 0.7517 \\
GMM & ++ & ++ & ++ & ++ &  &  &  &  &  &  &  &  &  &  &  &  & 0.7448 \\
OCSVM & ++ & + &  &  &  &  &  &  &  &  &  &  &  &  &  &  & 0.7342 \\
PCA &  &  &  &  &  &  &  &  &  &  & - &  &  &  &  & -{}- & 0.7271 \\
LUNAR & ++ & ++ &  &  &  &  &  &  &  &  &  &  &  &  &  &  & 0.7119 \\
SOD &  &  &  &  &  &  &  &  &  & - & -{}- &  &  & -{}- & -{}- & -{}- & 0.6956 \\
ELOF &  &  &  &  &  &  &  &  & -{}- & -{}- & -{}- & -{}- & -{}- & -{}- & -{}- & -{}- & 0.6422 \\
LOF &  &  &  &  &  &  &  &  & -{}- & -{}- & -{}- & -{}- & -{}- & -{}- & -{}- & -{}- & 0.6396 \\
COF &  &  &  &  &  & -{}- &  & - & -{}- & -{}- & -{}- & -{}- & -{}- & -{}- & -{}- & -{}- & 0.6147 \\
ODIN &  &  &  &  &  & -{}- &  & -{}- & -{}- & -{}- & -{}- & -{}- & -{}- & -{}- & -{}- & -{}- & 0.6089 \\
\bottomrule
\end{tabular}\end{scriptsize}
}
\end{subtable}

\vspace{1em} 

\begin{subtable}[t]{\textwidth}
\caption{Global cluster}
\resizebox{\textwidth}{!}{%
\begin{scriptsize}\begin{tabular}{lcccccccccccccccc|c}
\toprule
 & {COF} & {ODIN} & {LOF} & {ELOF} & {SOD} & {LUNAR} & {GMM} & {$k$-NN} & {$k$th-NN} & {MCD} & {OCSVM} & {PCA} & {DAD} & {DAD\_s} & {IF} & {EIF} & {\textbf{Mean AUC}} \\
\midrule
EIF & ++ & ++ & ++ & ++ & ++ & ++ &  &  &  &  &  &  &  &  &  &  & \textbf{0.8520} \\
IF & ++ & ++ & ++ & ++ & ++ &  &  &  &  &  &  &  &  &  &  &  & 0.8345 \\
DAD\_s & ++ & ++ & ++ & ++ & ++ & + &  &  &  &  &  &  &  &  &  &  & 0.8264 \\
DAD & ++ & ++ & ++ & ++ & ++ &  &  &  &  &  &  &  &  &  &  &  & 0.8248 \\
PCA & ++ & ++ & ++ & ++ & ++ &  &  &  &  &  &  &  &  &  &  &  & 0.8205 \\
OCSVM & ++ & ++ & ++ & ++ & ++ &  &  &  &  &  &  &  &  &  &  &  & 0.8193 \\
MCD & ++ & ++ & ++ & ++ & + &  &  &  &  &  &  &  &  &  &  &  & 0.8104 \\
$k$th-NN & ++ & ++ & ++ & ++ & ++ &  &  &  &  &  &  &  &  &  &  &  & 0.7759 \\
$k$-NN & ++ & ++ & ++ & ++ &  &  &  &  &  &  &  &  &  &  &  &  & 0.7643 \\
GMM & ++ & ++ & ++ & ++ &  &  &  &  &  &  &  &  &  &  &  &  & 0.7544 \\
LUNAR & ++ & + &  &  &  &  &  &  &  &  &  &  &  & - &  & -{}- & 0.7129 \\
SOD &  &  &  &  &  &  &  &  & -{}- & - & -{}- & -{}- & -{}- & -{}- & -{}- & -{}- & 0.7012 \\
ELOF &  &  &  &  &  &  & -{}- & -{}- & -{}- & -{}- & -{}- & -{}- & -{}- & -{}- & -{}- & -{}- & 0.6131 \\
LOF &  &  &  &  &  &  & -{}- & -{}- & -{}- & -{}- & -{}- & -{}- & -{}- & -{}- & -{}- & -{}- & 0.6121 \\
ODIN &  &  &  &  &  & - & -{}- & -{}- & -{}- & -{}- & -{}- & -{}- & -{}- & -{}- & -{}- & -{}- & 0.5781 \\
COF &  &  &  &  &  & -{}- & -{}- & -{}- & -{}- & -{}- & -{}- & -{}- & -{}- & -{}- & -{}- & -{}- & 0.5614 \\
\bottomrule
\end{tabular}\end{scriptsize}
}
\end{subtable}

\vspace{1em}

\begin{subtable}[t]{\textwidth}
\caption{Local cluster}
\resizebox{\textwidth}{!}{%
\begin{scriptsize}\begin{tabular}{lcccccccccccccccc|c}
\toprule
 & {PCA} & {OCSVM} & {IF} & {EIF} & {ODIN} & {LOF} & {SOD} & {ELOF} & {MCD} & {COF} & {LUNAR} & {$k$th-NN} & {GMM} & {DAD} & {$k$-NN} & {DAD\_s} & {\textbf{Mean AUC}} \\
\midrule
DAD\_s & ++ & ++ &  &  &  &  &  &  &  &  &  &  &  &  &  &  & \textbf{0.7632} \\
$k$-NN & ++ & ++ & ++ & ++ & ++ &  &  &  &  &  &  &  &  &  &  &  & 0.7509 \\
DAD & ++ & + &  &  &  &  &  &  &  &  &  &  &  &  &  &  & 0.7492 \\
GMM & ++ & ++ &  &  &  &  &  &  &  &  &  &  &  &  &  &  & 0.7460 \\
$k$th-NN & ++ & ++ & ++ & + & ++ &  &  &  &  &  &  &  &  &  &  &  & 0.7439 \\
LUNAR & ++ & ++ &  &  &  &  &  &  &  &  &  &  &  &  &  &  & 0.7305 \\
COF & ++ & ++ &  &  &  &  &  &  &  &  &  &  &  &  &  &  & 0.7230 \\
MCD & ++ & + &  &  &  &  &  &  &  &  &  &  &  &  &  &  & 0.7171 \\
ELOF & ++ &  &  &  &  &  &  &  &  &  &  &  &  &  &  &  & 0.7079 \\
SOD & ++ &  &  &  &  &  &  &  &  &  &  &  &  &  &  &  & 0.7039 \\
LOF & ++ &  &  &  &  &  &  &  &  &  &  &  &  &  &  &  & 0.6990 \\
ODIN &  &  &  &  &  &  &  &  &  &  &  & -{}- &  &  & -{}- &  & 0.6715 \\
EIF &  &  &  &  &  &  &  &  &  &  &  & - &  &  & -{}- &  & 0.6633 \\
IF &  &  &  &  &  &  &  &  &  &  &  & -{}- &  &  & -{}- &  & 0.6551 \\
OCSVM &  &  &  &  &  &  &  &  & - & -{}- & -{}- & -{}- & -{}- & - & -{}- & -{}- & 0.5927 \\
PCA &  &  &  &  &  & -{}- & -{}- & -{}- & -{}- & -{}- & -{}- & -{}- & -{}- & -{}- & -{}- & -{}- & 0.5607 \\
\bottomrule
\end{tabular}\end{scriptsize}
}
\end{subtable}
\end{table*}

\paragraph{Global Cluster}
As shown in Figure~\ref{fig:bresult_hpt_all}~(c) and (d), the performance of the two configurations of our method on the global cluster is compared to SOTA anomaly detection methods. While EIF and IF are ranked first and second overall on the global cluster, our methods -- although ranked third -- exhibit higher median values and a higher upper quartile than EIF and the other methods. This indicates that in a greater density of cases, DAD\_s tends to deliver stronger performance. Additionally, as annotated in Figure~\ref{fig:bresult_hpt_all}~(c), DAD\_s achieves the top performance on 14/33 datasets within the global cluster, significantly outperforming other methods in this regard. 

Referring to the outperform results in Table~\ref{tab:outperforms_HPT}~(b), EIF and IF achieve higher mean AUC scores (EIF = $0.8520$, IF = $0.8345$) compared to DAD\_s ($0.8264$). In terms of outperforming other methods, EIF surpasses the most ($6/15$) at $p=0.05$, while DAD\_s outperforms $5/15$ at $p=0.05$ and $6/15$ at $p=0.10$, making it the second most outperforming method. In addition, competition within the global cluster shows that IF, DAD, PCA, and OCSVM achieve very competitive results, both in terms of the number of outperforming methods and their mean AUC scores. Again, LOF, ELOF, ODIN, and COF are the worst-performing methods, with mean AUC values near $0.6$, suggesting their performance is almost equivalent to that of a random classifier on the global cluster. For a heatmap view of the method-dataset combination performance on the global cluster, see Figure~\ref{fig:bresult_cluster_hpt_all}.

As a result, EIF delivers better average performance on the global cluster datasets, while our DAD\_s method still achieves the top performance on a greater number of datasets within the global cluster.

\paragraph{Local Cluster}Figure~\ref{fig:bresult_hpt_all}~(e) and (f) presents the AUC performance comparison of SOTA methods on the local cluster. As shown, DAD\_s is ranked first, though $k$-NN achieves a higher median. Nevertheless, DAD\_s maintains the highest upper quartile among all methods. In Figure~\ref{fig:bresult_hpt_all}~(f), the other configuration of our method, DAD, is ranked second with a lower median than DAD\_s. 

Table~\ref{tab:outperforms_HPT}~(c) presents the outperforming counts for each method alongside their mean AUC performance on the local cluster. As observed, both $k$-NN variants ($k$-NN and $k$th-NN) outperform more methods than others, with $k$-NN surpassing $5/15$ methods at $p=0.05$ and $k$th-NN achieving the same count at $p=0.1$. However, DAD\_s maintains a higher mean AUC than $k$-NN, the second-highest method in this regard (mean AUC: DAD\_s = 0.7632, $k$-NN = 0.7509). Furthermore, Figure~\ref{fig:bresult_hpt_all}~(e) highlights that DAD\_s attains the top performance on a larger number of datasets, achieving this distinction on 7/17 datasets. EIF, IF, OCSVM, and PCA are the weakest performers on the local cluster, with OCSVM and PCA exhibiting performance levels close to that of a random classifier. For a detailed view of the method-dataset combination performance on the local cluster, see Figure~\ref{fig:bresult_cluster_hpt_all}.

As a result, $k$-NN outperforms more methods, while DAD\_s still maintains a higher mean AUC and achieves the top performance on a significantly larger number of datasets within the local cluster.

\paragraph{Summary of Findings}
DAD\_s stands out as the most effective anomaly detection method across a broad range of data and anomaly types. With a remarkable $22/50$ top performance across benchmark datasets, DAD\_s consistently outperforms other methods, exhibiting a mean AUC far superior to the competition. On the global cluster, DAD\_s holds its own against EIF, which achieves a higher mean AUC, but DAD\_s significantly surpasses EIF in terms of the number of datasets where it attains the top performance (EIF = $5/33$ vs. DAD\_s = $14/33$). On the local cluster, while $k$-NN outperforms more methods, DAD\_s excels with a higher mean AUC and achieves the top performance on a much greater number of datasets (DAD\_s = $7/17$, $k$-NN = $1/17$).

The key takeaway from this evaluation is that the competition among DAD\_s, EIF, and $k$-NN reveals clear distinctions: EIF struggles with local density anomalies, $k$-NN is less effective on the global cluster compared to EIF and DAD\_s, and only DAD\_s proves to be versatile enough to deliver robust performance in both local and global tasks. This makes DAD\_s the most reliable and effective anomaly detection solution across diverse scenarios, showcasing its superior generalization capabilities. Additional evaluation results on the benchmark dataset, following traditional benchmarking approaches such as peak, average, and default performances are provided in Appendix~\ref{sec:app_traditional}.

\subsubsection{Performance on High-Dimensional Data}
The demand for higher dimensional data is growing across many real-time applications. Therefore, it is crucial to assess whether a real-time anomaly detection method can handle the curse of high dimensionality. To evaluate our proposed method, we selected high-dimensional datasets within the benchmark with feature sizes $d > 100$. These datasets include  {\em musk} ($d=166$), {\em arrhythmia} ($d=257$), {\em speech} ($d=400$), and {\em internetads} ($d=1555$).

\begin{figure*}[t] 
\centering {\includegraphics[width=0.5\textwidth]{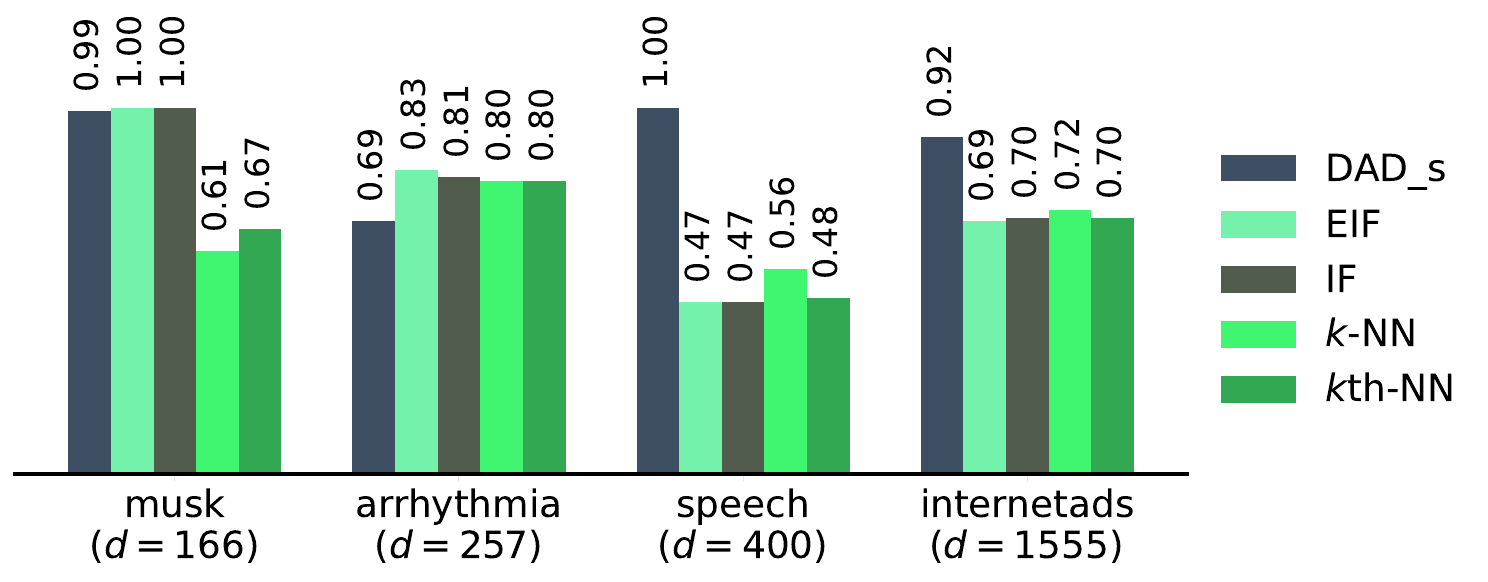}}
\caption{The AUC performance comparison of DAD\_s, EIF, IF, $k$-NN, and $k$th-NN on high dimensional datasets within the benchmark.} 
\label{fig:highdim} 
\end{figure*}

We selected the most competitive methods from our benchmark results, including EIF, IF, $k$-NN, and $k$th-NN, to compare with our proposed method, DAD\_s, for high dimensional data evaluation. The performance of these methods on high dimensional datasets should demonstrate their effectiveness. Figure~\ref{fig:highdim} illustrates the AUC results of the methods on these high dimensional datasets.

As seen in Figure~\ref{fig:highdim}, DAD\_s, EIF, and IF are the most effective methods on {\em musk} dataset, with AUC performances $\approx1.0$. For {\em arrhythmia} dataset, DAD\_s shows a lower performance ($\approx0.7$) compared to the other methods ($\approx0.8$). As the dimensionality increases, in the case of {\em speech} dataset ($d=400$), our method effectively handles the data, achieving an AUC performance of $1.0$, while other methods range between $0.47$ and $0.56$. The largest dimensional dataset, {\em internetads} with $d=1555$ features, reveals that our method still achieves the most effective performance ($\approx 0.9$) compared to others ($\approx0.7$).

Although this evaluation could be questioned due to the limited number of scenarios and datasets, as well as the types of anomalies, it underscores that DAD\_s performance is not adversely affected by increasing dimensionality as a crucial criterion for real-time high-dimensional data anomaly detection applications. We further discuss our explanation for this case in Section~\ref{sec:discussion}.

To better identify the most suitable method for real-time anomaly detection applications and align with the objectives of this work, we propose an additional analysis that examines the trade-off between performance and time complexity. Since the real-time applicability of anomaly detection methods heavily depends on this balance, it is crucial to investigate this aspect to provide a comprehensive understanding. The following section presents this analysis.

\subsubsection{Performance-Time Complexity Trade-Offs}
The benchmarking setup introduced in Section~\ref{sec:benchmark_setup} also records the wall-clock time for each method-dataset combination. This allows for an in-depth understanding of the real-time performance of SOTA anomaly detection methods in the benchmark. The benchmark was conducted on a machine equipped with four Intel(R) Xeon(R) E5-2650 v4 @ 2.20GHz CPU processors (48 cores in total) and 32 GB of memory. Since all benchmark runs were consistently executed on this system without interruption, the wall-clock time measurements remain reliable and comparable across methods.

\begin{figure}
    \centering
    {\includegraphics[width=0.5\textwidth]{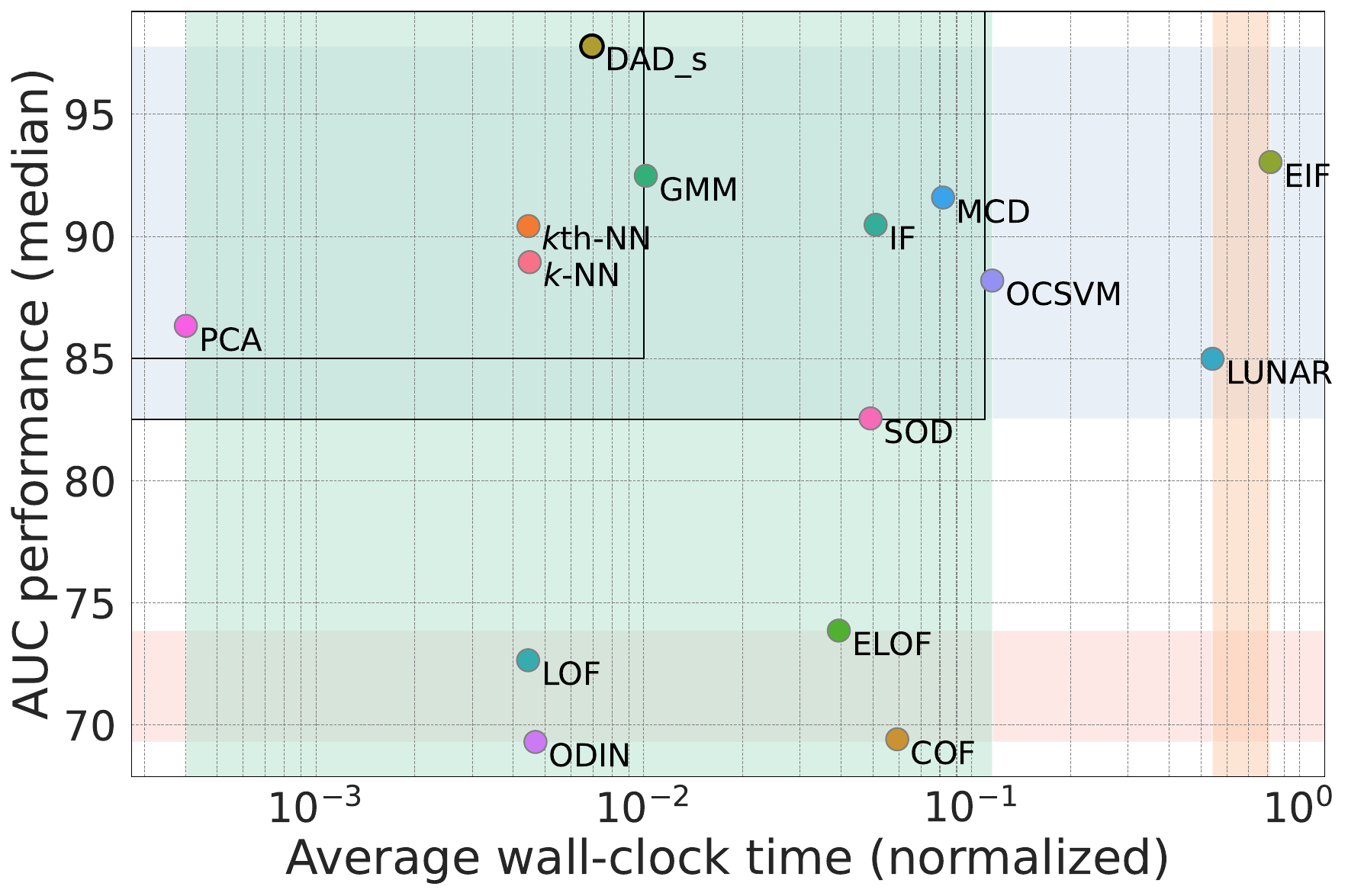}}        
    \caption{Scatter plot showing the median percentage of the maximum AUC versus the average wall-clock time for each method on the benchmark.}
    \label{fig:hpt_median_auc_vs_avg_time}
\end{figure}

The wall-clock time recorded for each method-dataset combination was averaged across datasets to provide a representative time complexity measure for each method. To visualize the trade-off between wall-clock time and AUC performance, the median AUC values from Fig.~\ref{fig:bresult_hpt_all}~(a) are used. This allows each method to be represented by its median performance plotted against its average wall-clock time. Fig.~\ref{fig:hpt_median_auc_vs_avg_time} illustrates these performance-time complexity trade-offs, where average wall-clock times are normalized relative to the maximum value among the methods.

To investigate the grouping of methods based on their performance and computational efficiency, we employed KMeans clustering with an automated cluster number estimation using the Elbow Method. This approach was applied separately to the full profiles of AUC values and wall-clock times across all datasets, enabling a two-dimensional analysis visualized in the scatter plot. The Elbow Method determined the optimal number of clusters by computing the within-cluster sum of squares (inertia) for $k=1$ to $k=min(10,N)$, where $N$ is the number of methods, and identifying the point of maximum curvature via the second derivative of the inertia curve. This heuristic balances cluster granularity and compactness without requiring manual specification of $k$. The resulting clusters were visualized as horizontal and vertical background strips in the scatter plot, representing AUC-based and wall-clock time-based groupings, respectively. This clustering approach highlights natural groupings of methods with similar performance or runtime characteristics, facilitating interpretation of trade-offs between efficacy and efficiency.

In this analysis, methods positioned in the top-left region of the scatter plot represent the best trade-offs, with high performance and low average wall-clock times. Ideally, the most optimal methods would be positioned in the top-left corner. The overlapping region in the top-left between the horizontal and vertical clusters suggests the optimal trade-off zone, where methods strike the best balance between efficiency and efficacy.

As shown, EIF, LUNAR, LOF, ELOF, ODIN, and COF exhibit the worst trade-offs between performance efficacy and computational complexity. For real-time exploitation, these methods would be less competitive compared to other methods. From the optimal trade-off zone, we can see two rectangles centered around the top-left corner: the first, positioned closer to the optimal point, encompasses DAD\_s, GMM, $k$th-NN, $k$-NN, and PCA. The second rectangle is a wider, including IF, MCD, OCSVM, and SOD in addition to the methods in the first rectangle. We focus on the narrower rectangle, as it represents a group of methods that are near the optimal trade-off point, indicating a higher degree of balance between performance efficacy and computational complexity for real-time applications.

Obviously, the narrower rectangle in the optimal trade-off zone highlights two key traits: DAD\_s, which achieves maximal performance and stands out as the most effective method, and PCA, which records the least average wall-clock time but notably weaker performance, making it the least complex method in terms of computational requirements. Since PCA is significantly less effective than the competition -- namely DAD\_s, GMM, $k$th-NN, and $k$-NN -- in terms of delivering anomaly detection performance, we exclude it from the optimal trade-off seeking procedure.

Prioritizing the efficacy of anomaly detection methods while minimizing computational complexity leads to a clear competition between DAD\_s and $k$th-NN. While GMM exhibits higher average wall-clock time than DAD\_s without delivering any performance benefit, $k$-NN shows a similar average wall-clock time compared to $k$th-NN but results in slightly lower performance. The median percentage of maximum AUC for DAD\_s is 97.8, significantly higher than $k$th-NN at 90.4. Regarding average wall-clock time, DAD\_s takes $6.9 \times 10^{-3}$ seconds, while $k$th-NN takes $4.4 \times 10^{-3}$ seconds on the normalized axis. Despite this, DAD\_s achieves a maximal performance gain without sacrificing significant wall-clock time compared to $k$th-NN. Through an optimal trade-off approach that prioritizes performance while preserving real-time throughput, DAD\_s outperforms $k$th-NN by delivering far superior performance with only a slight increase in runtime.

From a real-time exploitation perspective, our method is uniquely capable of operating in a truly streaming fashion -- it processes each incoming data point individually without storing past instances or requiring batch processing. In contrast, all other competing methods necessitate access to previous data, often employing a sliding window or batch-based approach. Importantly, while the reported wall-clock times for other methods were measured using static, preloaded datasets in a single-pass evaluation, the timing for our method was recorded under streaming conditions. This means that each sample was sequentially fed into the algorithm, mimicking real-time operation, and a corresponding anomaly score was produced on the fly. Consequently, the measured runtime for our method reflects a more realistic deployment scenario and arguably represents a more stringent evaluation condition. Therefore, the wall-clock time comparison is conservative in favor of competing methods, making our results particularly fair under practical, real-time constraints.

Ultimately, performance is a critical factor in system functionality and cannot be easily compromised, particularly in real-time applications. Ensuring high accuracy is essential for reliable and effective functionality, as errors or inaccuracies can severely impact the overall system's reliability and outcomes. While throughput remains important in time-sensitive scenarios, performance takes precedence, and methods like DAD\_s, which provide high performance without excessively sacrificing runtime, are typically the more favorable choice in real-world applications. Therefore, the trade-off analysis between performance (AUC) and computational complexity (average wall-clock time) firmly concludes that DAD\_s is the most effective and efficient method for real-time anomaly detection. Its ability to balance both high performance and manageable computational complexity positions it as the ideal choice for practical deployment in a wide range of time-sensitive applications.

\subsubsection{DAD's Hyperparameter Sensitivity Analysis}
\label{sec:Hyper_sen}
\begin{figure*}[t]
    \centering
    \begin{tabular}{cccc}
        {\includegraphics[width=0.22\textwidth]{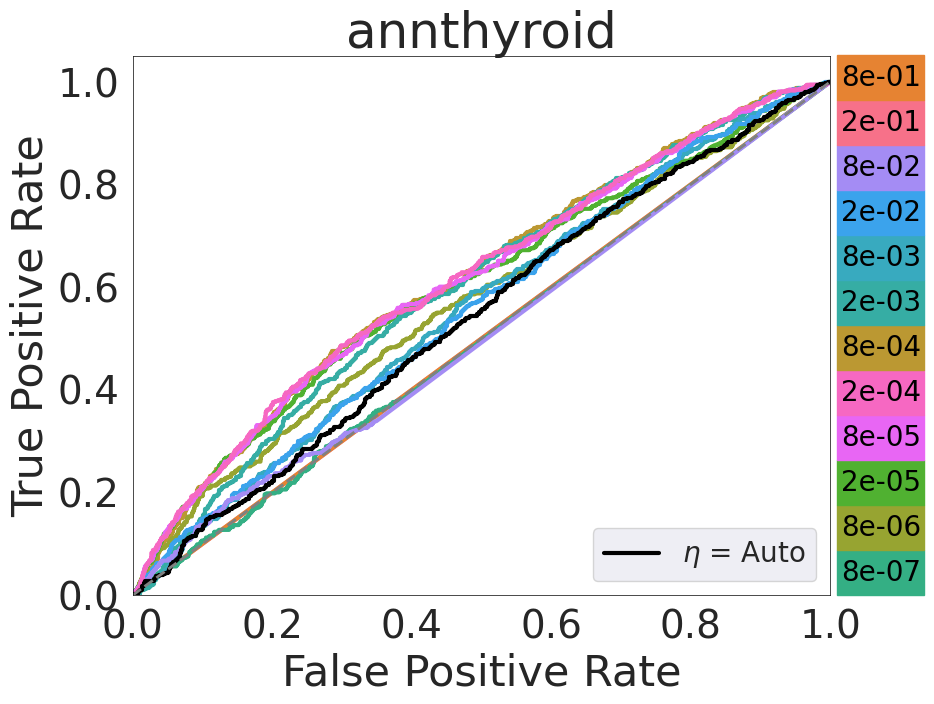}}      
        {\includegraphics[width=0.22\textwidth]{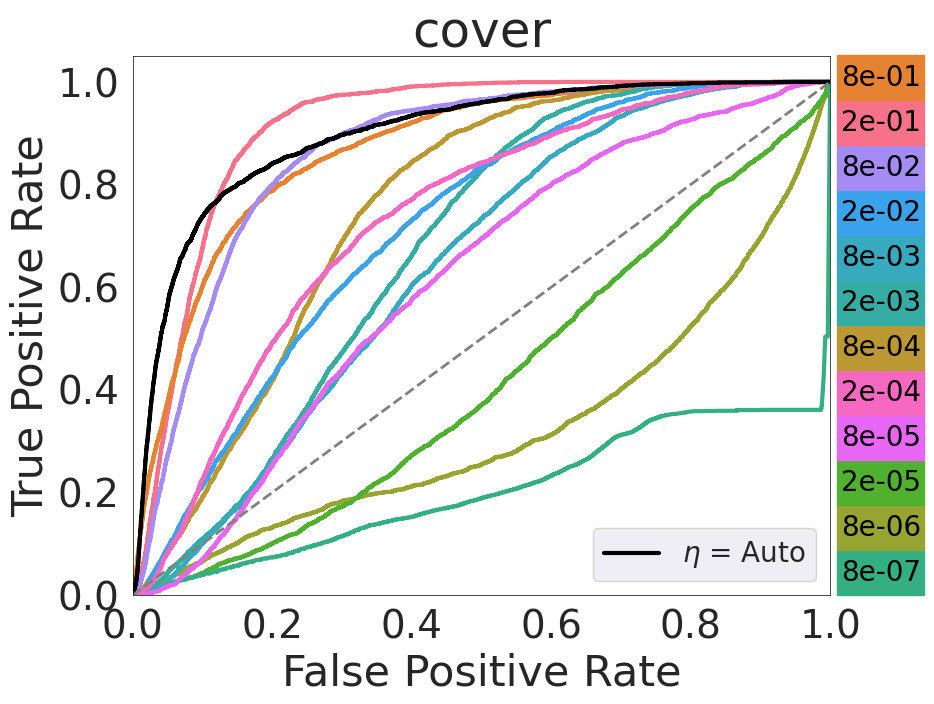}}               
        {\includegraphics[width=0.22\textwidth]{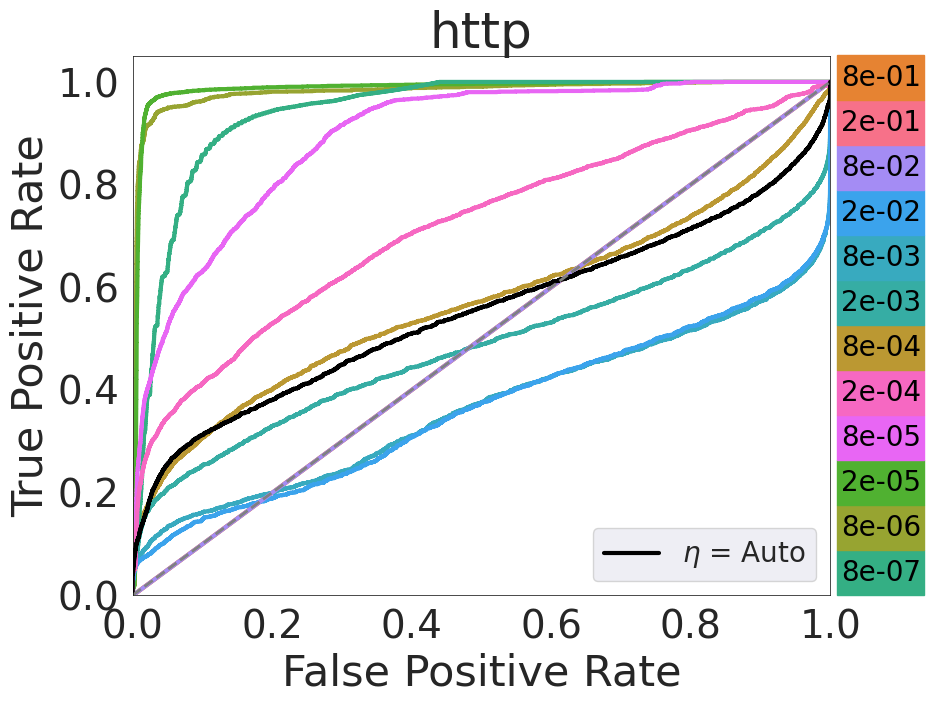}}        
        {\includegraphics[width=0.22\textwidth]{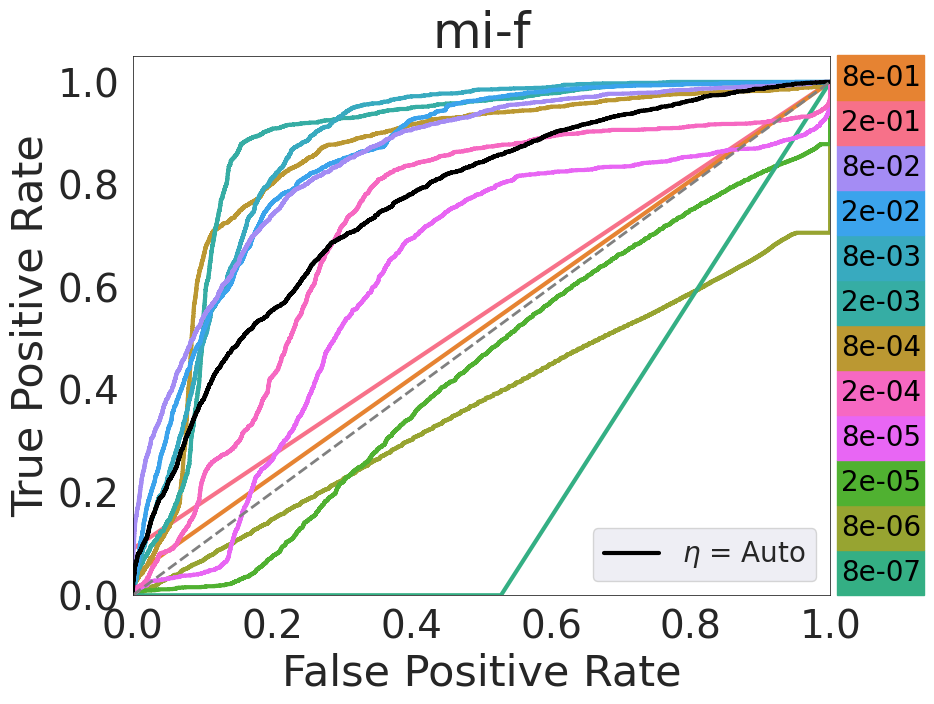}}     \\   
        {\includegraphics[width=0.22\textwidth]{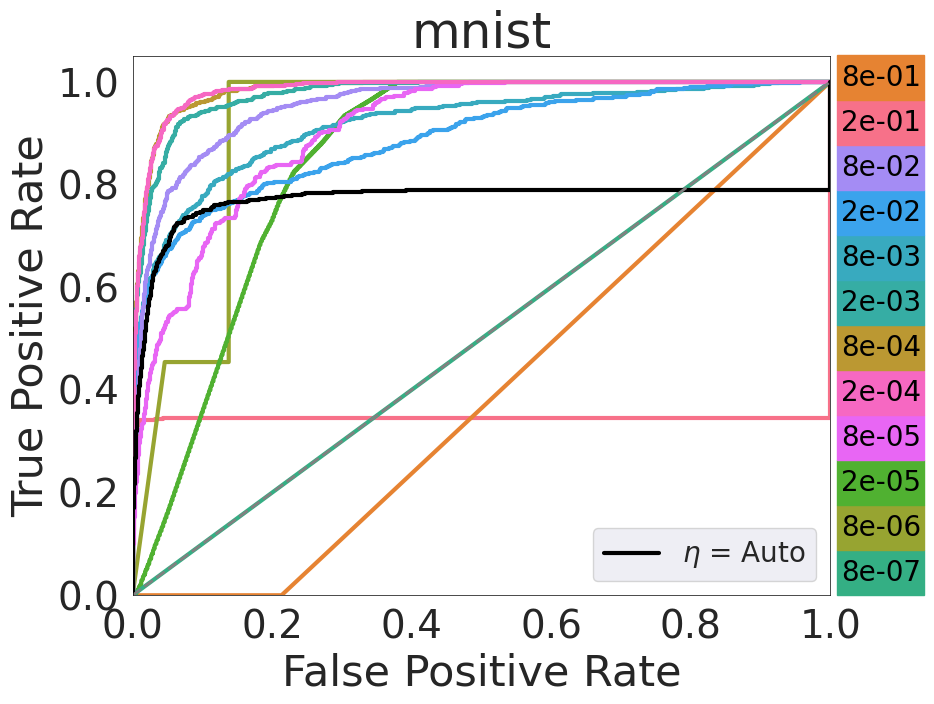}}     
        {\includegraphics[width=0.22\textwidth]{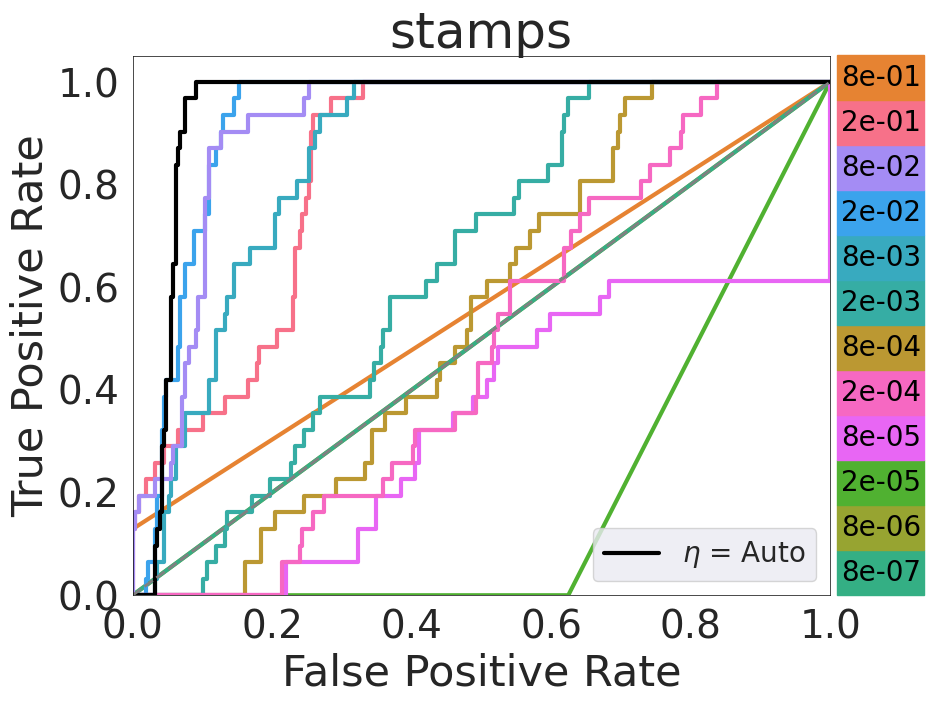}}  
        {\includegraphics[width=0.22\textwidth]{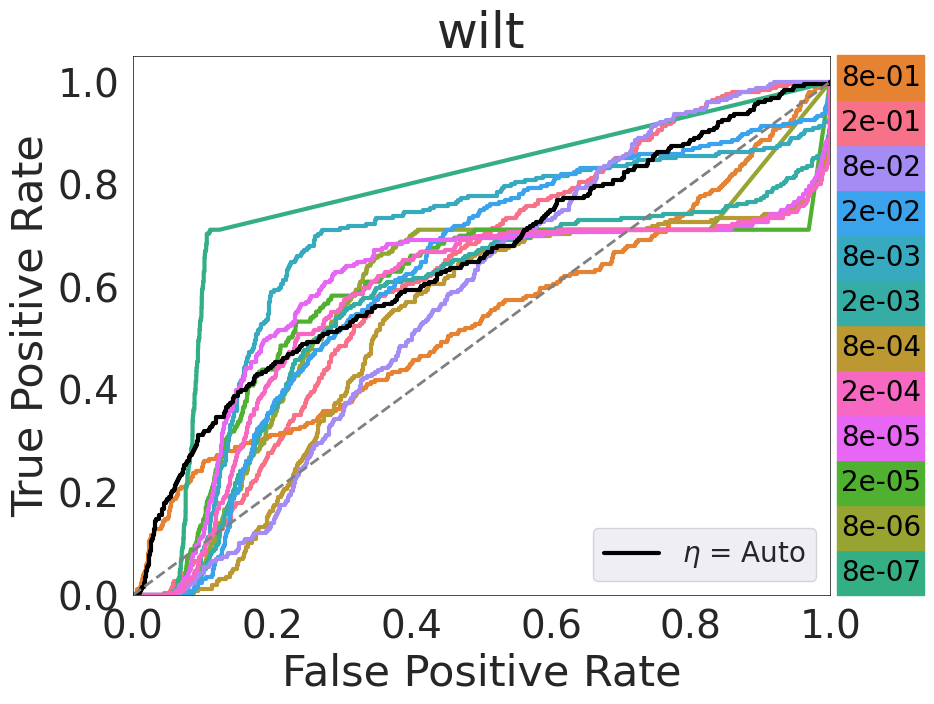}}       
        {\includegraphics[width=0.22\textwidth]{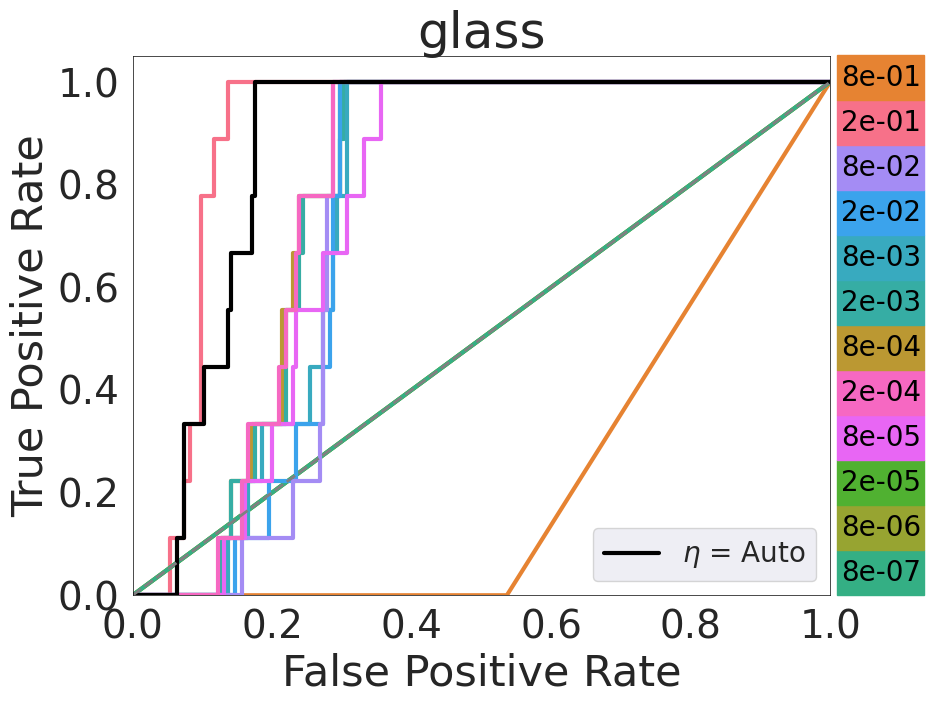}}      
    \end{tabular}
    \caption{Visualization of ROC curves for each dataset-hyperparameter ($\eta$) combinations. The color bar on the right of each plot maps the curve color to its corresponding $\eta$ value.}
    \label{fig:fig_Hyper_sen}
\end{figure*}

This section analyzes the effect of learning rate $\eta$ on DAD's performance. To simplify the visualization and reduce the size of the hyperparameter grid, we selected the smaller configuration of our method (DAD with a window size of $p=0$) for this analysis. Additionally, the automated variant of DAD, introduced using Algorithm~\ref{alg:DAD_alg_aut}, is included in the evaluation. To illustrate the impact of different $\eta$ configurations, we randomly selected eight datasets from the benchmark and visualized their ROC curves, including the automated variant. As shown in Figure~\ref{fig:fig_Hyper_sen}, the results highlight two key observations: First, performance variation across different $\eta$ configurations is substantial for some datasets, while minimal for others. Second, the automated variant, indicated by $\eta = Auto$, consistently identifies configurations with effective performance  --  achieving maximal or near-maximal results in 4 out of 8 datasets. Consequently, the automated version demonstrates robust performance across various scenarios.

Overall, DAD's performance is sensitive to the $\eta$ hyperparameter, which aligns with our hypothesis that the speed of learning interrelations between features directly depends on $\eta$. The optimal $\eta$ value varies based on the rate of change and variation within these interrelations, as well as the expected short- and long-term pattern detection in the scenario.
Hyperparameter tuning-based evaluation shows that tuning $\eta$ using a subset of data generalizes better performance to unseen data compared to SOTA methods. This characteristic is particularly advantageous when practitioners can obtain such a subset of data.

\section{Discussion}
\label{sec:discussion}
In this study, we introduced an innovative unsupervised anomaly detection method tailored for real-time applications. Through the generation of synthetic datasets that closely mimic real-world scenarios, we verified our concept and assessed its performance. Our method demonstrated distinct superiority over state-of-the-art methods, which proved ineffective on these synthetic datasets. Additionally, inspired by existing works, we proposed a hyperparameter tuning-based benchmarking approach for unsupervised anomaly detection methods. Benchmarking on comprehensive real-world data revealed that our method consistently outperformed others across a wide range of data and anomaly types. Specifically, compared to EIF and $k$-NN, our method showed the most consistent performance on both global and local anomaly types. While EIF performs better on global anomalies, its performance on local anomalies falls into the weakest group of methods. Moreover, $k$-NN shows moderate performance on global anomalies. Importantly, our method's performance remained robust despite increasing dimensionality, a crucial factor for real-time high-dimensional data anomaly detection applications. Furthermore, the performance-time complexity trade-offs highlighted that our method achieves an optimal balance by prioritizing performance while maintaining real-time throughput.

The proposed hyperparameter tuning approach assumes that practitioners are willing to provide a small amount of labeled data to enhance the accuracy and functionality of their systems. While this approach contradicts the principles of unsupervised learning, it can be supported by contemporary meta-learning techniques. By training a meta-learner on labeled datasets, patterns such as anomaly types and anomaly density can be identified and linked to appropriate hyperparameters for a given method. These hyperparameters can then be applied to unseen data in practice, where the meta-learner identifies similarities between the new data and its training data to tune the method accordingly. This integration of meta-learning techniques offers a pragmatic solution to improve the performance of unsupervised anomaly detection methods.

Since DAD leverages a similar concept to PCA-based methods, it is important to highlight the key distinctions between the two approaches, particularly in handling time-dependent multivariate data. A common PCA-based method first computes the covariance matrix $\Sigma$ of the original data matrix $D$, diagonalizes it as $\Sigma = P \Delta P^T$, and then transforms the data to a new decorrelated axis-system as $D' = DP$. Each column of $D'$ is standardized to unit variance by dividing it by its standard deviation. Finally, the outlier score of each row in $D'$ is determined by computing its squared Euclidean distance from the centroid of $D'$. Essentially, this approach diagonalizes the covariance matrix $\Sigma \in \mathbb{R}^{d \times d}$ using a decorrelator matrix $P \in \mathbb{R}^{d \times d}$. Hence, the decorrelator matrix is centered over all samples, and any deviation from this centroid is considered an outlier. However, multivariate time series data often exhibit time-dependent patterns of varying lengths (both short- and long-term dependencies). The static decorrelation matrix in PCA fails to capture these evolving dependencies, making it unsuitable for handling time-dependent multivariate data. In contrast, DAD employs an active learning-based decorrelator that updates sample by sample. This dynamic update process allows the decorrelator matrix to retain knowledge from past samples, with a learning rate determining how long previous information is preserved. A higher learning rate enables faster adaptation to new patterns, whereas a lower rate maintains longer-term dependencies. By tuning the learning rate, DAD can effectively capture both short- and long-term dependencies in multivariate time series, whereas PCA-based approaches remain limited in this regard.

Streaming data is endless, arrives swiftly, and frequently changes its distribution over time. This requires algorithms that can handle each incoming data point immediately, using minimal memory and without needing to revisit previous points. Storing entire sequences is impractical; instead, a method must maintain a concise internal representation of essential patterns. Existing solutions often utilize techniques like sliding windows, reservoir sampling, and forgetting mechanisms. While these methods can be effective, they often struggle with feature drift and concept evolution. The fixed-size window may either fail to adapt to rapid changes or combine multiple evolving patterns if it is too large. Our method addresses the requirement of real-time anomaly detection for data streams by processing each data point as it arrives without the need for storing past instances, thus avoiding memory burden on real-time processors such as edge devices. Abstracted information about past instances is only represented by the decorrelator matrix, which has a dimension of $d \times d$ and carries information about interrelations of features. The learning rate determines how quickly this decorrelator matrix should adapt to new time-dependent patterns, adjusting the trade-off between short- and long-term dependency learning. Therefore, no past instance of data in the non-sliding variant of our method (DAD) and a maximum of one past instance in the sliding variant (DAD\_s) contribute to real-time computation.

As the demand for higher dimensional data is dramatically growing in many applications, the curse of high-dimensional data further complicates anomaly detection tasks. Hence, understanding underlying interdependencies between features in such high-dimensional data is necessary. Our approach, which uses active learning of feature interdependencies through decorrelation with a decorrelator matrix, introduces a beneficial trait: as the dimension increases, the decorrelator matrix dimension scales up accordingly and continues to learn these interrelations. Therefore, it is believed that our proposed method is not adversely affected by these key requirements.

There are some insights into the method's future directions and its applications. It is evident that our anomaly detection method has superior potential in achieving maximal performance compared to SOTA methods if its learning rate is well-tuned. Therefore, a robust meta-learning technique can be embedded into our method, eventually converting this potential to reality. Additionally, empowering our method with an adaptive learning rate can assist in establishing an adaptive trade-off between short- and long-term time-dependency of evolving patterns, and significantly improve performance. Potential applications for our method include addressing covariate shift, as many real-world systems face the challenge of input data shifts, necessitating continuous monitoring of process behavior and prompt adaptive corrections. Our method can flag such patterns when observed, or even be used for early detection by monitoring changes in the orientation of the decorrelator learning curve. Additionally, the decorrelated input can facilitate adaptive corrections when covariate shift behavior chronically exists in a system.

In conclusion, this work endeavors to introduce an anomaly detection method that meets the key requirements of real-time data streaming applications, such as memory constraints, throughput, and high dimensionality, through single-pass processing of arriving data instances, making it suitable for deployment on edge devices. Our method effectively balances the trade-off between performance and time complexity by prioritizing performance as the main contributor to real-time system functionality, achieving the optimal trade-off with superior efficacy. Additionally, DAD's robustness to increasing dimensionality and its ability to adapt to evolving data patterns further underscore its potential for real-world applications involving massive, high-dimensional data streams.

\section{Acknowledgments}
This publication is part of the project ROBUST: Trustworthy AI-based Systems for Sustainable Growth with project number KICH3.L TP.20.006, which is (partly) financed by the Dutch Research Council (NWO), ASMPT, and the Dutch Ministry of Economic Affairs and Climate Policy (EZK) under the program LTP KIC 2020-2023. All content represents the opinion of the authors, which is not necessarily shared or endorsed by their respective employers and/or sponsors.


\begin{thebibliography}{10}

\bibitem{bouman2024unsupervised}
R.~Bouman, Z.~Bukhsh, and T.~Heskes, ``Unsupervised anomaly detection algorithms on real-world data: how many do we need?,'' {\em Journal of Machine Learning Research}, vol.~25, no.~105, pp.~1--34, 2024.

\bibitem{fisch2022real}
A.~T. Fisch, L.~Bardwell, and I.~A. Eckley, ``Real time anomaly detection and categorisation,'' {\em Statistics and Computing}, vol.~32, no.~4, p.~55, 2022.

\bibitem{9235582}
Y.~Jiang, S.~Yin, J.~Dong, and O.~Kaynak, ``A review on soft sensors for monitoring, control, and optimization of industrial processes,'' {\em IEEE Sensors Journal}, vol.~21, no.~11, pp.~12868--12881, 2021.

\bibitem{Jeschke2017}
S.~Jeschke, C.~Brecher, T.~Meisen, D.~Özdemir, and T.~Eschert, {\em Industrial Internet of Things and Cyber Manufacturing Systems}, pp.~3--19.
\newblock Springer Cham, 2017.

\bibitem{9915308}
H.~Nizam, S.~Zafar, Z.~Lv, F.~Wang, and X.~Hu, ``Real-time deep anomaly detection framework for multivariate time-series data in industrial {IoT},'' {\em IEEE Sensors Journal}, vol.~22, no.~23, pp.~22836--22849, 2022.

\bibitem{lampropoulos2019internet}
G.~Lampropoulos, K.~Siakas, and T.~Anastasiadis, ``Internet of things in the context of industry 4.0: An overview,'' {\em International Journal of Entrepreneurial Knowledge}, vol.~7, no.~1, 2019.

\bibitem{chen2017smart}
B.~Chen, J.~Wan, L.~Shu, P.~Li, M.~Mukherjee, and B.~Yin, ``Smart factory of industry 4.0: Key technologies, application case, and challenges,'' {\em IEEE Access}, vol.~6, pp.~6505--6519, 2017.

\bibitem{borsatti2021enabling}
D.~Borsatti, G.~Davoli, W.~Cerroni, and C.~Raffaelli, ``Enabling industrial {IoT} as a service with multi-access edge computing,'' {\em IEEE Communications Magazine}, vol.~59, no.~8, pp.~21--27, 2021.

\bibitem{liu2021towards}
Y.~Liu, R.~Zhao, J.~Kang, A.~Yassine, D.~Niyato, and J.~Peng, ``Towards communication-efficient and attack-resistant federated edge learning for industrial internet of things,'' {\em ACM Transactions on Internet Technology (TOIT)}, vol.~22, no.~3, pp.~1--22, 2021.

\bibitem{ADbench}
S.~Han, X.~Hu, H.~Huang, M.~Jiang, and Y.~Zhao, ``{ADB}ench: anomaly detection benchmark,'' in {\em Proceedings of the 36th International Conference on Neural Information Processing Systems}, NIPS '22, Curran Associates Inc., 2022.

\bibitem{ahmad2017unsupervised}
S.~Ahmad, A.~Lavin, S.~Purdy, and Z.~Agha, ``Unsupervised real-time anomaly detection for streaming data,'' {\em Neurocomputing}, vol.~262, pp.~134--147, 2017.

\bibitem{boniol2024dive}
P.~Boniol, Q.~Liu, M.~Huang, T.~Palpanas, and J.~Paparrizos, ``Dive into time-series anomaly detection: A decade review,'' {\em arXiv preprint arXiv:2412.20512}, 2024.

\bibitem{agyemang2006comprehensive}
M.~Agyemang, K.~Barker, and R.~Alhajj, ``A comprehensive survey of numeric and symbolic outlier mining techniques,'' {\em Intelligent Data Analysis}, vol.~10, no.~6, pp.~521--538, 2006.

\bibitem{marjani2017big}
M.~Marjani, F.~Nasaruddin, A.~Gani, A.~Karim, I.~A.~T. Hashem, A.~Siddiqa, and I.~Yaqoob, ``Big {IoT} data analytics: architecture, opportunities, and open research challenges,'' {\em IEEE access}, vol.~5, pp.~5247--5261, 2017.

\bibitem{han2019review}
Z.~Han, J.~Zhao, H.~Leung, K.~F. Ma, and W.~Wang, ``A review of deep learning models for time series prediction,'' {\em IEEE Sensors Journal}, vol.~21, no.~6, pp.~7833--7848, 2019.

\bibitem{de200625}
J.~G. De~Gooijer and R.~J. Hyndman, ``25 years of time series forecasting,'' {\em International Journal of Forecasting}, vol.~22, no.~3, pp.~443--473, 2006.

\bibitem{knox1998algorithms}
E.~M. Knox and R.~T. Ng, ``Algorithms for mining distancebased outliers in large datasets,'' in {\em Proceedings of the International Conference on Very Large Data Bases}, pp.~392--403, Citeseer, 1998.

\bibitem{knorr2000distance}
E.~M. Knorr, R.~T. Ng, and V.~Tucakov, ``Distance-based outliers: algorithms and applications,'' {\em The VLDB Journal}, vol.~8, no.~3, pp.~237--253, 2000.

\bibitem{ramaswamy2000efficient}
S.~Ramaswamy, R.~Rastogi, and K.~Shim, ``Efficient algorithms for mining outliers from large data sets,'' in {\em Proceedings of the 2000 ACM SIGMOD International Conference on Management of Data}, pp.~427--438, 2000.

\bibitem{samariya2023comprehensive}
D.~Samariya and A.~Thakkar, ``A comprehensive survey of anomaly detection algorithms,'' {\em Annals of Data Science}, vol.~10, no.~3, pp.~829--850, 2023.

\bibitem{LOFref}
M.~M. Breunig, H.-P. Kriegel, R.~T. Ng, and J.~Sander, ``{LOF}: identifying density-based local outliers,'' in {\em Proceedings of the 2000 ACM SIGMOD International Conference on Management of Data}, SIGMOD '00, p.~93–104, Association for Computing Machinery, 2000.

\bibitem{tang2002enhancing}
J.~Tang, Z.~Chen, A.~W.-C. Fu, and D.~W. Cheung, ``Enhancing effectiveness of outlier detections for low density patterns,'' in {\em Advances in Knowledge Discovery and Data Mining: 6th Pacific-Asia conference, PAKDD 2002 Taipei, Taiwan, May 6--8, 2002 proceedings 6}, pp.~535--548, Springer, 2002.

\bibitem{chandola2009anomaly}
V.~Chandola, A.~Banerjee, and V.~Kumar, ``Anomaly detection: A survey,'' {\em ACM Computing Surveys (CSUR)}, vol.~41, no.~3, pp.~1--58, 2009.

\bibitem{ruff2018deep}
L.~Ruff, R.~Vandermeulen, N.~Goernitz, L.~Deecke, S.~A. Siddiqui, A.~Binder, E.~M{\"u}ller, and M.~Kloft, ``Deep one-class classification,'' in {\em International Conference on Machine Learning}, pp.~4393--4402, PMLR, 2018.

\bibitem{barbariol2022review}
T.~Barbariol, F.~D. Chiara, D.~Marcato, and G.~A. Susto, ``A review of tree-based approaches for anomaly detection,'' {\em Control Charts and Machine Learning for Anomaly Detection in Manufacturing}, pp.~149--185, 2022.

\bibitem{belay2023unsupervised}
M.~A. Belay, S.~S. Blakseth, A.~Rasheed, and P.~Salvo~Rossi, ``Unsupervised anomaly detection for {IOT}-based multivariate time series: Existing solutions, performance analysis and future directions,'' {\em Sensors}, vol.~23, no.~5, p.~2844, 2023.

\bibitem{d12}
F.~T. Liu, K.~M. Ting, and Z.-H. Zhou, ``Isolation forest,'' in {\em 2008 Eighth IEEE International Conference On Data Mining}, pp.~413--422, IEEE, 2008.

\bibitem{8888179}
S.~Hariri, M.~C. Kind, and R.~J. Brunner, ``Extended isolation forest,'' {\em IEEE Transactions on Knowledge and Data Engineering}, vol.~33, no.~4, pp.~1479--1489, 2021.

\bibitem{aggarwal2015data}
C.~C. Aggarwal {\em et~al.}, {\em Data mining: the textbook}, vol.~1.
\newblock Springer, 2015.

\bibitem{hemalatha2015minimal}
C.~S. Hemalatha, V.~Vaidehi, and R.~Lakshmi, ``Minimal infrequent pattern based approach for mining outliers in data streams,'' {\em Expert Systems with Applications}, vol.~42, no.~4, pp.~1998--2012, 2015.

\bibitem{cai2020mifi}
S.~Cai, S.~Li, G.~Yuan, S.~Hao, and R.~Sun, ``Mifi-outlier: Minimal infrequent itemset-based outlier detection approach on uncertain data stream,'' {\em Knowledge-Based Systems}, vol.~191, p.~105268, 2020.

\bibitem{xie2018line}
K.~Xie, X.~Li, X.~Wang, J.~Cao, G.~Xie, J.~Wen, D.~Zhang, and Z.~Qin, ``On-line anomaly detection with high accuracy,'' {\em IEEE/ACM Transactions on Networking}, vol.~26, no.~3, pp.~1222--1235, 2018.

\bibitem{6200273}
Y.-J. Lee, Y.-R. Yeh, and Y.-C.~F. Wang, ``Anomaly detection via online oversampling principal component analysis,'' {\em IEEE Transactions on Knowledge and Data Engineering}, vol.~25, no.~7, pp.~1460--1470, 2013.

\bibitem{dong2018threaded}
Y.~Dong and N.~Japkowicz, ``Threaded ensembles of autoencoders for stream learning,'' {\em Computational Intelligence}, vol.~34, no.~1, pp.~261--281, 2018.

\bibitem{zhang2016sliding}
L.~Zhang, J.~Lin, and R.~Karim, ``Sliding window-based fault detection from high-dimensional data streams,'' {\em IEEE Transactions on Systems, Man, and Cybernetics: Systems}, vol.~47, no.~2, pp.~289--303, 2016.

\bibitem{schubert2013generalized}
E.~Schubert, {\em Generalized and efficient outlier detection for spatial, temporal, and high-dimensional data mining}.
\newblock PhD thesis, lmu, 2013.

\bibitem{souiden2022survey}
I.~Souiden, M.~N. Omri, and Z.~Brahmi, ``A survey of outlier detection in high dimensional data streams,'' {\em Computer Science Review}, vol.~44, p.~100463, 2022.

\bibitem{rousseeuw1999fast}
P.~J. Rousseeuw and K.~V. Driessen, ``A fast algorithm for the minimum covariance determinant estimator,'' {\em Technometrics}, vol.~41, no.~3, pp.~212--223, 1999.

\bibitem{ahmad2022constrained}
N.~Ahmad, E.~Schrader, and M.~van Gerven, ``Constrained parameter inference as a principle for learning,'' {\em arXiv preprint arXiv:2203.13203}, 2022.

\bibitem{ahmad2024correlations}
N.~Ahmad, ``Correlations are ruining your gradient descent,'' {\em arXiv preprint arXiv:2407.10780}, 2024.

\bibitem{vajda2024machine}
D.~L. Vajda, T.~V. Do, T.~B{\'e}rczes, and K.~Farkas, ``Machine learning-based real-time anomaly detection using data pre-processing in the telemetry of server farms,'' {\em Scientific Reports}, vol.~14, no.~1, p.~23288, 2024.

\bibitem{HPO_ref}
J.~Soenen, E.~Van~Wolputte, L.~Perini, V.~Vercruyssen, W.~Meert, J.~Davis, and H.~Blockeel, ``The effect of hyperparameter tuning on the comparative evaluation of unsupervised anomaly detection methods,'' in {\em Proceedings of the KDD'21 Workshop on Outlier Detection and Description}, pp.~1--9, Outlier Detection and Description Organising Committee, 2021.

\bibitem{d1}
F.~Ahmed and A.~Courville, ``Detecting semantic anomalies,'' in {\em Proceedings of the AAAI Conference on Artificial Intelligence}, vol.~34, pp.~3154--3162, 2020.

\bibitem{d8}
P.~Gopalan, V.~Sharan, and U.~Wieder, ``Pidforest: anomaly detection via partial identification,'' {\em Advances in Neural Information Processing Systems}, vol.~32, 2019.

\bibitem{d9}
E.~Gutflaish, A.~Kontorovich, S.~Sabato, O.~Biller, and O.~Sofer, ``Temporal anomaly detection: calibrating the surprise,'' in {\em Proceedings of the AAAI Conference on Artificial Intelligence}, vol.~33, pp.~3755--3762, 2019.

\bibitem{d20}
G.~Steinbuss and K.~B{\"o}hm, ``Benchmarking unsupervised outlier detection with realistic synthetic data,'' {\em ACM Transactions on Knowledge Discovery from Data (TKDD)}, vol.~15, no.~4, pp.~1--20, 2021.

\bibitem{d22}
H.~Trittenbach and K.~B{\"o}hm, ``One-class active learning for outlier detection with multiple subspaces,'' in {\em Proceedings of the 28th ACM International Conference on Information and Knowledge Management}, pp.~811--820, 2019.

\bibitem{d23}
H.~Trittenbach, K.~B{\"o}hm, and I.~Assent, ``Active learning of {SVDD} hyperparameter values,'' in {\em 2020 IEEE 7th International Conference on Data Science and Advanced Analytics (DSAA)}, pp.~109--117, IEEE, 2020.

\bibitem{campos2016evaluation}
G.~O. Campos, A.~Zimek, J.~Sander, R.~J. Campello, B.~Micenkov{\'a}, E.~Schubert, I.~Assent, and M.~E. Houle, ``On the evaluation of unsupervised outlier detection: measures, datasets, and an empirical study,'' {\em Data Mining and Knowledge Discovery}, vol.~30, pp.~891--927, 2016.

\bibitem{p3}
L.~Feremans, V.~Vercruyssen, B.~Cule, W.~Meert, and B.~Goethals, ``Pattern-based anomaly detection in mixed-type time series,'' in {\em Machine Learning and Knowledge Discovery in Databases: European Conference, ECML PKDD 2019}, pp.~240--256, Springer, 2020.

\bibitem{p4}
I.~Golan and R.~El-Yaniv, ``Deep anomaly detection using geometric transformations,'' {\em Advances in Neural Information Processing Systems}, vol.~31, 2018.

\bibitem{p7}
T.~Iwata and M.~Yamada, ``Multi-view anomaly detection via robust probabilistic latent variable models,'' {\em Advances in Neural Information Processing Systems}, vol.~29, 2016.

\bibitem{ruff2019deep}
L.~Ruff, R.~A. Vandermeulen, N.~G{\"o}rnitz, A.~Binder, E.~M{\"u}ller, K.-R. M{\"u}ller, and M.~Kloft, ``Deep semi-supervised anomaly detection,'' {\em arXiv preprint arXiv:1906.02694}, 2019.

\bibitem{raza2015ewma}
H.~Raza, G.~Prasad, and Y.~Li, ``{EWMA} model based shift-detection methods for detecting covariate shifts in non-stationary environments,'' {\em Pattern Recognition}, vol.~48, no.~3, pp.~659--669, 2015.

\bibitem{app132312800}
D.~Germano, N.~Sciaraffa, V.~Ronca, A.~Giorgi, G.~Trulli, G.~Borghini, G.~Di~Flumeri, F.~Babiloni, and P.~Aricò, ``Unsupervised detection of covariate shift due to changes in {EEG} headset position: Towards an effective out-of-lab use of passive brain–computer interface,'' {\em Applied Sciences}, vol.~13, no.~23, 2023.

\bibitem{NEURIPS2022_8511d06d}
N.~Wagh, J.~Wei, S.~Rawal, B.~M. Berry, and Y.~Varatharajah, ``Evaluating latent space robustness and uncertainty of {EEG-ML} models under realistic distribution shifts,'' in {\em Advances in Neural Information Processing Systems} (S.~Koyejo, S.~Mohamed, A.~Agarwal, D.~Belgrave, K.~Cho, and A.~Oh, eds.), vol.~35, pp.~21142--21156, Curran Associates, Inc., 2022.

\bibitem{refDamadics}
{IAIR Warsaw University of Technology}, ``Damadics benchmark.'' {https://iair.mchtr.pw.edu.pl/Damadics}, 2002.
\newblock Accessed: 2025-06-23.

\bibitem{goldstein2016comparative}
M.~Goldstein and S.~Uchida, ``A comparative evaluation of unsupervised anomaly detection algorithms for multivariate data,'' {\em PLOS ONE}, vol.~11, no.~4, pp.~1--31, 2016.

\bibitem{xu2018comparison}
X.~Xu, H.~Liu, L.~Li, and M.~Yao, ``A comparison of outlier detection techniques for high-dimensional data,'' {\em International Journal of Computational Intelligence Systems}, vol.~11, no.~1, pp.~652--662, 2018.

\bibitem{zhao2019pyod}
Y.~Zhao, Z.~Nasrullah, and Z.~Li, ``Pyod: A python toolbox for scalable outlier detection,'' {\em Journal of Machine Learning Research}, vol.~20, no.~96, pp.~1--7, 2019.

\end{thebibliography}

\vspace{9cm}

\appendix
\counterwithin{figure}{section}
\counterwithin{table}{section}

\section{Hyperparameter Tuning-based Benchmarking: Dataset Statistics}
\label{sec:app_hpt_statistics}

\begin{table}[h]
\centering
\caption{Original and downsampled datasets statistics.}
\label{tab:HPT_based_L_Exploration}
\resizebox{0.9\textwidth}{!}{%
\begin{tabular}{|l|c|c|c|c|c|c|c|c|}
\hline
\textbf{Dataset} & \textbf{\#Features} & \textbf{\#Samples} & \textbf{\#Outliers} & \textbf{Contamination} & \textbf{Downsampling} & \textbf{\#Samples} & \textbf{\#Outliers} & \textbf{Contamination Ratio} \\
&  &  & & \textbf{Ratio (\%)} & \textbf{(Downsampled)} & \textbf{(Downsampled)} & \textbf{(Downsampled)} & \textbf{Downsampled (\%)} \\
\hline
hepatitis & 19 & 80 & 13 & 16.3 & 0.5 & 40 & 9 & 22.5 \\
vertebral & 6 & 240 & 30 & 12.5 & 0.6 & 144 & 18 & 12.5\\
wine & 13 & 129 & 10 & 7.8 & 0.6 & 77 & 6 & 7.8\\
glass & 9 & 214 & 9 & 4.2 & 0.5 & 107 & 5 & 4.7\\
wbc2 & 9 & 223 & 10 & 4.5 & 0.5 & 111 & 5 & 4.5\\
stamps & 9 & 340 & 31 & 9.1 & 0.6 & 204 & 19 & 9.3\\
parkinson & 22 & 195 & 147 & 75.4 & 0.6 & 117 & 88 & 75.2\\
breastw & 9 & 683 & 239 & 35.0 & 0.6 & 409 & 156 & 38.1\\
pima & 8 & 768 & 268 & 34.9 & 0.6 & 460 & 160 & 34.8\\
wbc & 30 & 378 & 21 & 5.6 & 0.6 & 226 & 12 & 5.3\\
ionosphere & 33 & 351 & 126 & 35.9 & 0.6 & 210 & 76 & 36.2\\
yeast6 & 8 & 1484 & 35 & 2.4 & 0.6 & 890 & 20 & 2.2\\
yeast & 8 & 1484 & 507 & 34.2 & 0.6 & 890 & 294 & 33.0\\
pen-global & 16 & 808 & 90 & 11.1 & 0.6 & 484 & 58 & 12.0\\
vowels & 12 & 1456 & 50 & 3.4 & 0.6 & 873 & 30 & 3.4\\
thyroid & 6 & 3772 & 93 & 2.5 & 0.5 & 1886 & 48 & 2.5\\
wilt & 5 & 4819 & 257 & 5.3 & 0.4 & 1927 & 106 & 5.5\\
cardio & 21 & 1831 & 176 & 9.6 & 0.6 & 1098 & 105 & 9.6\\
annthyroid & 6 & 7200 & 534 & 7.4 & 0.5 & 3600 & 265 & 7.4\\
letter & 32 & 1600 & 100 & 6.3 & 0.6 & 960 & 60 & 6.3\\
fault & 27 & 1941 & 673 & 34.7 & 0.6 & 1164 & 403 & 34.6\\
seismic-bumps & 21 & 2584 & 170 & 6.6 & 0.4 & 1033 & 57 & 5.5\\
pageblocks & 10 & 5393 & 510 & 9.5 & 0.6 & 3235 & 298 & 9.2\\
waveform & 21 & 3442 & 99 & 2.9 & 0.6 & 2065 & 60 & 2.9\\
mammography & 6 & 11183 & 260 & 2.3 & 0.3 & 3354 & 78 & 2.3\\
pen-local & 16 & 6723 & 10 & 0.1 & 0.6 & 4033 & 8 & 0.2\\
arrhythmia & 257 & 452 & 66 & 14.6 & 0.6 & 271 & 43 & 15.9\\
pendigits & 16 & 6870 & 156 & 2.3 & 0.6 & 4122 & 98 & 2.4\\
nasa & 32 & 4687 & 755 & 16.1 & 0.5 & 2343 & 368 & 15.7\\
magic.gamma & 10 & 19020 & 6688 & 35.2 & 0.6 & 11412 & 4013 & 35.2\\
satimage-2 & 36 & 5803 & 71 & 1.2 & 0.6 & 3481 & 42 & 1.2\\
satellite & 36 & 6435 & 2036 & 31.6 & 0.6 & 3861 & 1240 & 32.1\\
landsat & 36 & 6435 & 1333 & 20.7 & 0.6 & 3861 & 817 & 21.2\\
spambase & 57 & 4206 & 1678 & 39.9 & 0.6 & 2523 & 1007 & 39.9\\
optdigits & 62 & 5216 & 150 & 2.9 & 0.6 & 3129 & 90 & 2.9\\
hrss & 18 & 19634 & 4517 & 23.0 & 0.4 & 7853 & 1806 & 23.0\\
smtp & 3 & 95156 & 30 & 0.0 & 0.6 & 57093 & 19 & 0.0\\
shuttle & 9 & 49097 & 3511 & 7.2 & 0.5 & 24548 & 1829 & 7.5\\
musk & 166 & 3062 & 97 & 3.2 & 0.6 & 1837 & 59 & 3.2\\
mnist & 78 & 7603 & 700 & 9.2 & 0.6 & 4561 & 420 & 9.2\\
skin & 3 & 245057 & 50859 & 20.8 & 0.5 & 122528 & 25430 & 20.8\\
mi-f & 40 & 25286 & 2161 & 8.5 & 0.6 & 15171 & 1297 & 8.5\\
mi-v & 40 & 25286 & 3942 & 15.6 & 0.4 & 10114 & 1577 & 15.6\\
aloi & 27 & 49999 & 1507 & 3.0 & 0.6 & 29999 & 905 & 3.0\\
speech & 40 & 3686 & 61 & 1.7 & 0.6 & 2211 & 37 & 1.7\\
http & 3 & 567498 & 2211 & 0.4 & 0.5 & 283749 & 1106 & 0.4\\
campaign & 62 & 41188 & 4640 & 11.3 & 0.6 & 24712 & 2803 & 11.3\\
internetads & 1555 & 1966 & 368 & 18.7 & 0.6 & 1179 & 221 & 18.7\\
cover & 10 & 286048 & 2747 & 1.0 & 0.6 & 171628 & 1655 & 1.0\\
donors & 10 & 619326 & 36710 & 5.9 & 0.5 & 309663 & 18439 & 6.0\\
\hline
\end{tabular}
}
\end{table}

\section{Hyperparameter Tuning-based Benchmarking: Evaluation}

\begin{table}[H]
\centering
\caption{Recommended list of hyperparameters for DAD and SOTA anomaly detection methods according to~\citep{bouman2024unsupervised}.}
\label{tab:methods_hyper}
\resizebox{0.7\textwidth}{!}{%
\begin{tabular}{c|c}
\hline
\textbf{Name} & \textbf{Hyperparameters} \\
\hline
DAD\_s & $\eta={0.8, 0.2, 0.08, 0.02, \dots, 2 \times 10^{-6}}$, $p=0,1$, $\gamma=0.25$ \\
DAD & $\eta={0.8, 0.2, 0.08, 0.02, \dots, 2 \times 10^{-6}}$, $p=0$, $\gamma=0.25$\\
COF & $k=5,10,15,20,25,30$\\
EIF & $n_{trees}=1000, n_{samples}=128,256,512,1024$, no replacement, extension levels: $1,2,3$\\
IF & $n_{trees}=1000, n_{samples}=128,256,512,1024$, no replacement\\
ELOF & maximum LOF score over $k=5,...,30$\\
GMM & $n_{gaussians}=2,...,14$\\
$k$-NN & $k=5,8,...,29$, mean distance\\
$k$th-NN & $k=5,8,...,29$, largest distance\\
LOF & $k=5,8,...,29$\\
LUNAR & $k=5,10,15,20,25,30$\\
MCD & subset fraction$=0.6,0.7,0.8,0.9$\\
OCSVM & RBF kernel, $\nu=0.5,0.6,0.7,0.8,0.9, \gamma=1/d$\\
ODIN & $k=5,8,...,29$\\
PCA & selected PCs explain $>30,50,70,90\%$ of variance\\
SOD & $k=20,30$, $l=10$, $\alpha=0.7,0.9$\\
\hline
\end{tabular}
}
\end{table}

\begin{figure}[H]
    \centering
    \setlength{\tabcolsep}{0.001pt} 
    \begin{tabular}{ccc}
        \subfloat[DAD\_s vs SOTA on all benchmark datasets]{\includegraphics[width=0.45\textwidth]{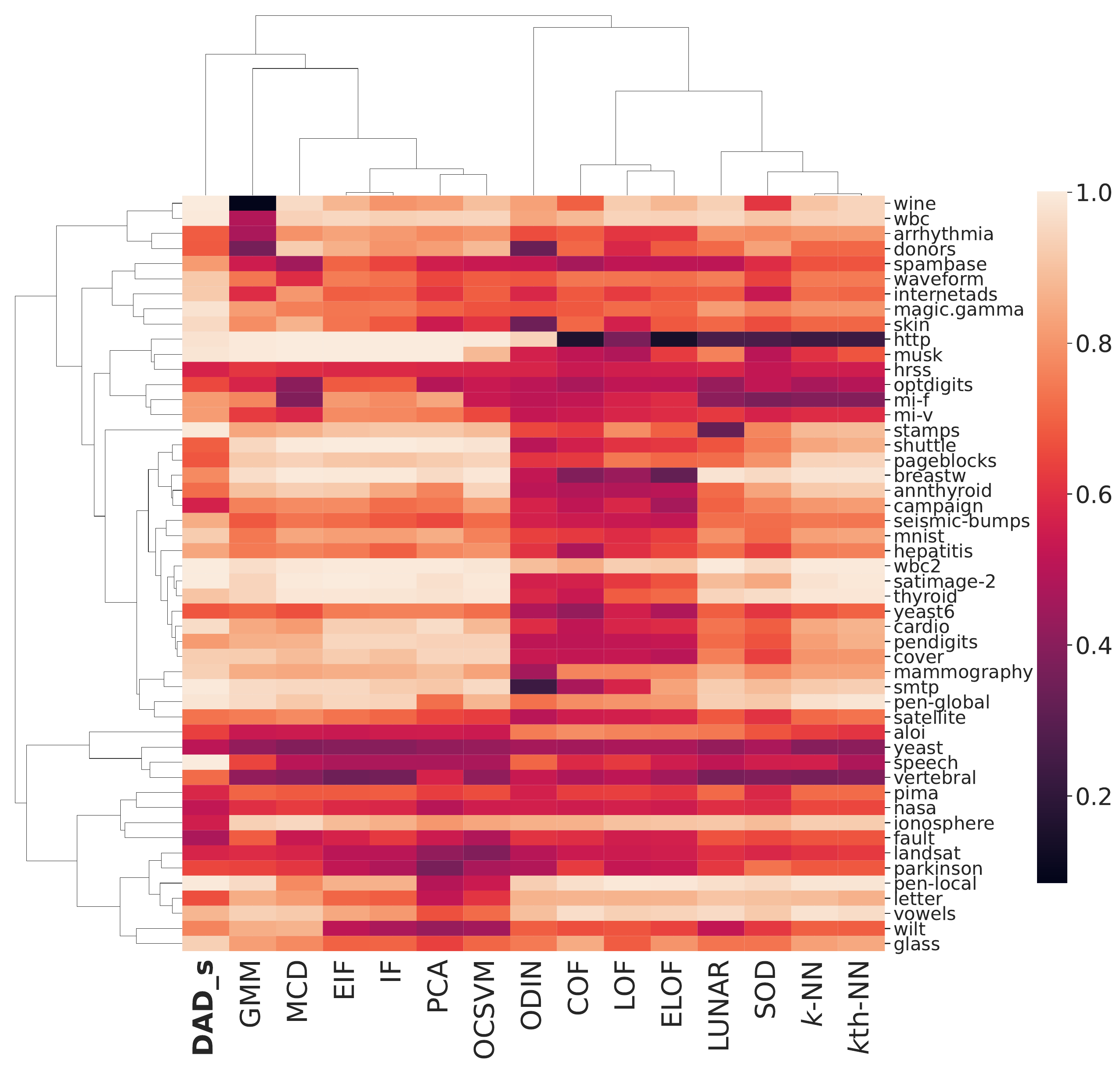}} &    
        \subfloat[DAD vs SOTA on all benchmark datasets]{\includegraphics[width=0.45\textwidth]{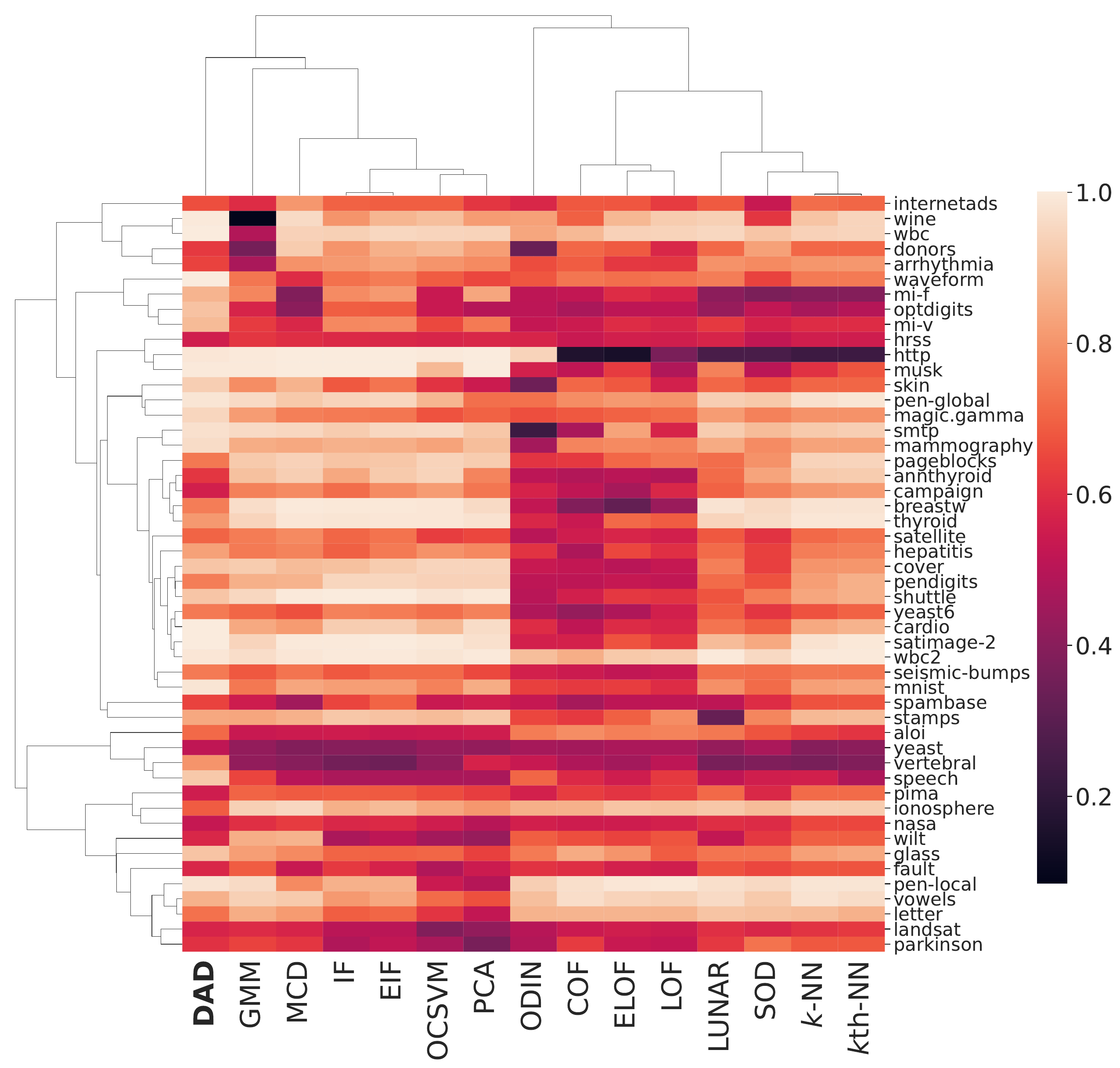}} \\
        \subfloat[DAD\_s vs SOTA on the local cluster]{\includegraphics[width=0.45\textwidth]{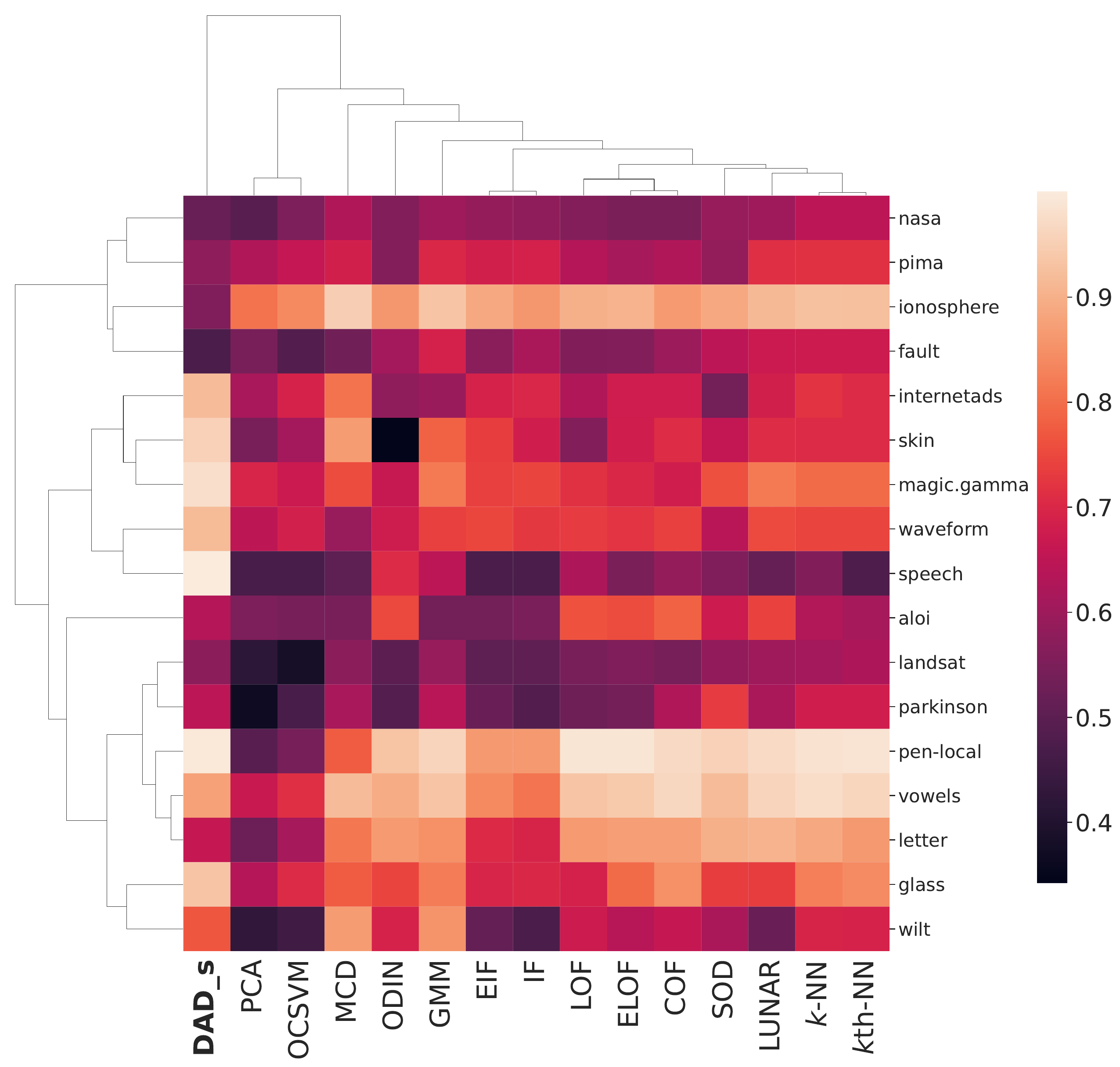}}  &
        \subfloat[DAD vs SOTA on the local cluster]{\includegraphics[width=0.45\textwidth]{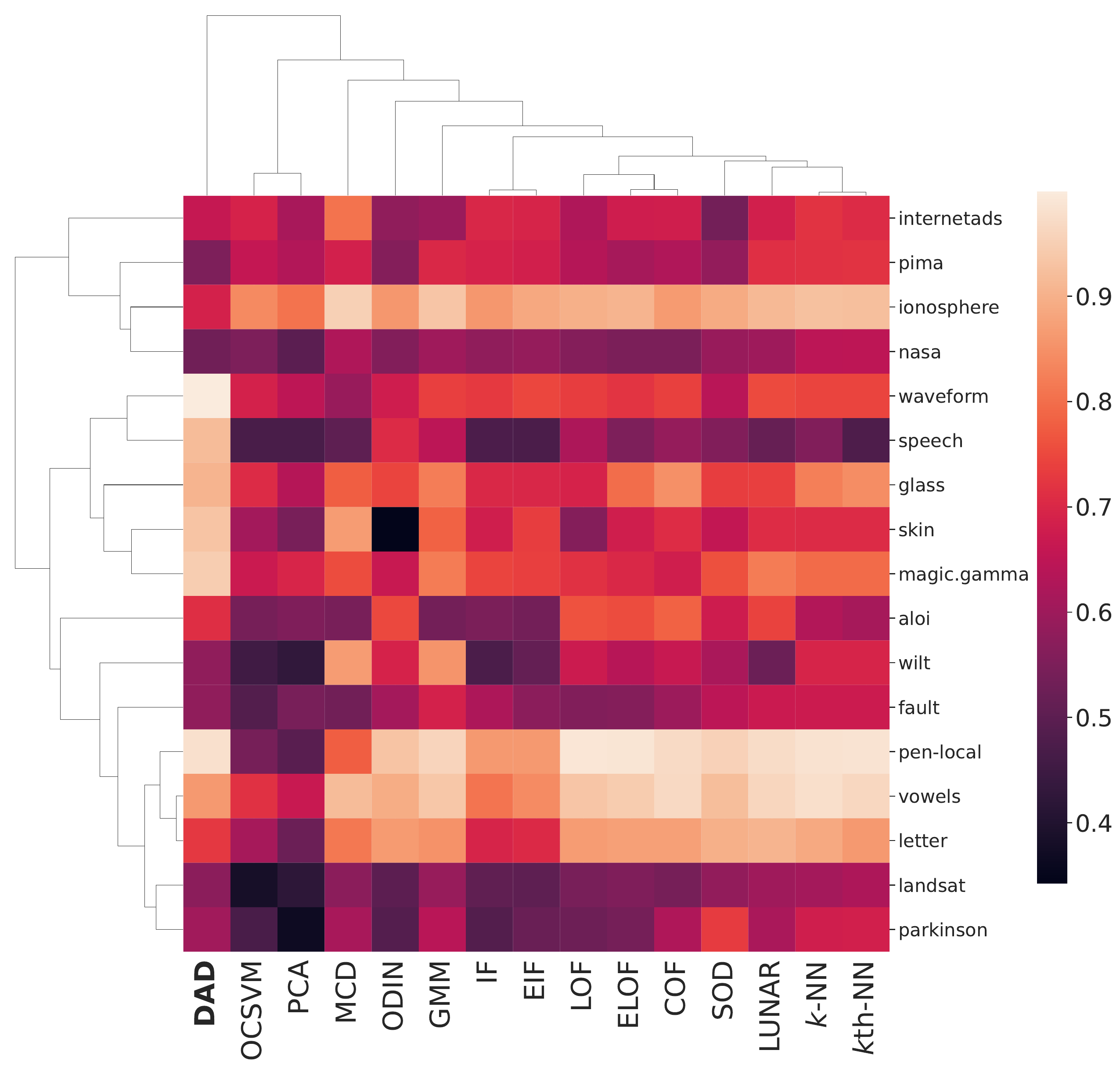}} \\    
        \subfloat[DAD\_s vs SOTA on the global cluster]{\includegraphics[width=0.45\textwidth]{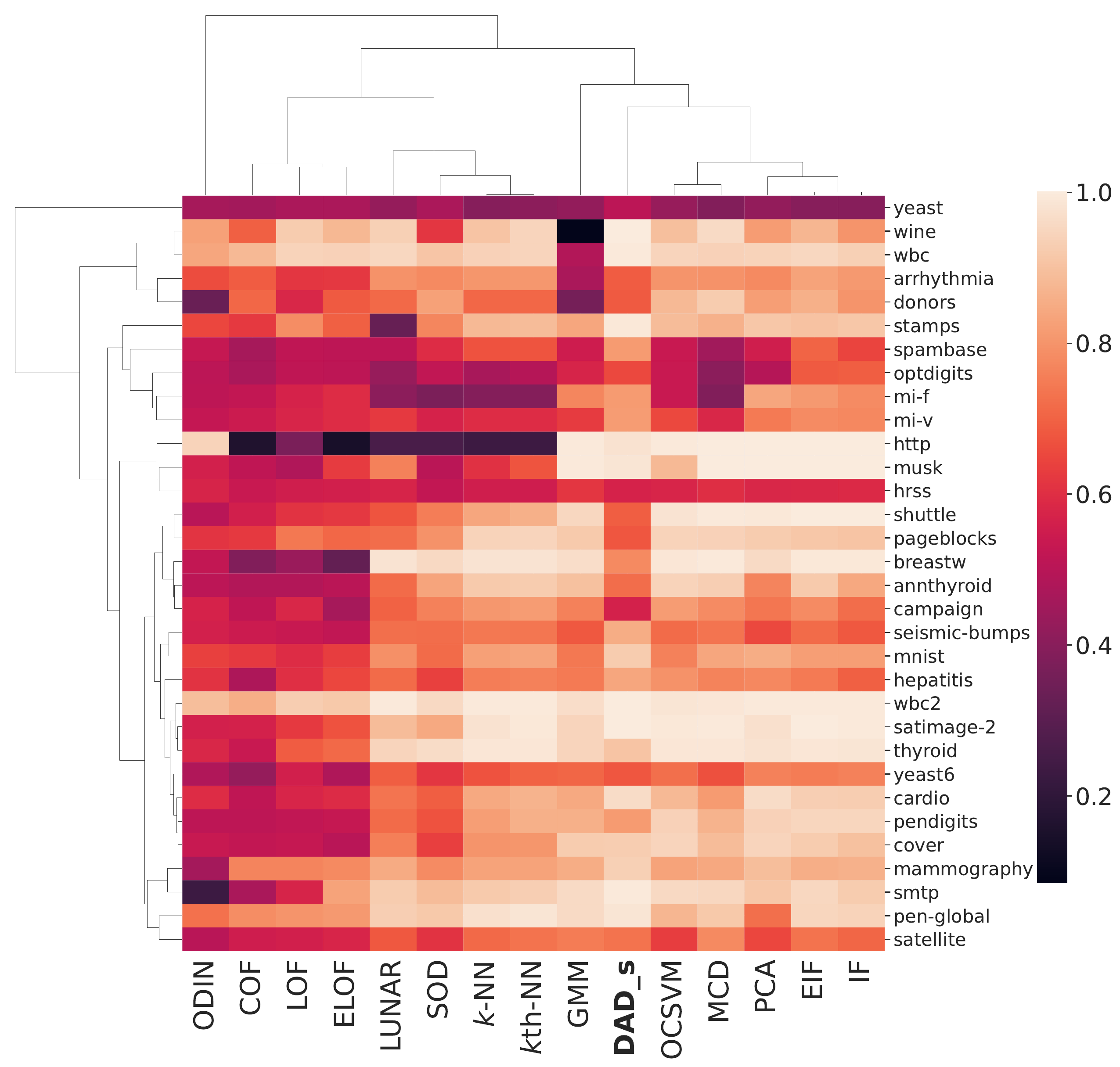}} &
        \subfloat[DAD vs SOTA on the global cluster]{\includegraphics[width=0.45\textwidth]{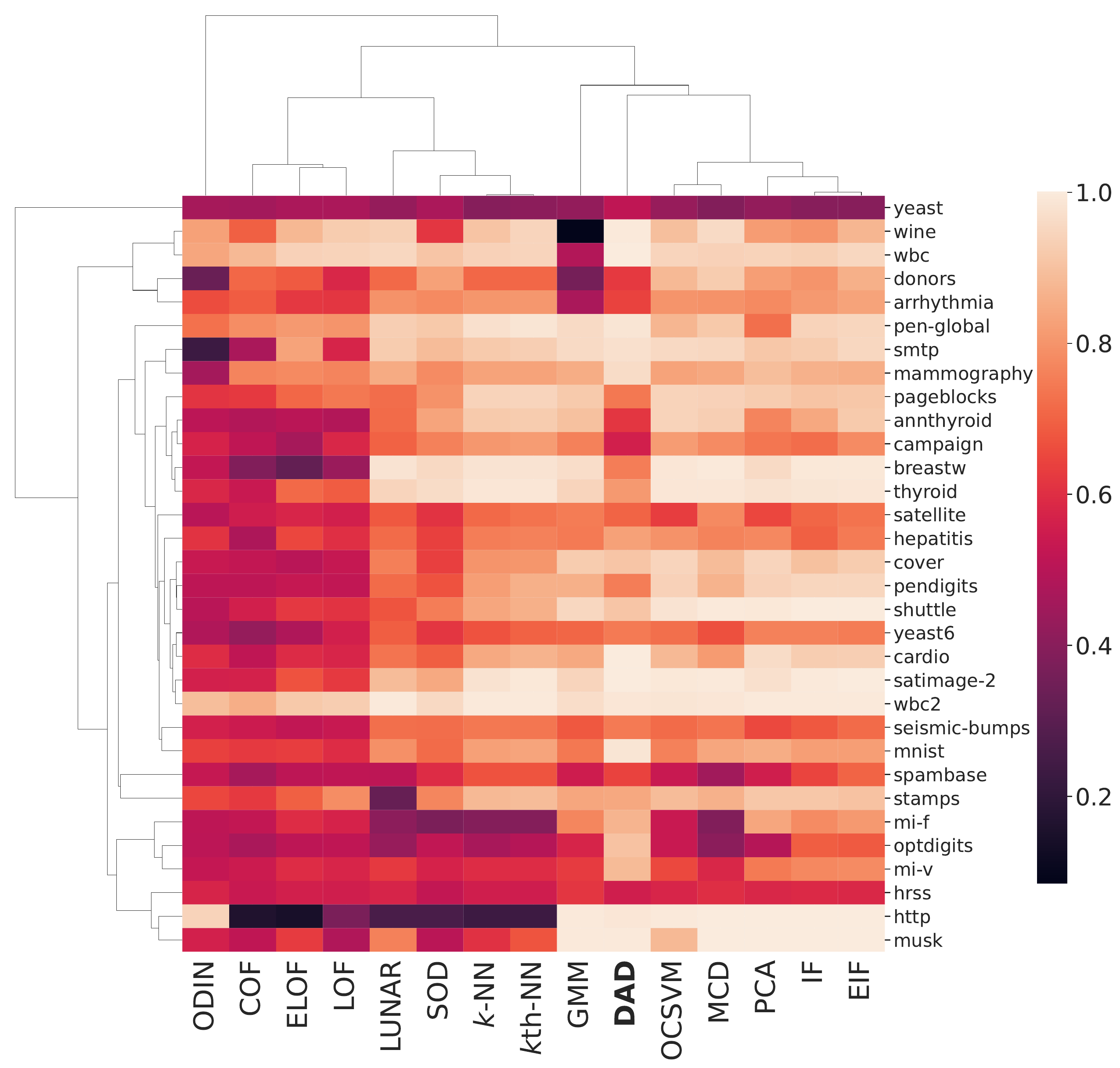}}          
    \end{tabular}
    \caption{Clustered heatmap (hyperparameter tuning-based benchmarking) of method-dataset combination performance: (a) and (b) all benchmark datasets, c) and d) the global cluster, e) and f) the local cluster.}        
    \label{fig:bresult_cluster_hpt_all}
\end{figure}

\section{Traditional Benchmarking}
\label{sec:app_traditional}
\begin{figure}[H]
    \centering
    \setlength{\tabcolsep}{0.001pt} 
    \begin{tabular}{cc}
        \subfloat[DAD\_s and DAD vs SOTA on all benchmark datasets]{\includegraphics[width=0.4999\textwidth]{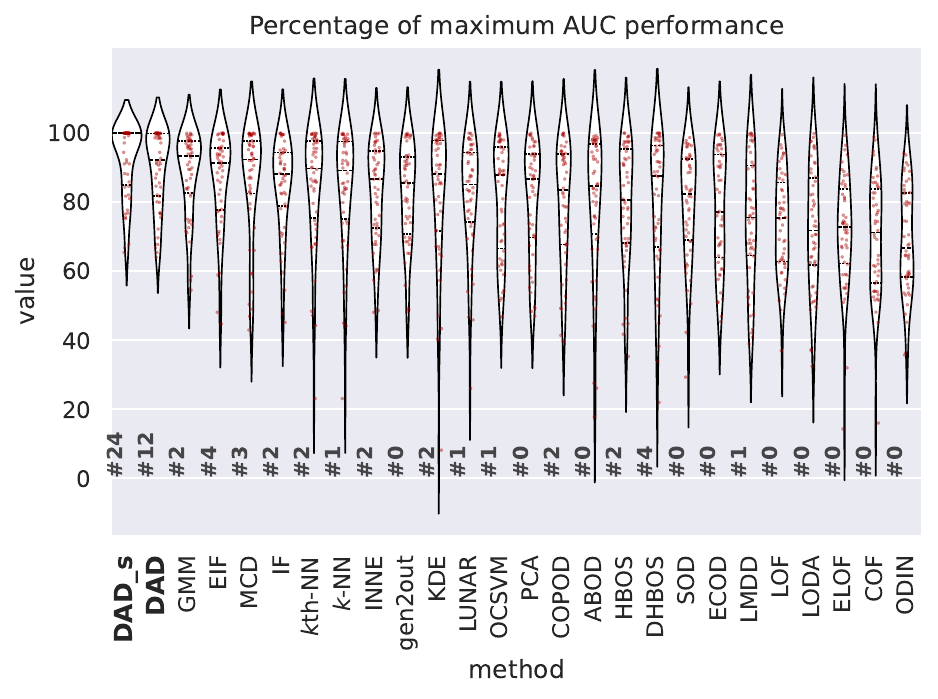}} &    
        \subfloat[Clustered heatmap of all benchmark datasets]{\includegraphics[width=0.4\textwidth]{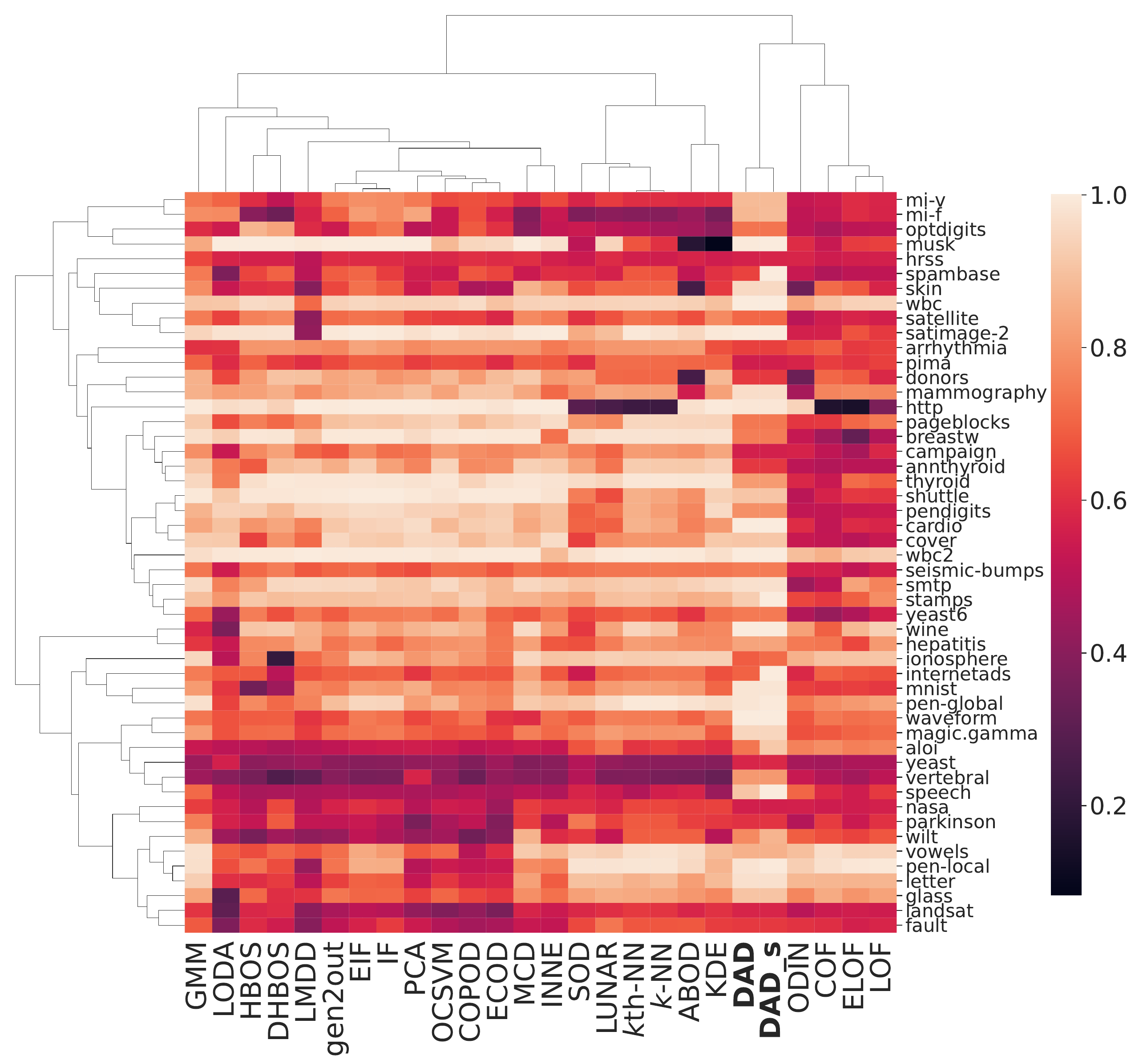}} \\
        \subfloat[DAD\_s and DAD vs SOTA on the global cluster]{\includegraphics[width=0.4999\textwidth]{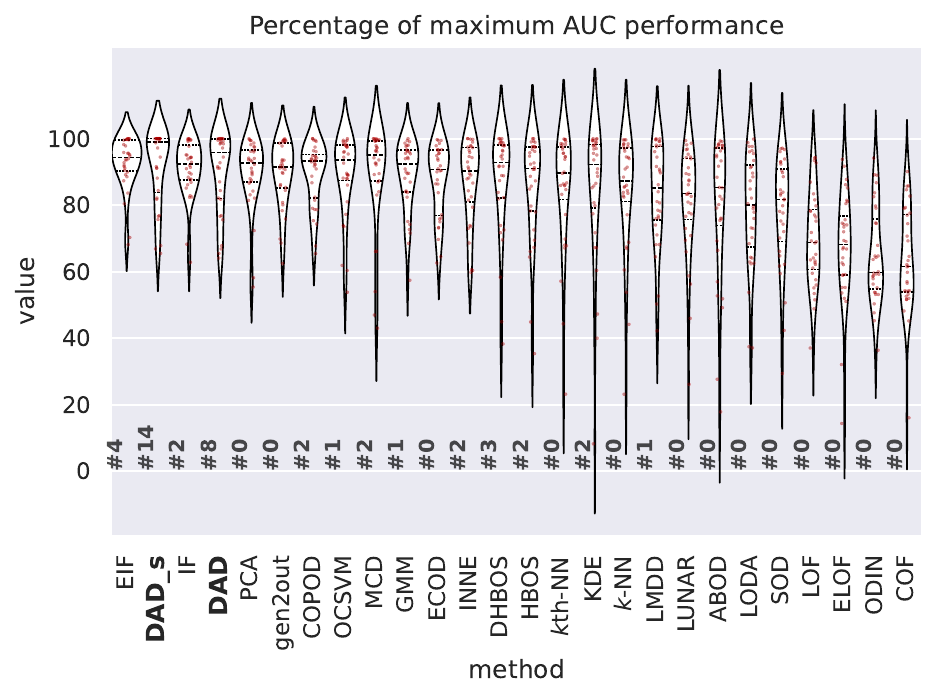}} &    
        \subfloat[Clustered heatmap of global cluster]{\includegraphics[width=0.4\textwidth]{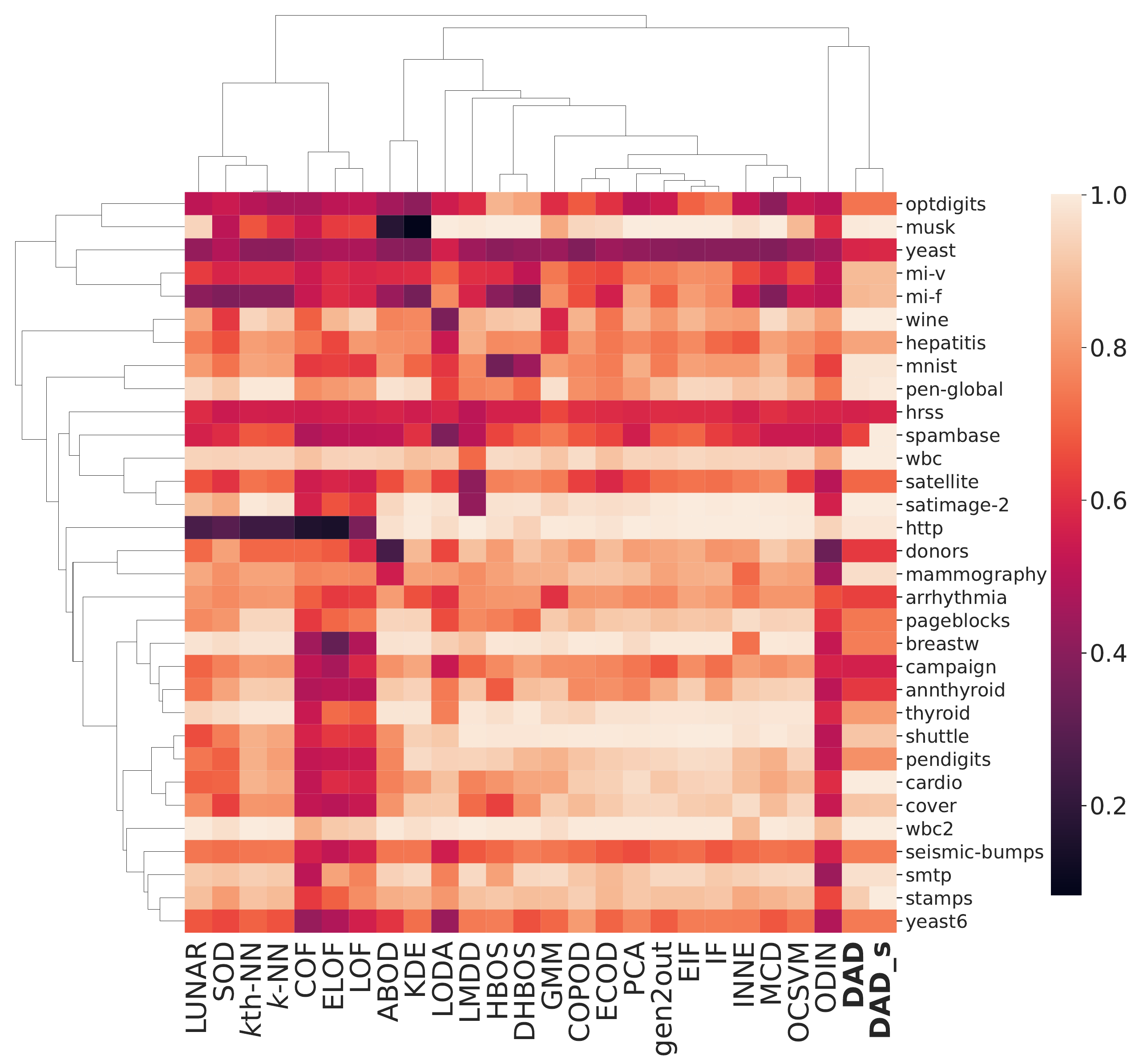}} \\
        \subfloat[DAD\_s and DAD vs SOTA on the local cluster]{\includegraphics[width=0.4999\textwidth]{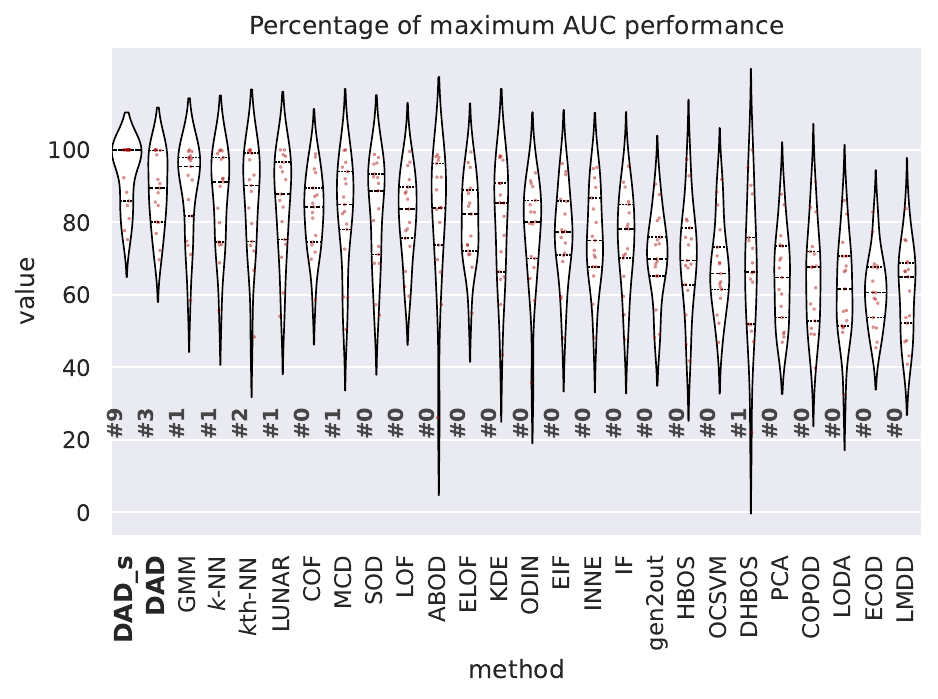}} &    
        \subfloat[Clustered heatmap of local cluster]{\includegraphics[width=0.4\textwidth]{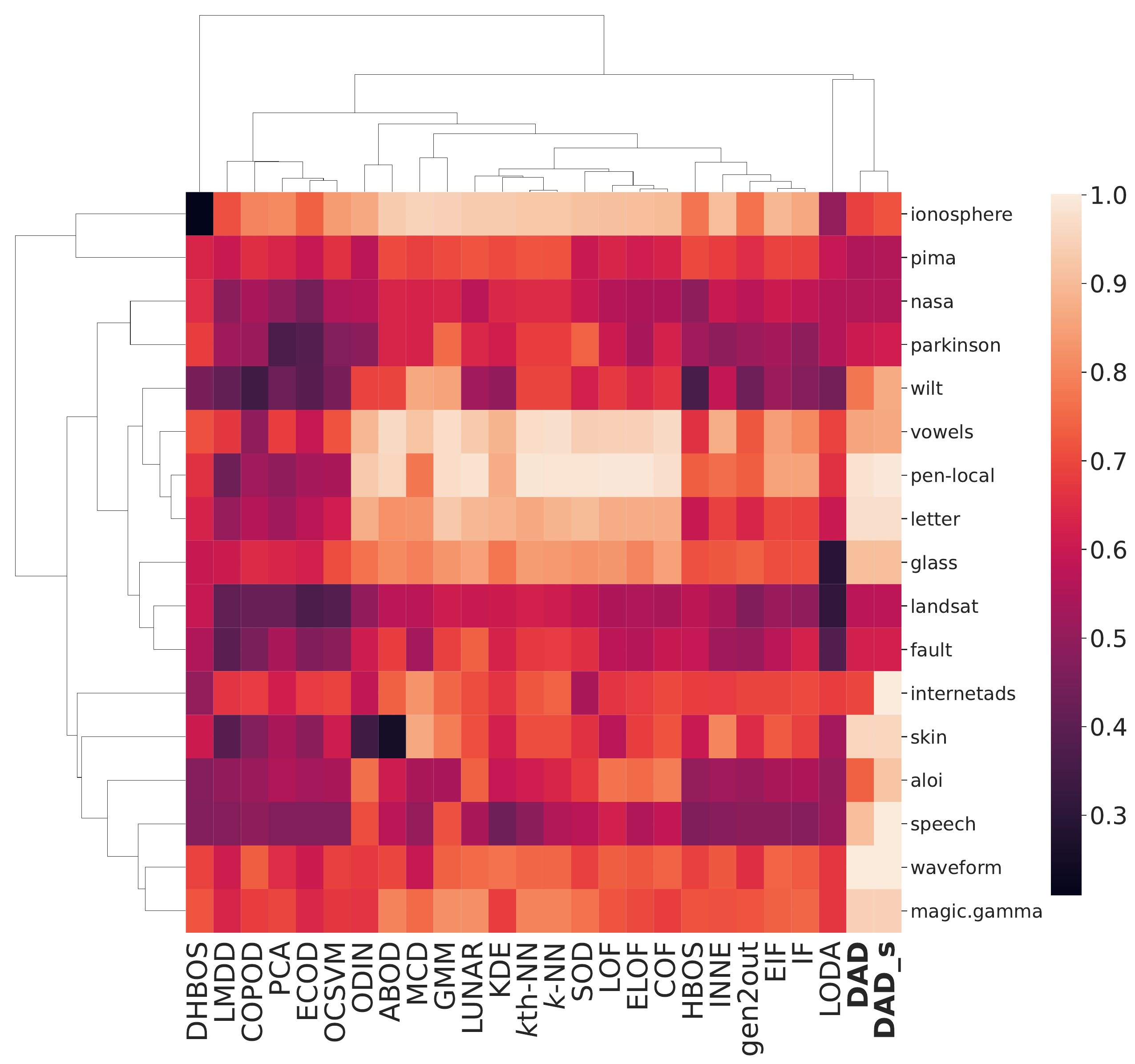}}     
    \end{tabular}
    \caption{AUC performance distribution (peak performance among a set of hyperparameters) and clustered heatmap of each method on: (a) and (b) all benchmark datasets, c) and d) the global cluster, e) and f) the local cluster.}        
    \label{fig:bresult_peak_all}
\end{figure}

\begin{figure}
    \centering
    \setlength{\tabcolsep}{0.001pt} 
    \begin{tabular}{cc}
        \subfloat[DAD\_Auto vs SOTA on all benchmark datasets]{\includegraphics[width=0.4999\textwidth]{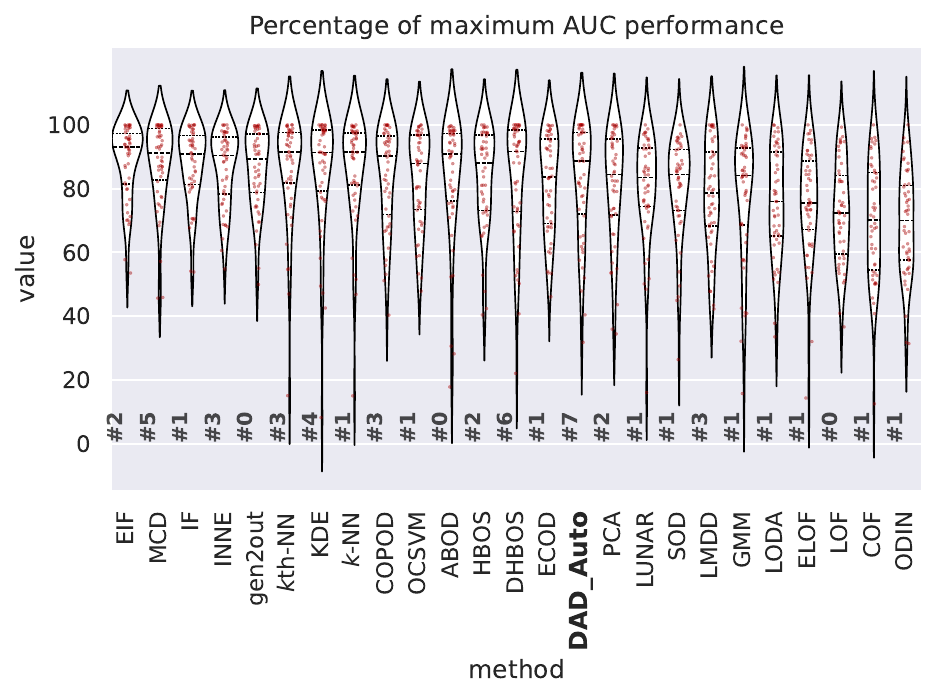}} &    
        \subfloat[Clustered heatmap of all benchmark datasets]{\includegraphics[width=0.4\textwidth]{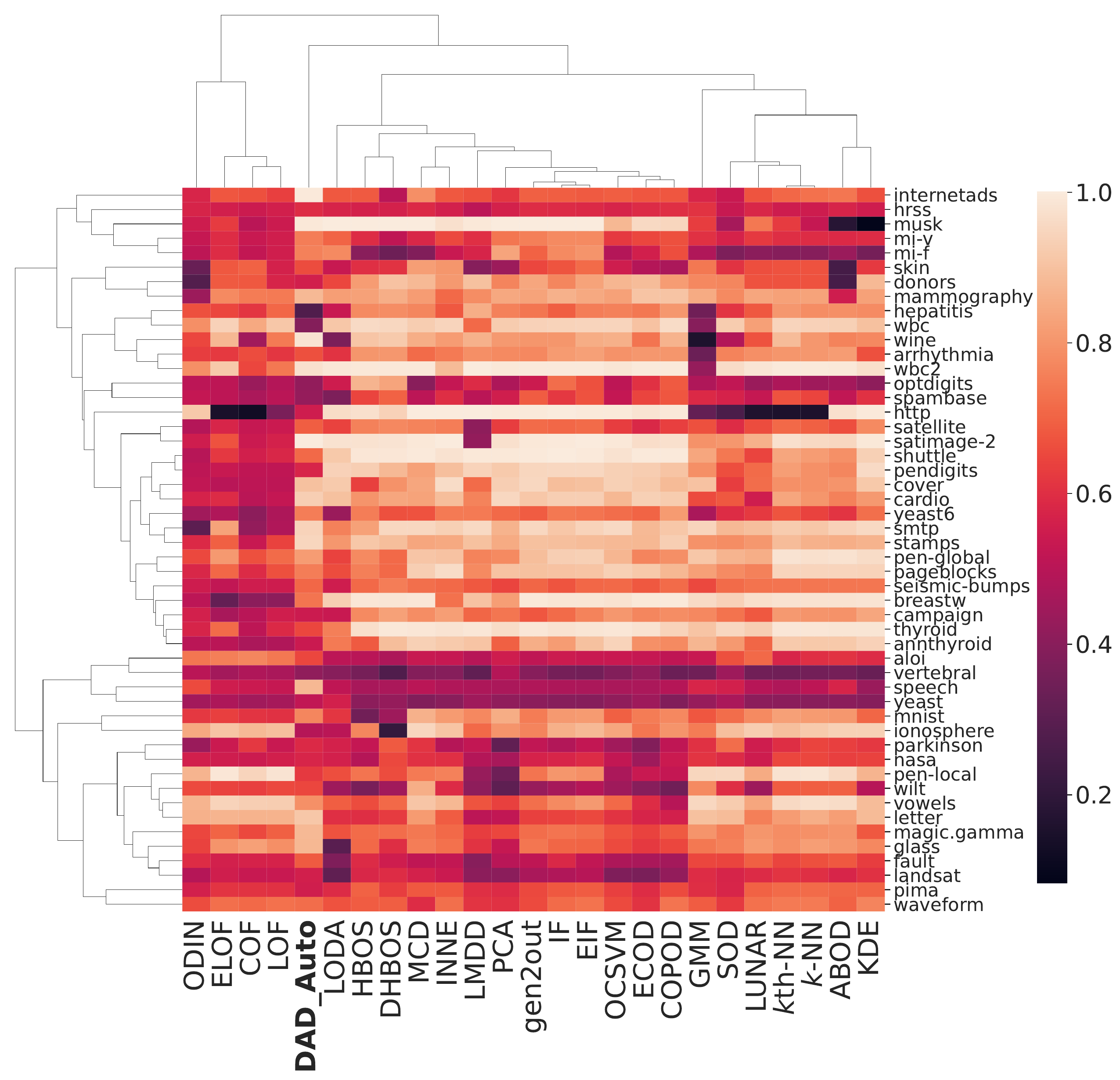}} \\
        \subfloat[DAD\_Auto vs SOTA on the global cluster]{\includegraphics[width=0.4999\textwidth]{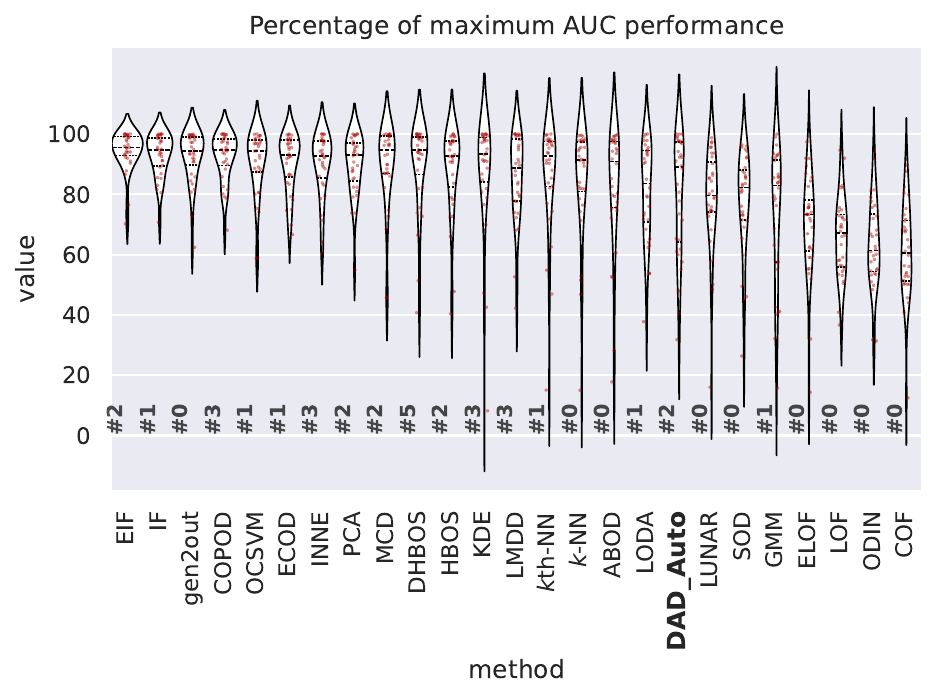}} &    
        \subfloat[Clustered heatmap of global cluster]{\includegraphics[width=0.4\textwidth]{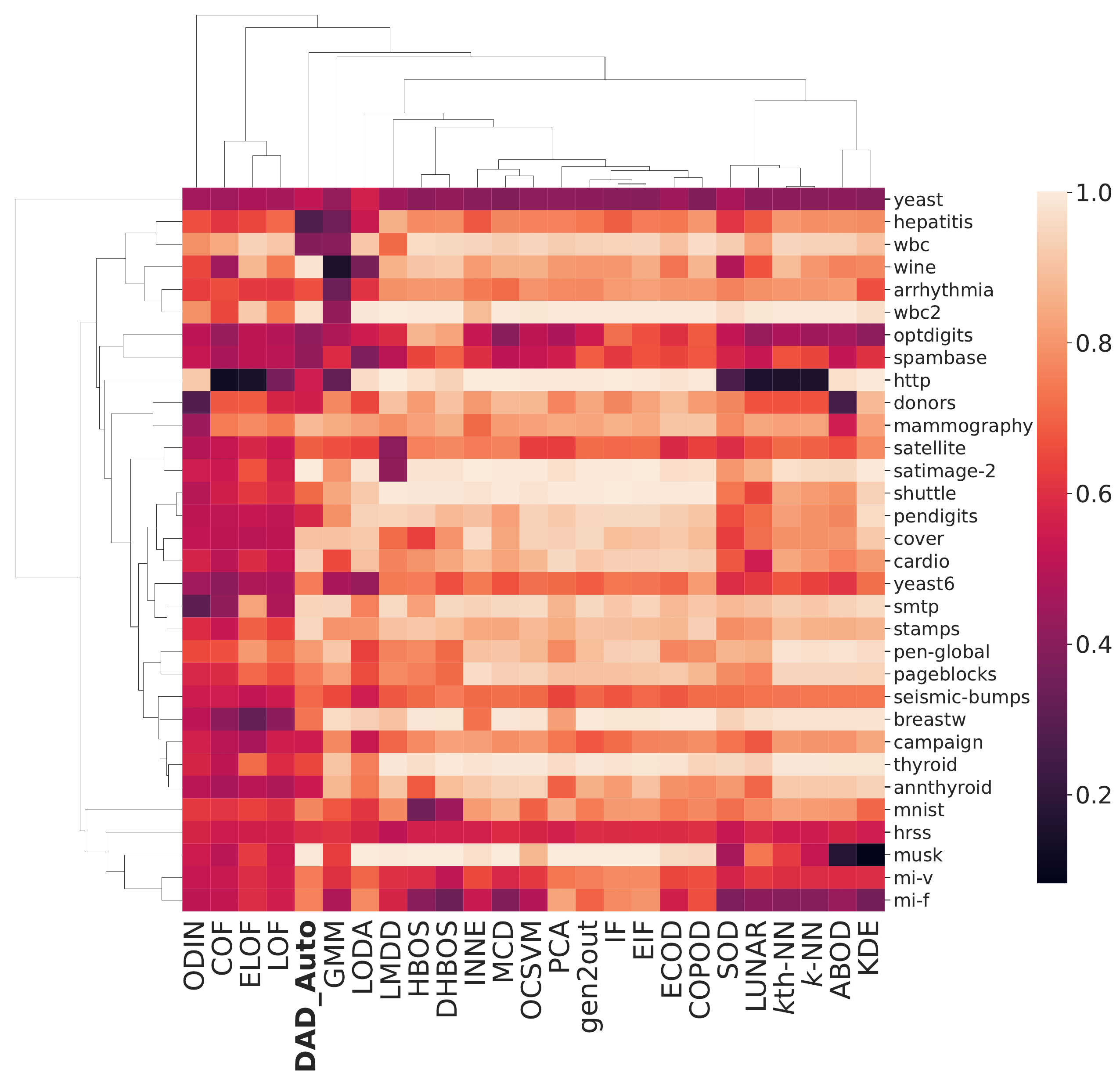}} \\
        \subfloat[DAD\_Auto vs SOTA on the local cluster]{\includegraphics[width=0.4999\textwidth]{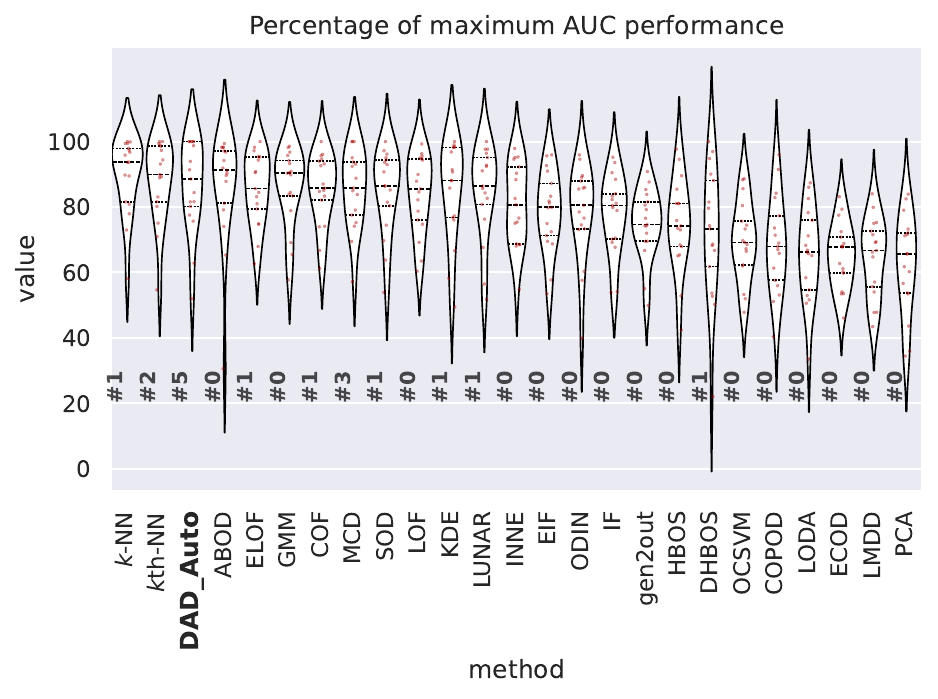}} &    
        \subfloat[Clustered heatmap of local cluster]{\includegraphics[width=0.4\textwidth]{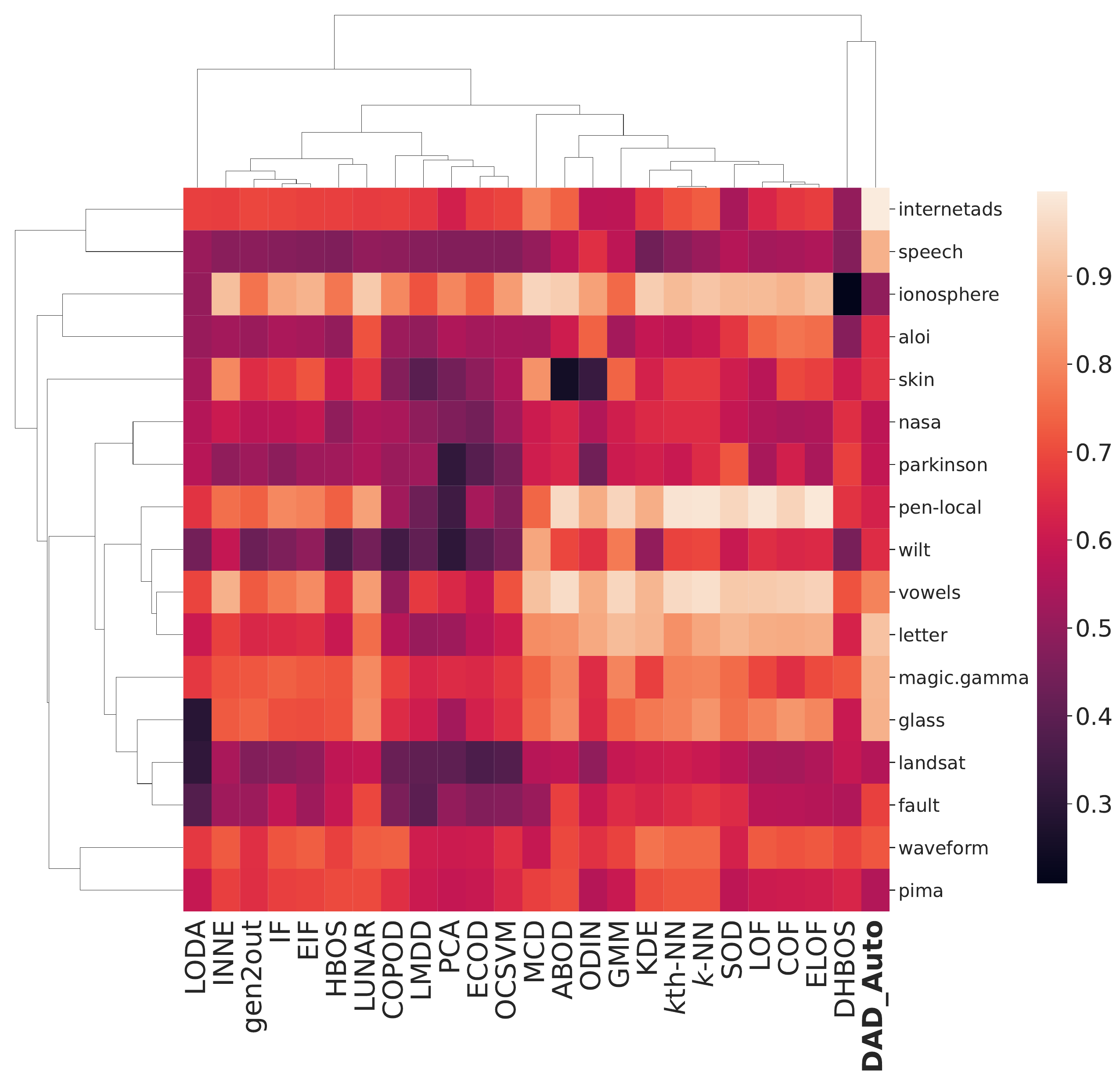}}     
    \end{tabular}
    \caption{AUC performance distribution (average performance across a set of hyperparameters) and clustered heatmap of each method on: (a) and (b) all benchmark datasets, c) and d) the global cluster, e) and f) the local cluster.}        
    \label{fig:bresult_average_all}
\end{figure}

\begin{figure}
    \centering
    \setlength{\tabcolsep}{0.001pt} 
    \begin{tabular}{cc}
        \subfloat[DAD\_Auto vs SOTA on all benchmark datasets]{\includegraphics[width=0.4999\textwidth]{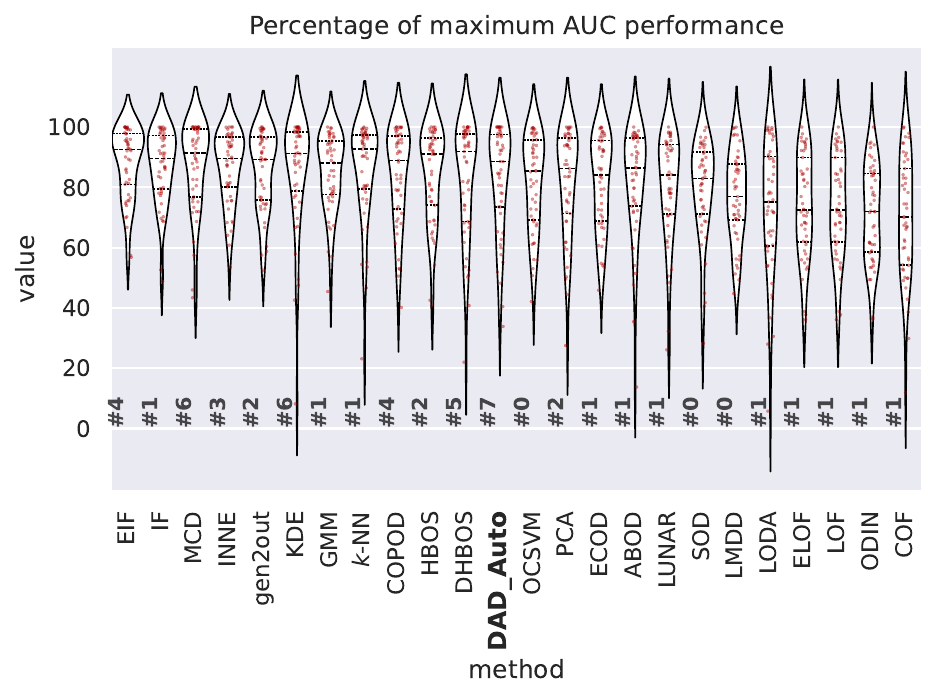}} &    
        \subfloat[Clustered heatmap of all benchmark datasets]{\includegraphics[width=0.4\textwidth]{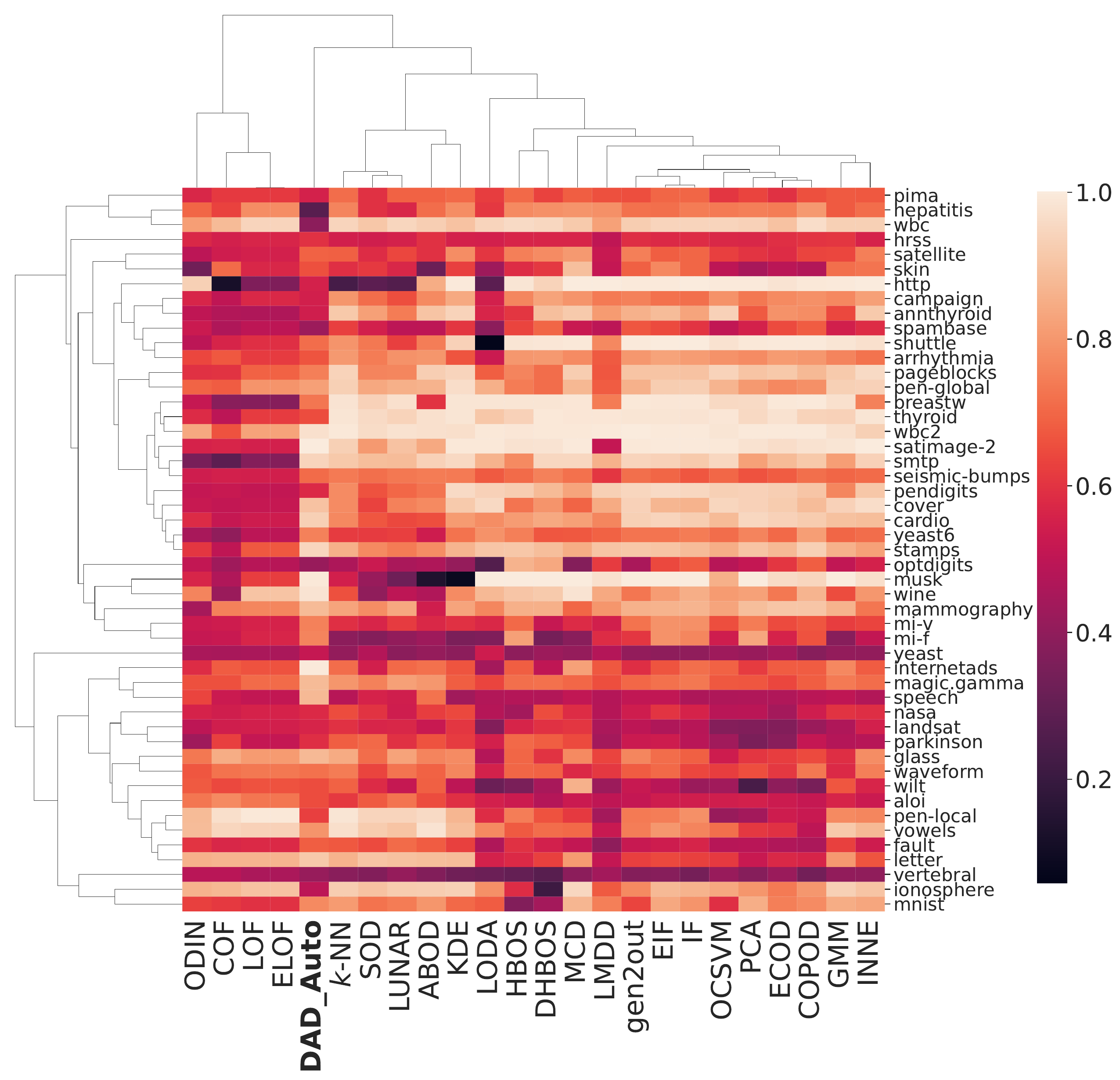}} \\
        \subfloat[DAD\_Auto vs SOTA on the global cluster]{\includegraphics[width=0.4999\textwidth]{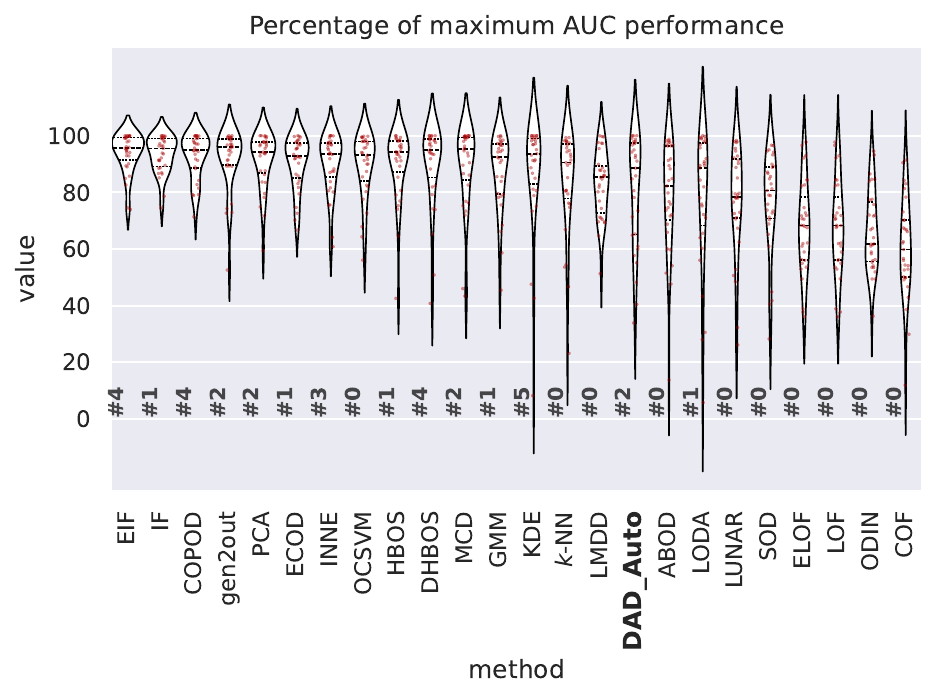}} &    
        \subfloat[Clustered heatmap of global cluster]{\includegraphics[width=0.4\textwidth]{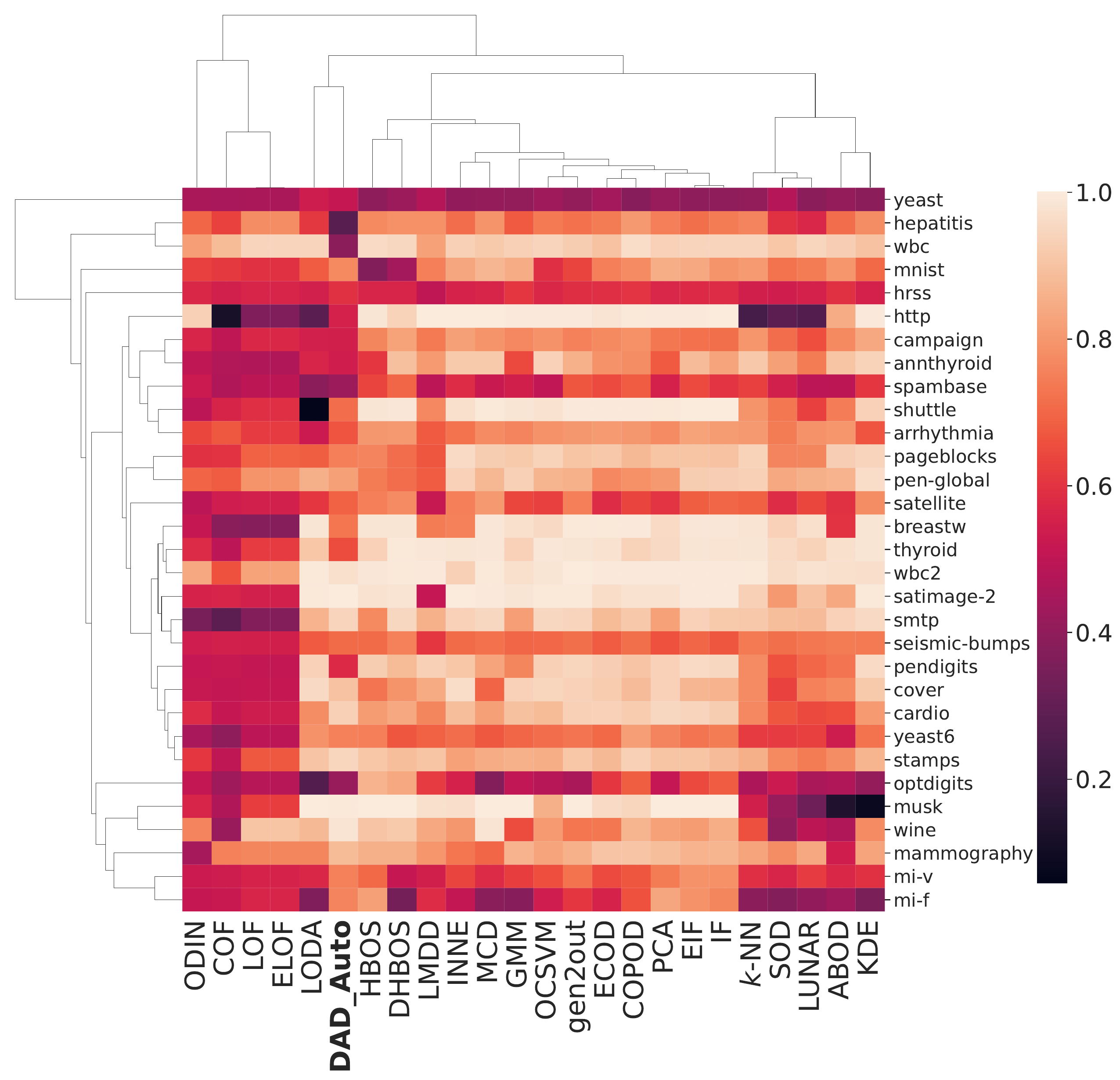}} \\
        \subfloat[DAD\_Auto vs SOTA on the local cluster]{\includegraphics[width=0.4999\textwidth]{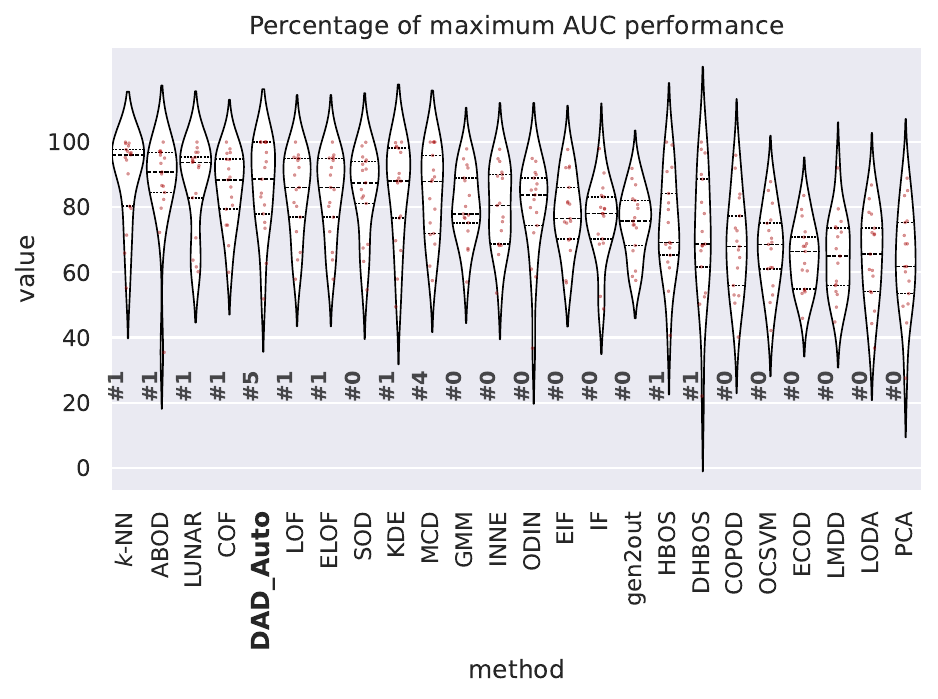}} &    
        \subfloat[Clustered heatmap of local cluster]{\includegraphics[width=0.4\textwidth]{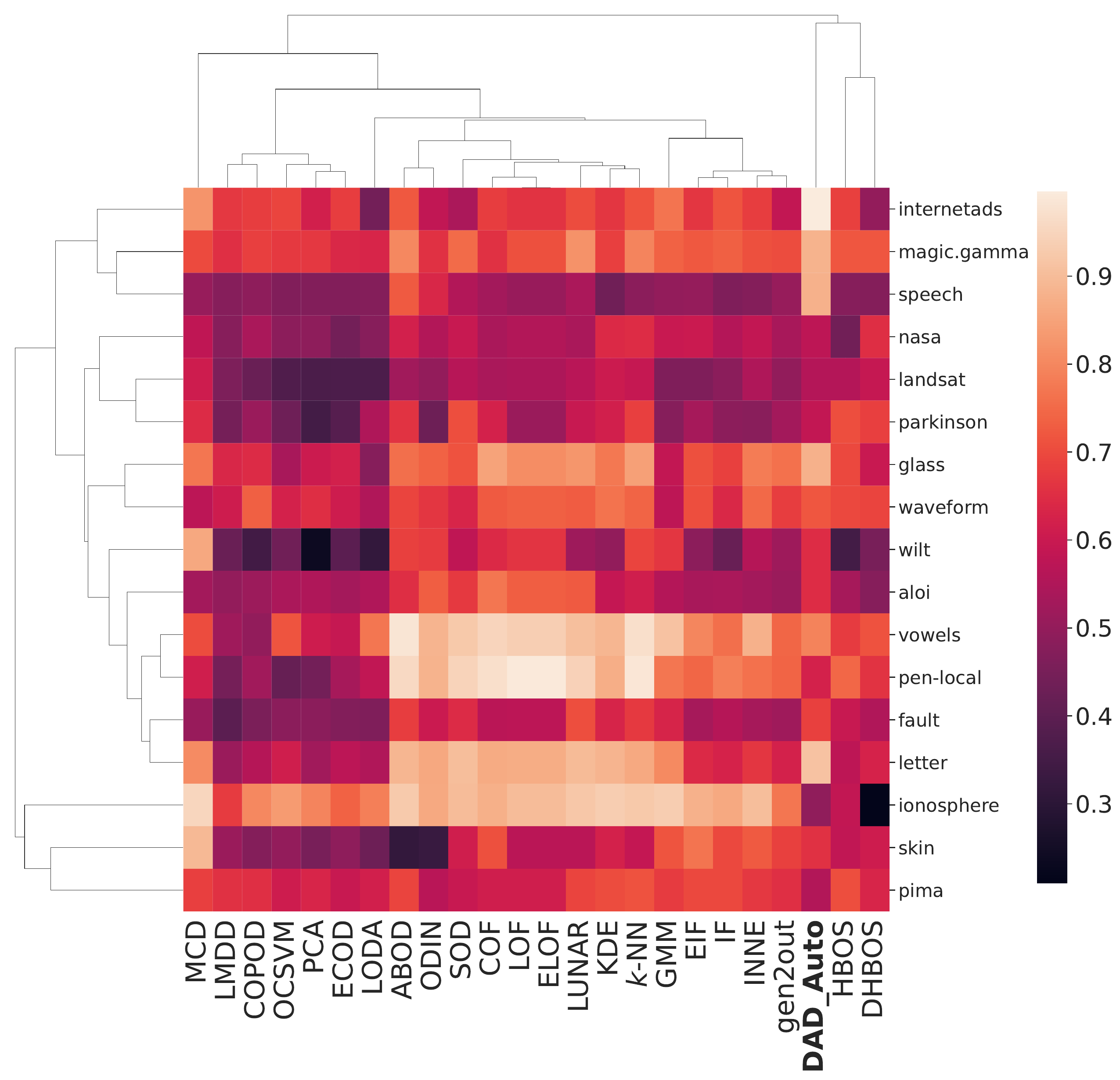}}     
    \end{tabular}
    \caption{AUC performance distribution (performance of default hyperparameter) and clustered heatmap of each method on: (a) and (b) all benchmark datasets, c) and d) the global cluster, e) and f) the local cluster.}        
    \label{fig:bresult_default_all}
\end{figure}

\end{document}